\documentclass[a4paper,fleqn]{cas-sc}
\usepackage{amsmath, amsfonts, mathtools, bm}
\usepackage[authoryear]{natbib}  % 如果使用 natbib 风格
\usepackage{graphicx}
\usepackage{subfig}  % 删除 subfigure，因为它已经被 subfig 包涵盖
\usepackage{float}  % 控制浮动对象
\usepackage{caption}
\usepackage{booktabs}  % 优化表格
\usepackage{algorithm}
\usepackage{algpseudocode}
\usepackage{url}
\usepackage{verbatim}  % 用于多行注释
\usepackage{lipsum}  % 用于生成占位符文本
\usepackage{color, xcolor}  % 颜色相关
\usepackage{textcomp}
\usepackage{hyperref}  % 超链接和引用
\usepackage[english]{babel}  % 设置文档语言为英文
\usepackage[intoc, spanish]{nomencl}  % 如果需要西班牙语的术语表

\usepackage{float}   % 之前用到的宏包
\usepackage{caption} % <--- 这就是你现在加的，用于支持 
\usepackage{subcaption}
\usepackage{tabularx} % 等宽列核心包
\usepackage{array}
\usepackage{dcolumn}

\makenomenclature
\def\tsc#1{\csdef{#1}{\textsc{\lowercase{#1}}\xspace}}
\tsc{WGM}
\tsc{QE}
\newcolumntype{d}{D{.}{.}{-1}}

\begin{document}
\let\WriteBookmarks\relax
\def\floatpagepagefraction{1}
\def\textpagefraction{.001}
%\let\printorcid\relax % 可去掉页面下方的ORCID(s)

% \shorttitle{<short title of the paper for running head>} 
\shorttitle{Dynamic Modeling, Parameter Identification and Numerical Analysis of Cables in Flexibly Connected Dual-AUV Systems} 

% \shortauthors{<short author list for running head>}
\shortauthors{Kuo Chen et al.}
% hybrid drive of data and model
\title[mode = title]{Dynamic Modeling, Parameter Identification and Numerical Analysis of  Flexible Cables in Flexibly Connected Dual-AUV Systems}

\author[1]{Kuo Chen}[type=editor,
    orcid=0000-0002-8707-4533]
\credit{Conceptualization of this study, Methodology, Software}

\author[4]{Minghao Dou}[type=editor,
    orcid=0009-0004-7622-3494]
\credit{Image creation, article grammar, content design, and layout formatting}

\author[1,2]{Qianqi Liu}[type=editor,
    orcid=]
\credit{Prototype experiments and data processing}

\author[4]{Yang An}[type=editor,
    orcid=]
\credit{Article content design}

\author[1,3]{Kai Ren}[type=editor,
    orcid=]
\credit{Article content design}

\author[6]{Zeming WU}[type=editor,
	orcid=0009-0004-1587-4015]
\credit{Article content design}	

\author[1]{Yu Tian}[type=editor,
    orcid=]
\credit{Article content design}

\author[1]{Jie Sun}[type=editor,
    orcid=]
\credit{Article content design}

\author[4,5]{Xinping Wang}[type=editor,
    orcid=]
\credit{Article content design}

\author[1,3]{Zhier Chen}
\cormark[1] 
\credit{Conceptualization of this study and funding acquisition}

\author[1]{Jiancheng Yu}[type=editor,
    orcid=]
\credit{Conceptualization of this study and Funding support
}

\address[1]{the State Key Laboratory of Robotics  and Intelligent Systems, Shenyang Institute of Automation, Chinese Academy of Sciences, Shenyang 110016,}
\address[2]{University of Chinese Academy of Sciences, Beijing 100049, China }
\address[3]{China-Portugal Belt and Road Joint Laboratory on Space $\&$ Sea Technology Advanced Research, Shanghai 201304, China}
\address[4]{Institute of Deep-Sea Science and Engineering, Chinese Academy of Sciences, Sanya 572000, China}
\address[5]{the Innovation Academy for Microsatellites,
Chinese Academy of Sciences, Shanghai 201109}
\address[6]{University of Nottingham Ningbo China, Ningbo 315199}

\cortext[1]{Corresponding author} 

\begin{abstract}
This research presents a dynamic modeling framework and parameter identification methods for describing the highly nonlinear behaviors of flexibly connected dual-AUV systems. The modeling framework is established based on the lumped mass method, integrating axial elasticity, bending stiffness, added mass and hydrodynamic forces, thereby accurately capturing the time-varying response of the forces and cable configurations. To address the difficulty of directly measuring material-related  and hydrodynamic coefficients, this research proposes a parameter identification method that combines the physical model with experimental data. High-precision inversion of the equivalent Young's modulus and hydrodynamic coefficients is performed through tension experiments under multiple configurations, effectively demonstrating that the identified model maintains predictive consistency in various operational conditions. Further numerical analysis indicates that the dynamic properties of flexible cable exhibit significant nonlinear characteristics, which are highly dependent on material property variations and AUV motion conditions. This nonlinear dynamic behavior results in two typical response states, 'slack' and 'taut', which are jointly determined by boundary conditions and hydrodynamic effects, significantly affecting the cable configuration and endpoint loads.  In this research, the dynamics of flexible cables under complex boundary conditions is revealed, providing a theoretical foundation for the design, optimization and further control research of similar systems.
\end{abstract}


% \begin{highlights}
% \item The proposal of a flexible connection dual-AUV system with a cable, addressing the issue of limited payload capacity in long cable deployments in towed systems.
% \item Develop a method for constructing reduced-order models, adeptly projecting high-dimensional state data into a low-dimensional subspace, thereby enhancing the applicability of control system development.
% \item The MEDMD reduced-order modeling method introduced in this paper by executing precise clustering of the input data from the flexible linked dual-AUV system, markedly enhancing the predictive accuracy of the reduced-order model.
% \item The creation of a flexible connection dual-AUV experimental platform and the execution of physical experiments to validate the efficacy of the proposed model.
% \end{highlights}

% Keywords
\begin{keywords}
Towed cable \sep
Flexibly connected dual-AUV system \sep
Dynamic model \sep
\end{keywords}

\maketitle
% Main text
% \nomenclature{\(c\)}{Speed of light in a vacuum}
% \nomenclature{\(h\)}{Planck constant}

% \printnomenclature

\section{Introduction}

In the development of marine engineering equipment and underwater inspection technologies, flexible cables serve as a crucial medium for force transmission and sensing applications \citep{baggeroer2005sonar}, demonstrating significant advantages in operations such as seafloor geological exploration \citep{sun2021review,tossman1979underwater}, oil and gas resource detection\citep{lyutikova2021management, qi2025review}, and underwater archaeological investigation \citep{yamaguchi2022development, thompson2023underwater}. The attitude and dynamic stability of towed cables directly determine the detection accuracy of sensor arrays. Studies have shown that the acoustic and magnetic detecting accuracy is significantly affected when the cable geometric offset reaches ten percent of the signal wavelength \citep{liu2006review,yan2020analysis}. Therefore, maintaining the stable configuration and controllable loads of flexible cables under complex hydrodynamic conditions has become one of the critical challenges limiting the performance of towing systems.

Conventional towing systems usually depend on a single platform providing propulsion, which leads to limited cable length, intense hydrodynamic disturbances, and the AUV formation instability. To enhance formation control capability, our research group proposes a novel flexibly connected dual-AUV system,  which is shown in Figure \ref{Figure1} \citep{chen2025hybrid, chen2024reduced}. Without relying on high-speed navigation, the dual-AUV system forms the time-varying boundary conditions by adjusting relative positions and velocities, thereby maintaining formation stability. This characteristics provides unique advantages in long cable deployment, multi-sensor arrangements, and resistance to flow disturbances. However, compared with the conventional towing system, the introduction of dual boundary excitation leads to more complex nonlinear dynamic behaviors, including asymmetric tension distribution, geometric reconfiguration, and sudden load variations, which create new challenges for  parameter identification and modeling works.
\begin{center}
    \includegraphics[width=9cm]{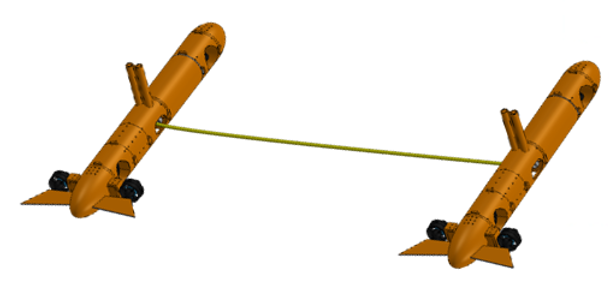}
    \captionof{figure}{Schematic of a flexibly connected dual-AUV system} 
    \label{Figure1}
\end{center}

Several methods have been applied to establish the dynamic model of flexibly connected dual-AUV systems, including the lumped mass method \citep{chen2025hybrid,lian1987theoretical}, the finite element method \citep{gobat2001application, shen2011postprocessing}, and the finite difference method \citep{shen2011postprocessing}. In these methods, the model is transformed into a system of nonlinear equations, which are solved numerically through time integration schemes \citep{chen2025hybrid, lian1987theoretical}. However, these methods are based on over-simplified assumptions, such as a single moving boundary, steady inflow or fixed-end constraints, which lead to fundamental differences from the actual state of dual-AUV systems. In addition, published works in parameter identification of underwater flexible cables are limited. Existing research mainly focuses on identification of drag coefficients or empirical fitting under simplified conditions \citep{quan2016nonlinear,yang2022dynamic,yang2022study,feng2022study} , while the identification of the equivalent Young's modulus is rarely considered. However, as a typical composite structure, flexible cables are composed of outer coating materials and internal communication payloads, which leads to significant challenges in directly measuring the equivalent Young's modulus through traditional stress-strain experiments \citep{fedotov2022mori}. Existing methods, including the Mori-Tanaka model\citep{mori1973average,benveniste1987new} and numerical methods (FEM) \citep{pierard2007micromechanics, doghri2011second}, still require further refinement to reduce the discrepancies between computational predictions and experimental data \citep{fedotov2022mori}.

To address the above limitations, this research presents a unified framework for dynamic modeling and parameter identification of flexibly connected dual-AUV systems.  Firstly, the cable is discretized using the lumped mass method. By integrating axial elasticity, bending stiffness, buoyancy, gravity, added mass, and tangential/normal hydrodynamic forces of each segment, a complete system of nonlinear multi-body dynamic equations is formulated. The AUVs are regarded as moving boundaries while the positions and velocities serve as inputs to drive the dynamic response of the cable system. Secondly, based on the multi-configuration tension dataset  obtained from tank experiments, a fitness function for parameter inversion is constructed. The Genetic Algorithm (GA) is applied under physical constraints to jointly identify the equivalent Young's modulus and hydrodynamic coefficients, deriving the parameter set that accurately reflects the material and fluid characteristics. Finally, based on the modeling results, simulation analysis is conducted to investigate the effects of different material properties and moving boundary conditions. Results indicate that flexible cables exhibit two typical states under the influence of two moving boundary conditions: 'slack' and 'taut'. In the slack state, cables exhibit low tension, bending-dominated behavior, and is more significantly influenced by hydrodynamic effects. In the taut state, cables tend to straighten, leading to rapid  increases in tension and redistribution of boundary loads. This cable configuration switching process between the two states reflects the competition among inertia, gravity and hydrodynamic forces, which is the critical dynamic mechanism that determines the system formation stability and the peak force responses.

\section{Dynamic Modeling of the Flexibly Connected Dual-AUV System}
\subsection{Coordinate settings and modeling assumptions}
A system consisting of two AUV platforms interconnected by a flexible cable is shown in Figure \ref{CoordinateSystems}, which is called the flexibly connected dual-AUV system. The dual-AUV system is able to perform collaborative operations, jointly controlling the flexible cable to execute detection missions in the target area. In this paper, the two AUVs are targeted as $AUV_j$, where $\textit{$j$} \in ({1, 2})$.  To establish the dynamic model that accurately captures the coupled motion within the dual-AUV system, the following reasonable assumptions are introduced:

\begin{enumerate}
  \item Cable homogeneity: Assume the mass and density of the cable are uniformly distributed along the length direction.
  \item Attitude simplification: Assume the pitch and roll angles of the AUV platforms are relatively small. The impact of attitude motions on cable torsional deformation is  negligible.
\end{enumerate}

The lumped mass method is applied in establishing the dynamic model of the flexible cable. The cable is discretized into \textit{$N_c$}	mass points connected by \textit{$N_{c}-1$}	massless spring segments. This simplification facilitates the simulation of the characteristics of continuity, elasticity and flexibility. To establish the mathematical model with higher precision, the coordinate systems defined in this paper include the Earth-fixed coordinate system \textit{$S_E({O_E}{x_E}{y_E}{z_E})$}, which is fixed\\[2pt] to the ground; and a series of local coordinate systems \textit{$S^C_i({O^C_i}{t_i}{n_i}{b_i})$}, $\textit{$i$} \in ({1, 2, \dots, N_c})$, which are  attached to each mass point .

\begin{center}
    \includegraphics[width=9cm]{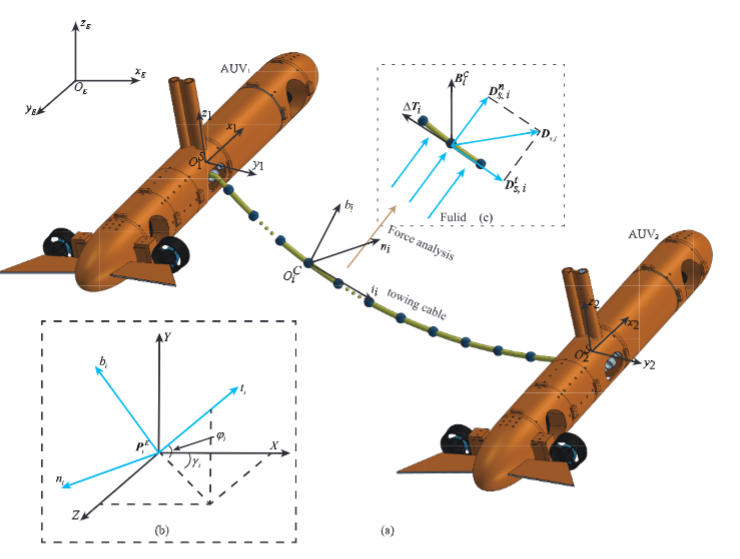}
    \captionsetup{justification=centering} % 局部设置
    \captionof{figure}{Schematic of the flexibly connected dual-AUV system in collaborative operations: \\ (a) Overall view, (b) Coordinate systems, (c) Force analysis on the mass point} 
    \label{CoordinateSystems}
\end{center}

% \begin{center}
%     \includegraphics[width=9cm]{Figure1.png}
%     \captionof{figure}{Schematic of a flexible-connection dual-AUV system} 
%     \label{Figure1}
% \end{center}

In the application of lumped mass method, to simulate the geometry of the flexible cable, the position of $i$-th mass point is defined as \textit{$\boldsymbol{P}^E_i$} = (\textit{$X_i$}, \textit{$Y_i$}, \textit{$Z_i$})$^T$, while the direction of $i$-th spring segment is defined as a set of euler angles ($\varphi_i$, $\gamma_i$). $\varphi_i$ is the rotation angle at which the $S_E$ rotates around the $x_E$ axis to align $z_E$ with the axis $b_i$; Based on the above rotation, $\gamma_i$ is the rotation angle at which the $S_E$ further rotates around the $b_i$ to align $x_E$ axis with  the axis $t_i$, and  $y_E$ with the axis $n_i$. The relative position of the coordinate systems is shown in Figure \ref{CoordinateSystems} (b). The mathematical relationship between $\boldsymbol{P}^E_i$ and euler angles ($\varphi_i$, $\gamma_i$) is:
\begin{equation}
	\varphi_i= \mathrm{arcsin}	(Y_i - \frac{Y_{i-1}}{l_i})
\end{equation}
\begin{equation}
	{\gamma_i}=\left\{ \begin{array}{*{35}{l}}
   {\mathrm{arcsin}} (\dfrac{Z_{i} - Z_{i-1}}{l_i cos{\varphi_i}}) \hspace{3.7em} X_i - X_{i-1} \leq	0 \\[15pt]
   \pi - {\mathrm{arcsin}} (\dfrac{Z_{i-1} - Z_{i-1}}{l_i cos{\varphi_i}}) \hspace{1.2em} X_i - X_{i-1} >	0\\ 
\end{array} \right.	
\end{equation}
where $l_i$ is the length of $i$-th massless spring segment, which is represented as:
\begin{equation}
	l_i= \sqrt{(X_i - X_{i-1})^2 + (Y_i - Y_{i-1})^2 + (Z_i - Z_{i-1})^2}
\end{equation}

Therefore, the transformation matrix between $S^C_i$ and $S_E$ is expressed as:
\begin{equation}
\boldsymbol{R}_E^i = \begin{bmatrix}
\cos\gamma_i \cos\varphi_i & -\cos\gamma_i \sin\varphi_i & \sin\gamma_i  \\
\sin\varphi_i & \cos\varphi_i & 0 \\
-\sin\gamma_i \cos\varphi_i & \sin\gamma_i \sin\varphi_i & \cos\gamma_i
\end{bmatrix}
\end{equation}

To describe the motion conditions of ${AUV}_j$ in the Earth-fixed reference frame, the pose vector is defined as $\boldsymbol{\eta}^E_j = \left[{\boldsymbol{\eta}^E_{1,j}, \boldsymbol{\eta}^E_{2,j}}\right]$, where $\boldsymbol{\eta}^E_{1,j} = \left[ x_j, y_j, z_j \right]^T$ represents the position vector, and $\boldsymbol{\eta}^E_{2,j} = \left[ \phi_j, \theta_j, \psi_j \right]^T$ is the  attitude \\[4pt] vector.  The  attitude includes roll angle $\phi_j$, pitch angle $\theta_j$, and yaw angle $\psi_j$. The position vector of the  connection point between the cable and ${AUV}_j$ in the body-fixed frame is defined as $\boldsymbol{\eta}^S_{c,j} = \left[ x^S_{c,j}, y^S_{c,j}, z^S_{c,j} \right]^T$.
 
\subsection{Dynamic model of the AUV system}
For simplicity of expression, we assume the two AUVs within the flexibly connected dual-AUV system are identical in terms of external geometry and technical parameters. The vector $\boldsymbol{V}^S_{1,j} = \left[{\boldsymbol{V}^S_{1,j}, \boldsymbol{V}^S_{2,j}}\right]$ is defined to describe the translational and rotational velocities in the body-fixed coordinate system, where $\boldsymbol{V}^E_{1,j} = \left[ u_j, v_j, w_j \right]^T$ is the translational velocity vector, and $\boldsymbol{V}^E_{2,j} = \left[ p_j, q_j, r_j \right]^T$ is the rotational velocity vector. Based on the above derivation, \\[4pt] the kinematic model of ${AUV}_j$ is further formulated as follow:
\begin{equation}
	\dot{\boldsymbol{\eta}}_j^E= \boldsymbol{R}_E^S(\phi_j, \theta_j, \psi_j)
\end{equation}
where $\boldsymbol{R_E}^S(\phi_j, \theta_j, \psi_j)$ is the rotational matrix to realize the velocity transformation from the body-fixed coordinate system to the Earth-fixed system. The complete structure of the rotational matrix is:
\begin{equation}
\boldsymbol{R}_{E}^S(\phi_j, \theta_j, \psi_j) = \begin{bmatrix} \boldsymbol{R}_{E,1}^S(\phi_j, \theta_j, \psi_j) & {0}_{3 \times 3} \\ {0}_{3 \times 3} & \boldsymbol{R}_{E,2}^S(\phi_j, \theta_j, \psi_j) 
\end{bmatrix}
\end{equation}
where $\boldsymbol{R}_{E,1}^S(\phi_j, \theta_j, \psi_j)$ is the rotational matrix for linear velocities, which is represented as: 
\begin{equation}
\boldsymbol{R}_{E,1}^S(\phi_j, \theta_j, \psi_j) =
\begin{bmatrix}
\cos\theta_j \cos\psi_j & \sin\phi_j \sin\theta_j \cos\psi_j - \cos\phi_j \sin\psi_j & \cos\phi_j \sin\theta_j \cos\psi_j + \sin\phi_j \sin\psi_j \\
\cos\theta_j \sin\psi_j & \sin\phi_j \sin\theta_j \sin\psi_j + \cos\phi_j \cos\psi_j & \cos\phi_j \sin\theta_j \sin\psi_j - \sin\phi_j \cos\psi_j \\
-\sin\theta_j & \sin\phi_j \cos\theta_j & \cos\phi_j \cos\theta_j
\end{bmatrix}
\end{equation}
$\boldsymbol{R}_{E,2}^S(\phi_j, \theta_j, \psi_j)$ is the matrix for angular velocities, the detailed structure of which is shown as follow:
\begin{equation}
\boldsymbol{R}_{E,2}^S(\phi_j, \theta_j, \psi_j) =
\begin{bmatrix}
1 & \sin\phi_j \tan\theta_j & \cos\phi_j \tan\theta_j \\[2pt]
0 & \cos\phi_j & -\sin\phi_j \\[2pt]
0 & \dfrac{\sin\phi_j}{\cos\theta_j} & \dfrac{\cos\phi_j}{\cos\theta_j}
\end{bmatrix}
\end{equation}

The mathematical model of AUV dynamics is formulated as:
\begin{equation}
	\boldsymbol{\tau}_{c,j} + \boldsymbol{T}_{prop} + \boldsymbol{\tau}_{H,j} = - \boldsymbol{M}_A \boldsymbol{\dot V}_j^S - \boldsymbol{C}_A(\boldsymbol{V}_j^S)\boldsymbol{V}_j^S - \boldsymbol{D}_H(\boldsymbol{V}_j^S)\boldsymbol{V}_j^S
\end{equation}
where $\boldsymbol{\tau}_{H,j} = \left[\boldsymbol{\tau}_{H,u,j}, {\tau}_{H,v,j}, {\tau}_{H,w,j}, {\tau}_{H,p,j}, {\tau}_{H,q,j}, {\tau}_{H,r,j} \right]^T$ is the hydrodynamic forces and moments in Six Degrees of Freedom (DoF); $\boldsymbol{T}_{prop}$ is the thrust from propulsion system; $\boldsymbol{\tau}_{c,j} = \left[\boldsymbol{F_{c,j}},  \boldsymbol{M_{c,j}}\right]^T$ contains the forces and moments exerted by the flexible cable on the ${AUV}_j$, where $\boldsymbol{F}_{c,j} = \left[ F_{c,j,X}, F_{c,j,Y}, F_{c,j,Z}\right]^T$ and $\boldsymbol{M}_{c,j} = \left[ M_{c,j,K}, M_{c,j,M}, M_{c,j,N}\right]^T$. \\[4pt] The  derivation of $\boldsymbol{F}_{c,j}$ and $\boldsymbol{M}_{c,j}$ is given by:
\begin{equation}
\boldsymbol{F}_{c,j} = 
\begin{bmatrix}
F_{c,j,X} \\
F_{c,j,Y} \\
F_{c,j,Z}
\end{bmatrix}
= \boldsymbol{R}_E^i \boldsymbol{R}_E^S(\phi_j, \theta_j, \psi_j)
\begin{bmatrix}
T_{AUVj} \\
0 \\
0
\end{bmatrix}
\end{equation}
\begin{equation}
\boldsymbol{M}_{c,j} = 
\begin{bmatrix}
M_{c,j,K} \\
M_{c,j,M} \\
M_{c,j,N}
\end{bmatrix}
= \boldsymbol{\eta}_{c,j}^S \times \boldsymbol{F}_{c,j} =
\begin{bmatrix}
y_{c,j}^S F_{c,j,Z} - z_{c,j}^S F_{c,j,Y} \\
z_{c,j}^S F_{c,j,X} - x_{c,j}^S F_{c,j,Z} \\
x_{c,j}^S F_{c,j,Y} - y_{c,j}^S F_{c,j,X}
\end{bmatrix}
\end{equation}
where $T_{AUV_j}$ is the tension exerted on the $AUV_j$. The added mass matrix $\boldsymbol{M}_A$ of the ${AUV_j}$ is expressed as:
\begin{equation}
\boldsymbol{M}_A = 
\begin{bmatrix}
X_{\dot{u}} & 0 & 0 & 0 & 0 & 0 \\
0 & Y_{\dot{v}} & 0 & 0 & 0 & Y_{\dot{r}} \\
0 & 0 & Z_{\dot{w}} & 0 & Z_{\dot{q}} & 0 \\
0 & 0 & 0 & K_{\dot{p}} & 0 & 0 \\
0 & 0 & M_{\dot{w}} & 0 & M_{\dot{q}} & 0 \\
0 & N_{\dot{v}} & 0 & 0 & 0 & N_{\dot{r}}
\end{bmatrix}
\end{equation}
where $X_{\dot{u}}$, $Y_{\dot{v}}$, $Y_{\dot{r}}$, $Z_{\dot{w}}$, $Z_{\dot{q}}$, $K_{\dot{p}}$, $M_{\dot{w}}$, $M_{\dot{q}}$, $N_{\dot{v}}$, $N_{\dot{r}}$ are added mass coefficients. $\boldsymbol{C}_A(\boldsymbol{V}_j^S)$ is the corresponding coriolis matrix, which is formulated as:
\begin{equation}
  \boldsymbol{C}_A(\boldsymbol{V}_j^S) = -
\begin{bmatrix}
0 & 0 & 0 & 0 & -Z_{\dot{w}} w_j  & Y_{\dot{v}} v_j  \\
0 & 0 & 0 & Z_{\dot{w}} w_j   & 0 & -X_{\dot{u}} u_j  \\
0 & 0 & 0 & -Y_{\dot{v}} v_j  & X_{\dot{u}} u_j  & 0 \\
0 & -Z_{\dot{w}} w_j  &   Y_{\dot{v}} v_j & 0 & - N_{\dot{r}} r_j &   M_{\dot{q}} q_j \\
Z_{\dot{w}} w_j  & 0 & -X_{\dot{u}} u_j & N_{\dot{r}} r_j  & 0 & -K_{\dot{p}} p_j \\
-Y_{\dot{v}} v_j  & X_{\dot{u}} u_j  & 0 &- M_{\dot{q}} q_j  &  K_{\dot{p}} p_j & 0
\end{bmatrix}
\end{equation}

The damping matrix $\boldsymbol{D}_H(\boldsymbol{V}_j^S)$ is arranged as:
\begin{equation}
\boldsymbol{D}_H(\boldsymbol{V}_j^S) =-\mathrm{diag}\left\{ X_u, Y_v, Z_w, K_p, M_q, N_r     \right\}
\end{equation}
where $X_u, Y_v, Z_w, K_p, M_q, N_r$ are damping coefficients.
\subsection{Dynamic model of the flexible cable}
As the core component of flexibly connected dual-AUV system, the dynamic model of flexible cable is of great importance in the model formulation process. The force analysis schematic of the $i$-th mass point is shown in Figure \ref{CoordinateSystems} (c), where $\boldsymbol{F_i}$ represents the resultant force acting on the $i$-th mass point, formulated as:
\begin{equation}
  \begin{aligned}
    \boldsymbol{F}_i &= \boldsymbol{M}_i \boldsymbol{\ddot{P}}_i^E \\
    \boldsymbol{M}_i &= m \boldsymbol{I}_{3 \times 3} + \boldsymbol{m}_{a,i}
  \end{aligned}, \qquad i = 1,2,3,\dots, N_c
  \label{F_i_M_i}
\end{equation}
where $\boldsymbol{M}_i $ is the mass matrix of  $i$-th mass point, including inertia mass $m_i$ and added mass $\boldsymbol{m}_{a,i}$. These two components are respectively denoted as:
\begin{equation}
    \begin{aligned}
m_i &= \frac{1}{2}(\rho_c \sigma l_i + \rho_c \sigma l_{i+1})\\
\boldsymbol{m}_{a,i} &= \frac{1}{2}(\rho_c \sigma l_i k_a \boldsymbol{I}_{3 \times 3} + \rho_c \sigma l_{i+1} k_a \boldsymbol{I}_{3 \times 3})\\
  \end{aligned}, \qquad i = 1,2,3,\dots, N_c
\end{equation}
where $\rho_c$ is the density of the flexible cable, $\rho$ is the sea-water density, $\sigma$ is the cross sectional area of a single spring segment, $k_a$ is the added mass coefficient of the cable, $\boldsymbol{I_{3 \times 3}}$ represents the $3 \times 3$ identity matrix. $\ddot{P}_i^E = (\ddot{X}_i, \ddot{Y}_i, \ddot{Z}_i)$ is the acceleration vector of the $i$-th mass point. According to the Figure \ref{CoordinateSystems} (b), the dynamic  model of the $i$-th mass point in the inertial coordinate system is given by:
\begin{equation}
\boldsymbol{F}_i = \boldsymbol{B}_i^c + \boldsymbol{D}_{s,i} + \Delta \boldsymbol{T}_i + \boldsymbol{F}_{bend,i}, \qquad i = 1,2,3,\dots, N_c
\label{F_i}
\end{equation}
where $\boldsymbol{B}_i^c$ is the net buoyancy, $\boldsymbol{D}_{s,i}$ is the hydrodynamic forces, $\Delta \boldsymbol{T}_i$ is the tension from adjacent massless spring segments exerting on the $i$-th mass point. Assuming the strain satisfies $|\epsilon| \ll 1$, the tension vector of flexible cable is formulated based on the Hooke's law, in combination with the rotation matrix. For the the $i$-th mass point, $\Delta\boldsymbol{T}_i$  is expressed as:
\begin{equation}
\Delta\boldsymbol{T}_i = \boldsymbol{R}_E^i E \sigma \epsilon_i \boldsymbol{\tau}_i - \boldsymbol{R}_E^{i+1} E \sigma \epsilon_{i+1} \boldsymbol{\tau}_{i+1}, \qquad i = 1,2,3,\dots, N_c
\end{equation}
where $E$ is the Young's modulus, $\boldsymbol{\tau_i}$ is the unit tangent vector along the direction of the $i$-th spring segment, $\epsilon_i$ is the strain of the flexible cable, the equation of which is presented as follows:
\begin{equation}
	{\epsilon_i}=\left\{ \begin{array}{*{35}{l}}
   \dfrac{l_i}{l_0}-1 \hspace{3.7em} l_i >	l_0 \\[15pt]
   0 \hspace{5.7em} l_i <	l_0\\ 
\end{array} \right.	
\end{equation}
where $l_0$ is the original length of the spring segment. The net bouyancy of  $i$-th mass point is:
\begin{equation}
\boldsymbol{B}_i^c = (\rho_c - \rho) m_i \boldsymbol{g}, \qquad i = 1,2,3,\dots, N_c
\end{equation}
where $\boldsymbol{g}$ is the acceleration of gravity. In the calculation of hydrodynamic forces, the drag forces of flexible cable are decomposed into normal and tangential components, which are defined as $\boldsymbol{D}_{s,i}^n$ and $\boldsymbol{D}_{s,i}^t$. The drag force of $i$-th massless spring model in the local coordinate system is expressed as:
\begin{equation}
\begin{aligned}
\boldsymbol{D}_{s,i} =& \boldsymbol{D}_{s,i}^n + \boldsymbol{D}_{s,i}^t \approx
-\frac{1}{2} \rho \sqrt{1+\varepsilon} C_n l_i d' \left| \left( \dot{\boldsymbol{P}}_i^E - \boldsymbol{J}_i \right) \boldsymbol{\tau}_i^n \right| \cdot \left( \dot{\boldsymbol{P}}_i^E - \boldsymbol{J}_i \right) \boldsymbol{\tau}_i^n \\
&- \frac{1}{2} \rho \pi \sqrt{1+\varepsilon} C_t l_i d' \left| \left( \dot{\boldsymbol{P}}_i^E - \boldsymbol{J}_i \right) \boldsymbol{\tau}_i^t \right| \cdot \left( \dot{\boldsymbol{P}}_i^E - \boldsymbol{J}_i \right) \boldsymbol{\tau}_i^t,  \quad i = 1,2,3,\dots,N_c
\end{aligned}
\end{equation}
where $\dot{\boldsymbol{P}}_i^E$ is the velocity vector of the $i$-th mass point, $\boldsymbol{J}_i$ is the ocean current velocity  defined within the $S_i^C$ system of the flexible cable, $d^{'}$ is the cable diameter, $C_n$ and $C_t$ are normal and tangential drag coefficients, which are treated as constants.

The bending force $\boldsymbol{F}_{bend,i}$  describes the restoring force of the flexible cable induced by the material elastic stiffness. \\[2pt] Under the small-deformation assumption, $\boldsymbol{F}_{bend,i}$ is approximated using the finite difference method. For the internal \\[2pt] mass points ($i$ = 1, 2, 3, \dots, $N_c-1$), $\boldsymbol{F}_{bend,i}$ is formulated as:
\begin{equation}
\boldsymbol{F}_{bend,i} = -\frac{EI}{l_0^3}\left( \boldsymbol{P}_{i+1}^E - 2\boldsymbol{P}_{i}^E + \boldsymbol{P}_{i-1}^E \right)
\end{equation}
where $EI$ is the bending stiffness, $l_0$ is the original length of the segment, $\boldsymbol{P}_{i}^E $ is the position of the $i$-th mass point in the Earth-fixed coordinate system. To improve the approximation accuracy of boundary conditions, the first-order modified finite difference scheme is applied at the connection points, which is presented in the following equations. For the left boundary point ($i$ = 1):
\begin{equation}
\boldsymbol{F}_{bend,1} = -\frac{EI}{l_0^3}\left( 2\boldsymbol{P}_{1}^E - 5\boldsymbol{P}_{2}^E + 4\boldsymbol{P}_{3}^E - \boldsymbol{P}_{4}^E \right)
\end{equation}
For the right boundary point ($i$ = $N_c$):
\begin{equation}
\boldsymbol{F}_{bend,N_c} = -\frac{EI}{l_0^3}\left( 2\boldsymbol{P}_{N_c}^E - 5\boldsymbol{P}_{N_c-1}^E + 4\boldsymbol{P}_{N_c-2}^E - \boldsymbol{P}_{N_c-3}^E \right)
\end{equation}

The boundary conditions are established based on the fixed connection between both endpoints of the flexible cable and the AUV, implying that the positions and velocities of the cable endpoints are identical to the corresponding AUVs. The mathematical equations of the boundary conditions are expressed as:

\begin{equation}
\begin{aligned}
\boldsymbol{P}_i        &= \boldsymbol{\eta}_{i,j}^E \hspace{3.7em} (i = 1, j=1) \\
\boldsymbol{P}_i        &= \boldsymbol{\eta}_{i,j}^E \hspace{3.7em} (i = N_c, j=2) \\
\boldsymbol{\dot{P}}_i  &= \boldsymbol{\dot{\eta}}_{i,j}^E \hspace{3.7em} (i = 1, j=1) \\
\boldsymbol{\dot{P}}_i  &= \boldsymbol{\dot{\eta}}_{i,j}^E \hspace{3.7em} (i = N_c, j=2) \\
\boldsymbol{T}_i        &= \boldsymbol{T}_{AUV_j} \hspace{2.6em} (i = 1, j=1) \\
\boldsymbol{T}_i        &= \boldsymbol{T}_{AUV_j} \hspace{2.6em} (i = N_c, j=2)
\end{aligned}
\label{boundaryconditions}
\end{equation}

Integrating the elements of the equations \ref{F_i_M_i} and  \ref{F_i} with the boundary conditions in equation \ref{boundaryconditions}, a comprehensive mathematical model for the flexibly connected dual-AUV system is formulated in this research.

\subsection{Parameter identification method for underwater flexible cables}
In the flexibly connected dual-AUV system, in order to accurately capture the physical behaviors of the cable, system identification of critical parameters for the flexible cable is conducted. The unknown parameters to be identified mainly include the cable diameter, added mass coefficients of segments, material density,  equivalent Young's modulus, tangential and normal drag coefficients. Some of the parameters  are directly measured through experiments: the diameter is measured by using a vernier caliper, and the density  is obtained  through weighing and volume calculation. However, other parameters, such as the equivalent Young's modulus and hydrodynamic related parameters, are not directly accessible through standard experimental methods.

In this research, a series of water tank experiments involving multiple operational conditions is designed and conducted to identify these unknown parameters. During the identification experiments, the AUVs are controlled to perform uniform linear motions under different separation distances, thereby generating multiple sets of experimental samples. The three-dimensional position information of the AUV is collected by the positioning system, while the heading angle is derived from multi-sensor data fusion, thereby enabling the precise localization of both cable endpoints. In conjunction with the 3-DoF force sensors integrated into  the AUV platforms, the actual cable force data for each operating conditions are collected.

To identify the critical parameters, this paper proposes the objective function targeting for minimizing the deviation between the predictions and measurements at both endpoints. The objective function is shown as:
\begin{equation}
\mathcal{L}_{cable} = \dfrac{1}{N_{cable}} \sum_{k=1}^{N_{cable}}\left| \left| \boldsymbol{F}_{Real}(k) - \boldsymbol{F}_{Predicted}(X_d(k), Y_d(k)) \right|\right|^2
\end{equation}
where $\mathcal{L}_{cable}$ is the objective function value, $N_{cable}$ is the total number of all the involved samples, $k$ is the sample number, $\boldsymbol{F}_{Real}(k)$ is the force measurements of the $k$-th sample, $\boldsymbol{F}_{Predicted}(X_d(k), Y_d(k))$ is the prediction results based on the physical model, $X_d(k)$ and $Y_d(k)$ are the lateral and longitudinal relative position information between the two AUVs in the $k$-th sample. Serving as the fitness evaluation function of the genetic algorithm, this objective function drives the iterative update of model parameters by minimizing the discrepancy between experimental measurements and model predictions. Consequently, the joint identification of critical physical parameters, including the equivalent Young's modulus and hydrodynamic drag coefficients, is systematically  achieved.

\section{Water Tank Experiments and Parameter Identification}
\subsection{Experimental system design}
As shown in Figure \ref{CoordinateSystems}, the experimental system is composed of two identical AUVs interconnected by a flexible cable. The flexible cable in this system serves as a representation of the underwater cables applied in the practical  sensor equipment. The cable is composed of multiple constituent materials, forming a composite structure: the high-strength Kevlar woven material is utilized in the outer protective sheath to enhance the tensile strength and wear resistance, while the copper conductor is placed inside to facilitate data transmission and power supply. For the avoidance of bouyancy interference affecting the attitude and dynamics of the system, the cable is designed to achieve  neutral buoyancy and is connected to the the middle section of the AUVs.
\begin{center}
    \includegraphics[width=9.5cm]{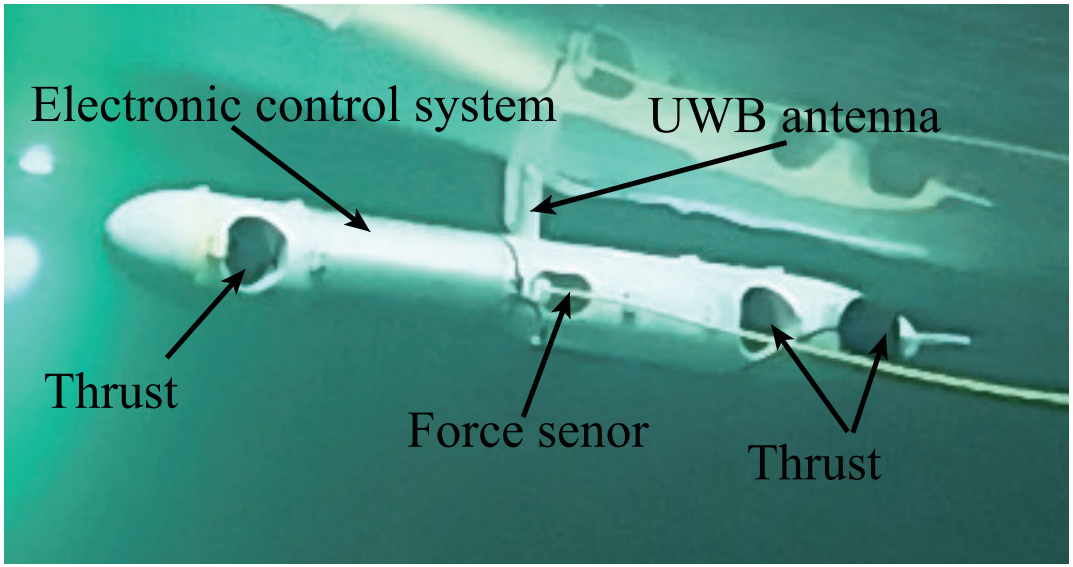}
    \captionof{figure}{Physical view of a single AUV} 
    \label{single_AUV}
\end{center}

The structure of a single AUV is shown in Figure \ref{single_AUV}. The AUV has an overall length of 1049 mm, a diameter of 160 mm, and a total mass of 13.17 kg. During the experiments, The two AUV platforms are adjusted to an approximately neutrally buoyant state to ensure attitude stability. The AUV is designed with four modular sections, including the bow section, controller section, tension sensor section, and the stern section. The propulsion system is composed of a main thruster and two lateral thrusters, which enables the AUV to perform forward motions and yaw control. Since the platform is intended for principle verification, vertical thrusters are not equipped for system simplicity. 3-DoF force sensors are integrated in the tension sensor section to provide   real-time force measurements  at the endpoints. UWB indoor positioning modules and IMU attitude sensors are integrated in the electrical control system  to record the motion data of the dual-AUV system. The prototype platform of flexibly connected dual-AUVs is shown in Figure \ref{Prototype_Platform}, and the experimental environment is shown in Figure \ref{experimental_environments}.

 \begin{center}
    \includegraphics[width=9.5cm]{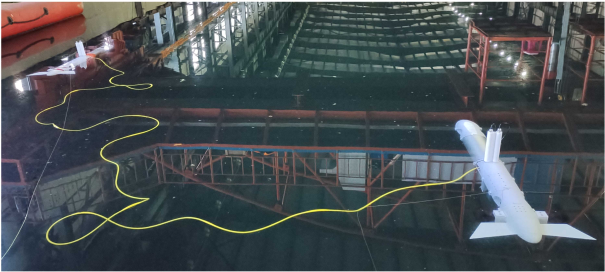}
    \captionof{figure}{Prototype platform of the flexibly connected dual-AUV system} 
    \label{Prototype_Platform}
\end{center}

\begin{center}
    \includegraphics[width=9.5cm]{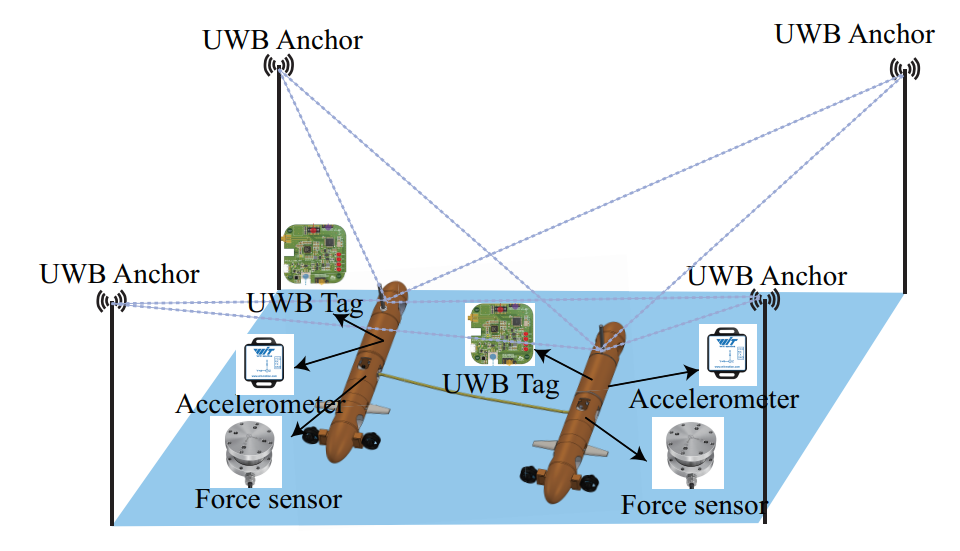}
    \captionof{figure}{Sensor and positioning architecture of the AUV system} 
    \label{experimental_environments}
\end{center}

\subsection{Parameter identification of the flexible cable}
During the identification experiments, a PID controller is implemented to stabilize the heading and X-direction positions. Motion control and dynamic response tests are carried out in the water tank environment. The two AUVs are controlled to perform uniform motion at a cruising speed of 1 knot. During this process, the tension and position data are recorded once the cable reaches a steady state. By varying the lateral and longitudinal separation between the two AUV platforms, a dataset containing multiple operating conditions is constructed. A representative set of collected data is shown in Figure \ref{Curves}, where Figure \ref{Curves} (a) and (c) present the tension variation process, while Figure \ref{Curves} (b) and (d) depicts the corresponding motion trajectory. As indicated by the experimental results, the X-direction positions of both AUVs remain stable while the motion is mainly along the Y-direction. These results demonstrates that the PID controller effectively suppresses lateral deviations, ensuring consistent boundary conditions throughout the experimental process.

According to the force measurements, the tension is negligible during the initial thruster startup stage. As the thrust gradually increases, the platforms accelerate and eventually reach a steady state, and correspondingly, the cable tension increases and approaches a stable level. When the thrusters are shut down, the whole system continues to move forward for a distance. The velocity and cable tension exhibit a time-dependent decay to zero. The tension  exhibits a continuous transition through the acceleration, steady-state, and deceleration stages, which is consistent with the dynamic characteristics of the underwater multibody system. Excellent continuity and physical consistency are demonstrated in the correspondence between the force measurements and the position data, which verifies the stability and reliability of the experiments. Therefore, the constructed dataset is utilized for subsequent dynamic model validation and parameter identification.

\captionsetup{justification=centering}
% 定义高度
\newlength{\imgHeight}
\setlength{\imgHeight}{3.5cm}

\begin{center}
    \begin{minipage}{0.48\textwidth}
        \centering
        \includegraphics[height=\imgHeight,keepaspectratio]{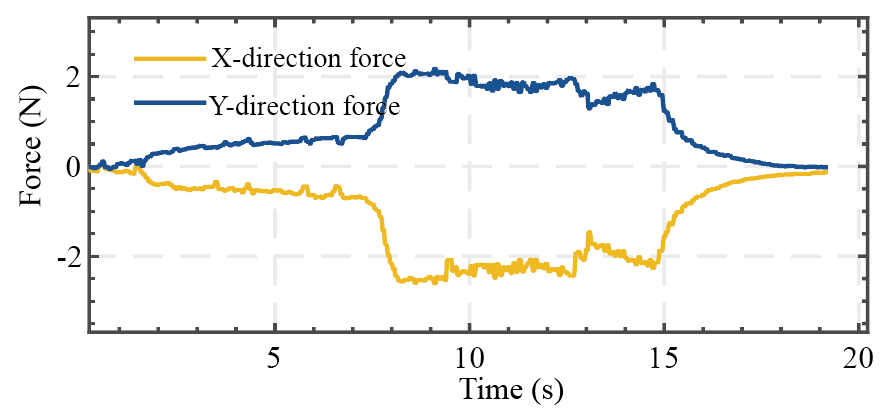}
        \vspace{0.5em}
        \text{(a) Force measurements of $AUV_1$} % 自定义标题样式
    \end{minipage}
    \hfill
    \begin{minipage}{0.48\textwidth}
        \centering
        \includegraphics[height=\imgHeight,keepaspectratio]{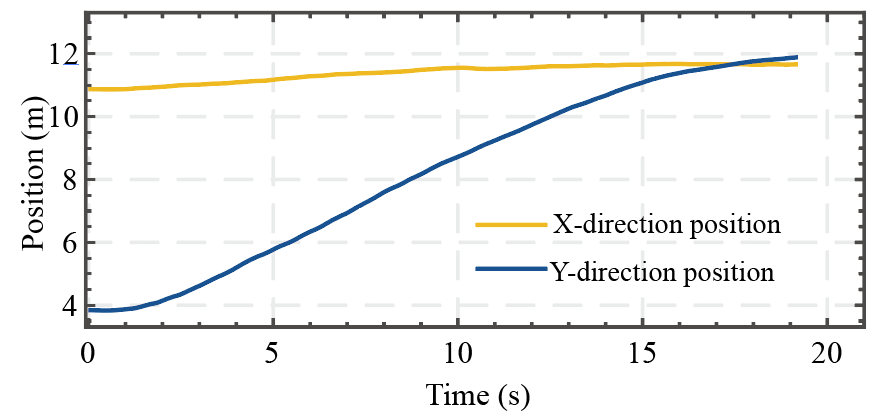}
        \vspace{0.5em}
        \text{(b) Position data of $AUV_1$} % 自定义标题样式
    \end{minipage}
    
   \begin{minipage}{0.48\textwidth}
        \centering
        \includegraphics[height=\imgHeight,keepaspectratio]{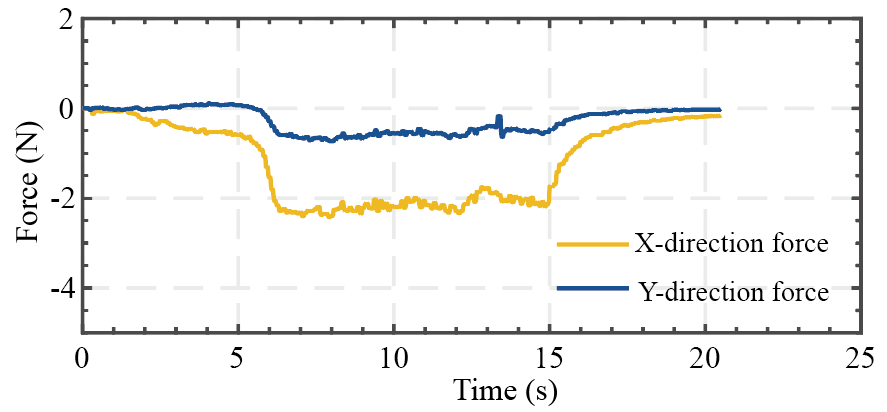}
        \vspace{0.5em}
        \text{(c) Force measurements of $AUV_2$} % 自定义标题样式
    \end{minipage}
    \hfill
    \begin{minipage}{0.48\textwidth}
        \centering
        \includegraphics[height=\imgHeight,keepaspectratio]{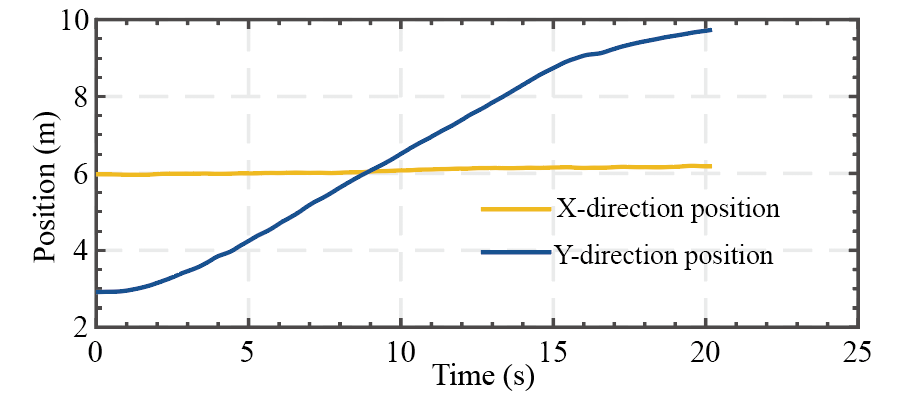}
        \vspace{0.5em}
        \text{(d) Position data of $AUV_2$} % 自定义标题样式
    \end{minipage}

    \captionof{figure}{Position and force variation curves during the experiment of \\ the flexibly connected dual-AUV system}
    \label{Curves}
\end{center}

Photographs of the experimental process are presented in Figure \ref{experim_photo}, which demonstrate the collaborative experiments of the flexibly connected dual-AUV system for steady-state cable tension data acquisition. Based on the force data obtained at multiple  separation distances with a forward speed of 1 knot, the Genetic Algorithm is implemented for  parameter identification. The identification results and hyperparameters of the cable dynamics model are listed in Table \ref{Hydrodynamic_parameters_and_hyperparameters}. In further analysis, numerical calculations of the cable tension under various operating conditions are performed using the established model. The comparison between the numerical results and experimental measurements is presented in Figure \ref{comparison}, where the scatter points represent the tension measurements from the experiments, and the continuous surface is the theoretical force predictions. The overall results indicate that the cable tension exhibits a nonlinear growth trend with the increase in the separation distance between the AUV platforms. In addition, a high degree of consistency is observed between the theoretical calculations and the experimental measurements.

\begin{center}
\captionof{table}{Hydrodynamic parameters and hyperparameters from calculation process}
    \footnotesize
    \renewcommand{\arraystretch}{1.5} % 增加50%的行高
    \begin{tabularx}{\textwidth}{@{}*{10}{>{\centering\arraybackslash}X}@{}} % 添加居中对齐
    \toprule
    Parameters & $N_c$ & $\rho_c$ & $\rho$ & $\sigma$ & $k_a$ & $E$ & $C_n$ & $C_t$ & $d'$ \\
    \midrule
    value & 30 & 1025\,$\text{kg}/\text{m}^3$ & 1025\,$\text{kg}/\text{m}^3$ & 0.00384\,$\text{m}^2$ & 1.3 & 3.68\,$\text{MPa}$ & 1.8306 & 0.0756 & 0.07\,$\text{m}$ \\
    \bottomrule
    \end{tabularx}
    \label{Hydrodynamic_parameters_and_hyperparameters}
\end{center}

\begin{center}
    \includegraphics[width=16cm]{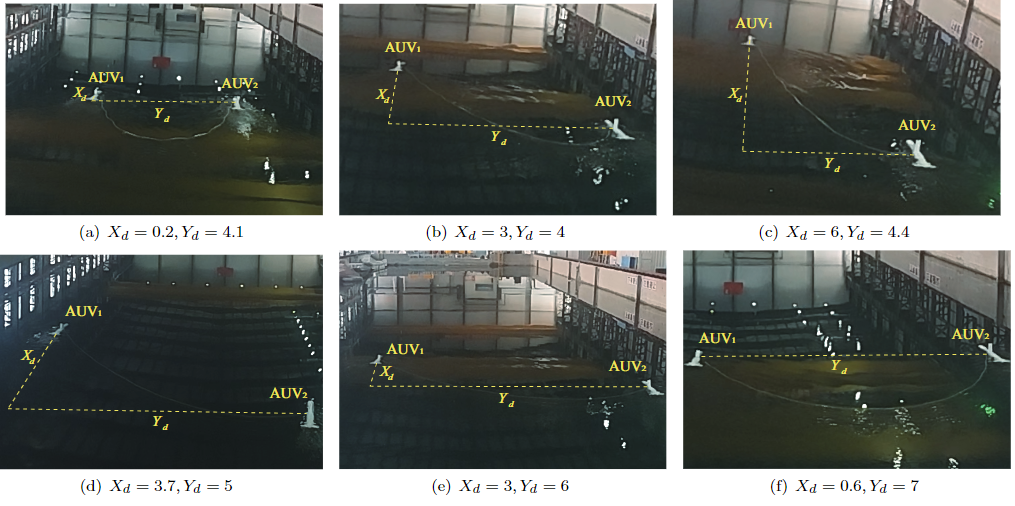}
    \captionof{figure}{Photographs of the flexibly connected dual-AUV system in experiments} 
    \label{experim_photo}
\end{center}

The  distribution of deviation between the theoretical predictions and experimental measurements is shwon in Figure \ref{error_distribution}. The maximum of average tension deviation for all separation conditicons is lower than 0.7 N, which verifies the accuracy and stability of  the proposed identification method. The RMSE and MSE indicators under different operational conditicons are summarized in Table \ref{RMSE_and_MSE}. According to the table, the deviation levels between the force predictions and measurements are consistently low for all operating conditions. These results collectively validate the applicability and accuracy of the dynamic model and numerical solution methods. Therefore, the methodology developed in this research is implementable for tension prediction and analysis of AUV boundary loads in similar towing systems.

\newlength{\imgHeightcomparison}
\setlength{\imgHeightcomparison}{4.5cm}

\begin{center}
    \begin{minipage}{0.48\textwidth}
        \centering
        \includegraphics[height=\imgHeightcomparison,keepaspectratio]{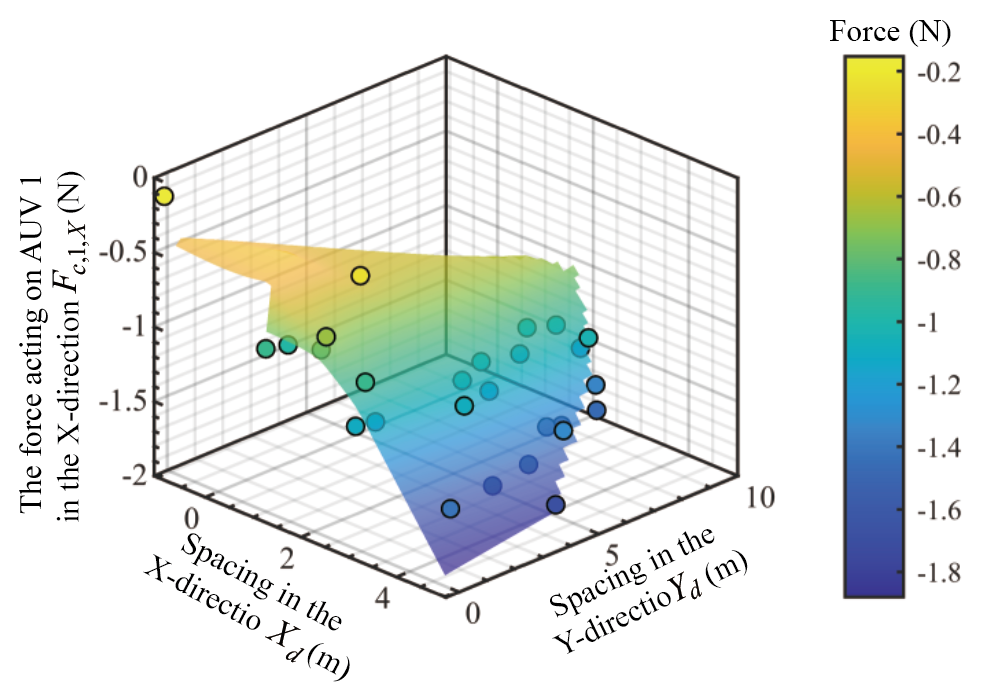}
        \vspace{0.5em}
        \text{\makecell{(a)  Comparison of experimental and theoretical forces\\ $F_{c,1,X}$ in X-direction for $AUV_{1}$}}

    \end{minipage}
    \hfill
    \begin{minipage}{0.48\textwidth}
        \centering
        \includegraphics[height=\imgHeightcomparison,keepaspectratio]{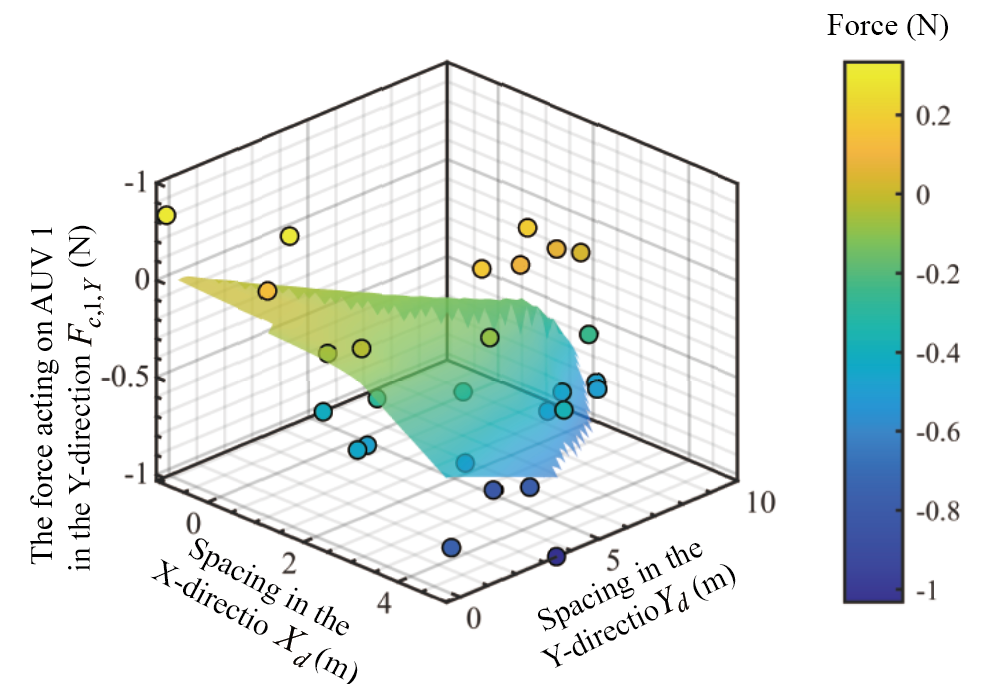}
        \vspace{0.5em}
        \text{\makecell{(b)  Comparison of experimental and theoretical forces\\ $F_{c,1,Y}$ in Y-direction for $AUV_{1}$}}
    \end{minipage}
    
   \begin{minipage}{0.48\textwidth}
        \centering
        \includegraphics[height=\imgHeightcomparison,keepaspectratio]{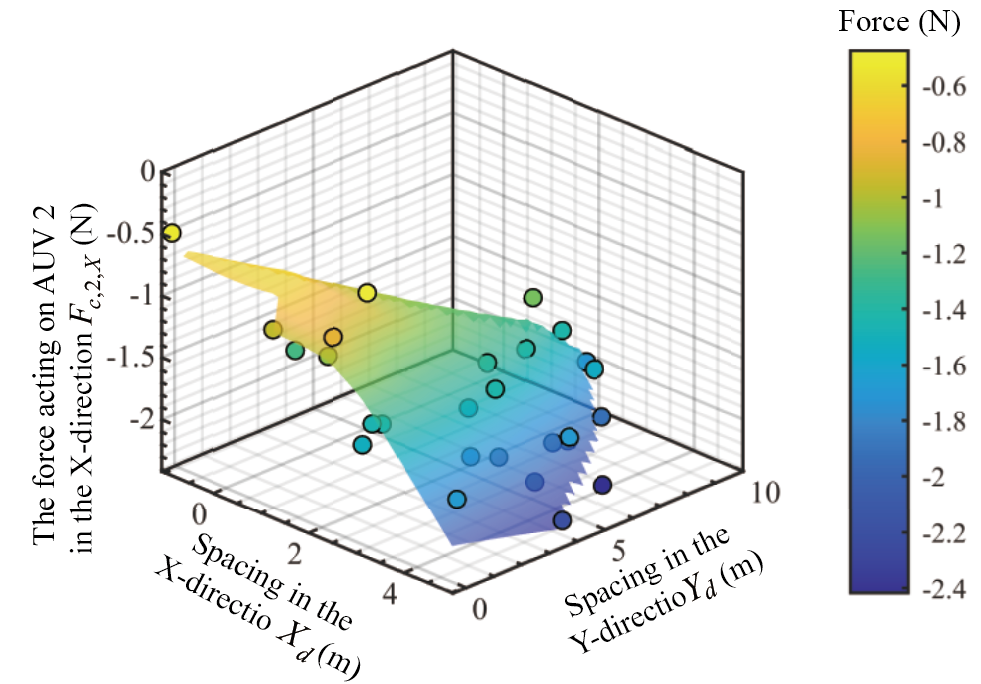}
        \vspace{0.5em}
        \text{\makecell{(c)  Comparison of experimental and theoretical forces\\ $F_{c,2,X}$ in X-direction for $AUV_{2}$}}
    \end{minipage}
    \hfill
    \begin{minipage}{0.48\textwidth}
        \centering
        \includegraphics[height=\imgHeightcomparison,keepaspectratio]{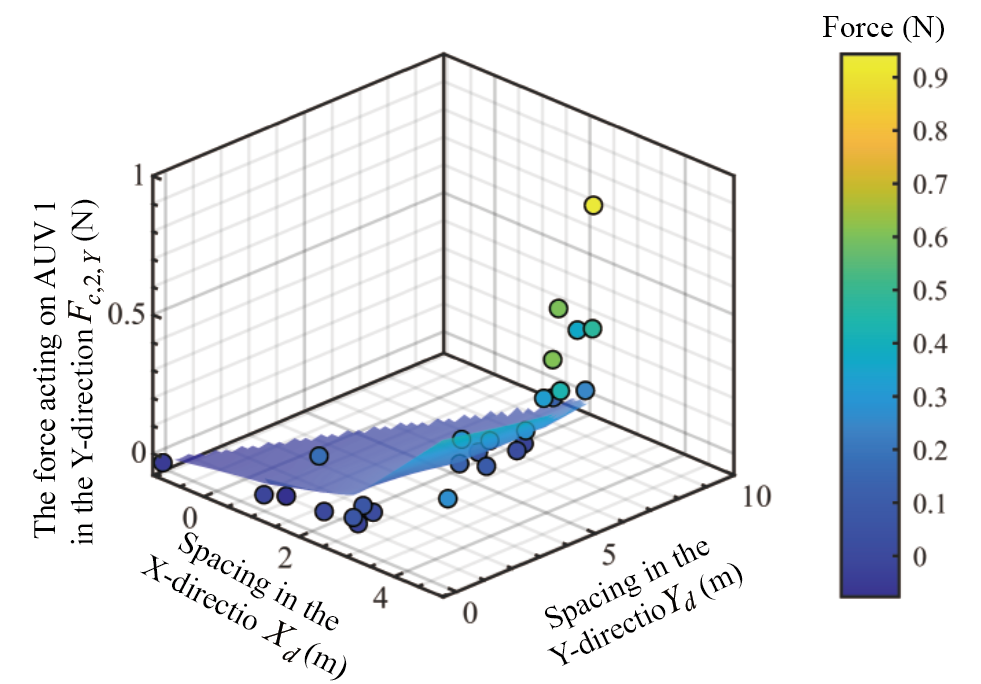}
        \vspace{0.5em}
      \text{\makecell{(b)  Comparison of experimental and theoretical forces\\ $F_{c,2,Y}$ in Y-direction for $AUV_{2}$}}
    \end{minipage}

    \captionof{figure}{Comparison of the experimental and theoretical  forces for the flexible connected dual-AUV system} 
    \label{comparison}
\end{center}

\begin{center}
    \includegraphics[width=11cm]{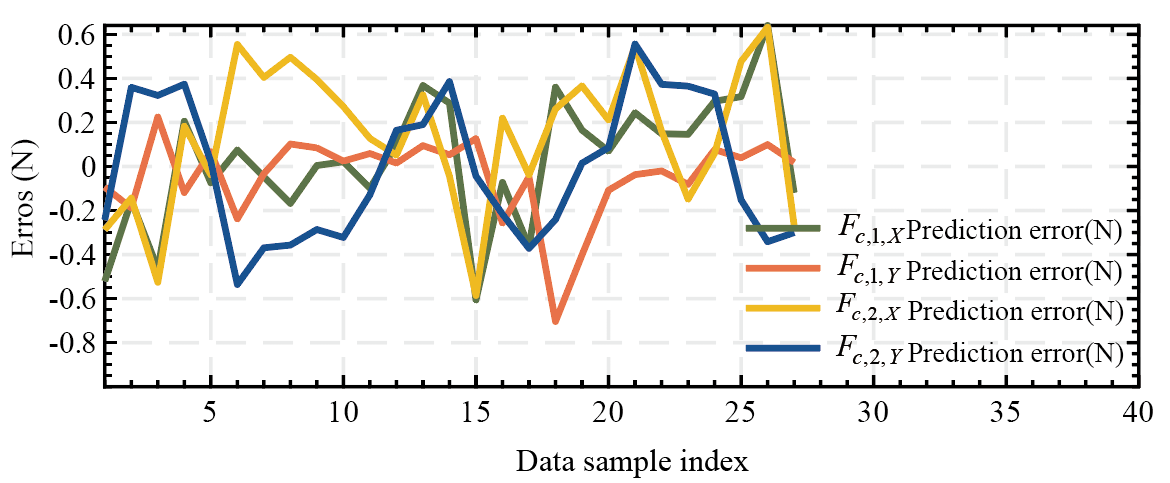}
    \captionof{figure}{Comparison of deviations between theoretical and experimental results of \\ the flexibly connected dual-AUV system} 
    \label{error_distribution}
\end{center}

\begin{center}
    \captionof{table}{Summary of RMSE and MSE  indicators}
    \footnotesize
    \renewcommand{\arraystretch}{1.5} % 增加50%的行高
    \begin{tabularx}{0.9\textwidth}{@{}*{3}{>{\centering\arraybackslash}X}@{}} % 添加居中对齐
    \toprule
    Indicator Definition                & RMSE(N)  & MSE(N$^2$)  \\ 
    \midrule
    Cable tension on $AUV_{1}$ (Y-direction) & 0.286327 & 0.081983 \\ 
    Cable tension on $AUV_{1}$ (Y-direction) & 0.190809 & 0.036408 \\ 
    Cable tension on $AUV_{1}$ (Y-direction) & 0.342861 & 0.117554 \\ 
    Cable tension on $AUV_{1}$ (Y-direction) & 0.308454 & 0.095144 \\     \bottomrule
    \end{tabularx}
    \label{RMSE_and_MSE}
\end{center}

\section{Simulation Experiments}
\subsection{Convergence analysis of the spatial and temporal discretization settings}
\subsubsection{Convergence analysis of the spatial discretization settings}

To verify the numerical convergence of the proposed algorithm as the number of discrete segments increases, and to further evaluate the influence on  cable modeling accuracy and computational stability, this section presents a series of systematic simulation investigations of the physical model established for the flexibly connected dual-AUV system. In the simulation analysis, the velocities of AUV platforms are set to 1 m/s, while the separation distance is set to 5 m. By setting the number of discretized segments to $n$ = 5, 10, 20, 30, 40, 50, 60 and 70, multiple groups of simulations are conducted, thereby analyzing the steady-state cable configuration, the endpoint tension, and the simulation runtime.

\begin{center}
    \includegraphics[width=7cm]{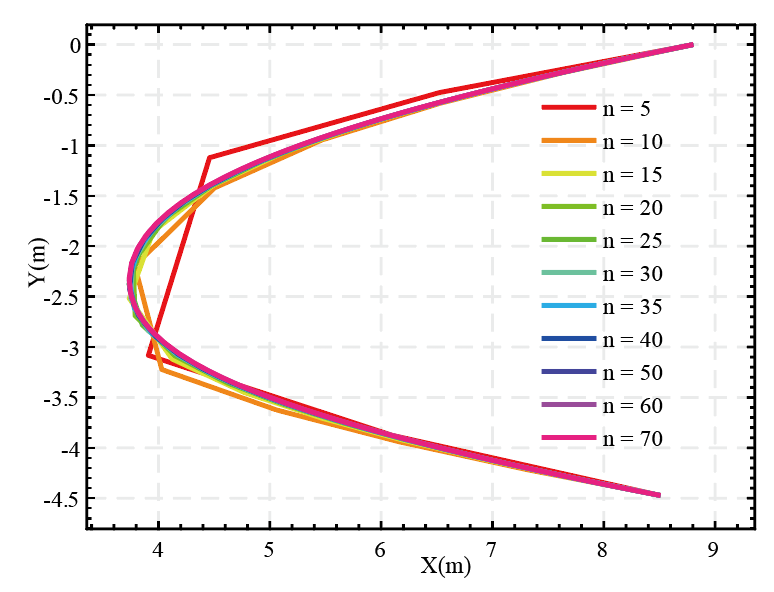}
    \captionof{figure}{Steady-state cable shape under different numbers of discrete segments} 
    \label{Steady_State_Cable_Shape}
\end{center}

Figure \ref{Steady_State_Cable_Shape} illustrates the steady-state bending configurations of the flexible cable in the horizontal plane under different spatial discretization levels. According to the simulation results, as $n$ increases, the system response exhibits a trend of convergence, and the cable configuration transitions from a distinct polyline to a smooth, continuous curve. To further quantify the geometric formation convergence, the midpoint deviations relative to the reference solution ($n = 70$) for multiple spatial discretization settings are computed and listed in Table \ref{Comparison_of_cable_shape_deviations_and_computation_time}. Additionally, the associated average deviations, maximum deviations, and simulation runtimes  are also presented. The results indicate that the geometric deviation of the cable decreases significantly as the number of segments increases from 10 to 30. For $n \geq 30$, all deviation indicators converge to an acceptable range, with the average deviation less than 0.01 m and the maximum deviation becomes stable. The improvement in shape prediction accuracy is limited, whereas the computational time rises rapidly.

\begin{center}
    \captionof{table}{Comparison of cable shape deviations and computational time for different spatial discretization levels (Reference: n=70)}
    \footnotesize
    \renewcommand{\arraystretch}{1.5} % 增加50%的行高
    \begin{tabularx}{1.0\textwidth}{@{}*{5}{>{\centering\arraybackslash}X}@{}} % 添加居中对齐
    \toprule
        Number of Segments  $n$            & Midpoint deviations (m) & Average deviations (m) & Max deviations (m) & Simulation runtime (s)\\ 
    \midrule
          5	  & 0.394	& 0.109	& 0.394	& 3  \\ 
          10	& 0.085	& 0.062	& 0.207	& 7  \\ 
          15	& 0.094	& 0.022	& 0.094	& 11 \\ 
          20	& 0.030	& 0.013	& 0.054	& 13 \\ 
          25	& 0.042	& 0.011	& 0.042	& 17 \\ 
          30	& 0.018	& 0.008	& 0.029	& 21 \\ 
          35	& 0.024	& 0.006	& 0.024	& 25 \\ 
          40	& 0.011	& 0.005	& 0.019	& 32 \\ 
          50	& 0.006	& 0.003	& 0.011	& 38 \\ 
          60	& 0.003	& 0.002	& 0.006	& 45 \\ 
          70	& 0.000	& 0.000	& 0.000	& 62 \\      \bottomrule
    \end{tabularx}
    \label{Comparison_of_cable_shape_deviations_and_computation_time}
\end{center}

To further evaluate the influence of the number of segments $n$ on the dynamic responses, the distribution of external forces at the endpoints in the dual-AUV system is illustrated in Figure \ref{Force_Distribution}.The corresponding relative deviations between the force components and reference solutions are listed in Table \ref{Relative_deviation_of_AUV_tension_response}. According to the results,  as $n$ increases from 5 to 60, the cable tensions  tend to stabilize in all directions and the system responses exhibit excellent convergence characteristics. Especially when $n \geq 30$, the deviations of all tension components are significantly converged. The deviation remains below 0.3 N, which indicates that the sensitivity of tension responses to spatial discretization settings is  significantly reduced.

\begin{center}
    \includegraphics[width=7cm]{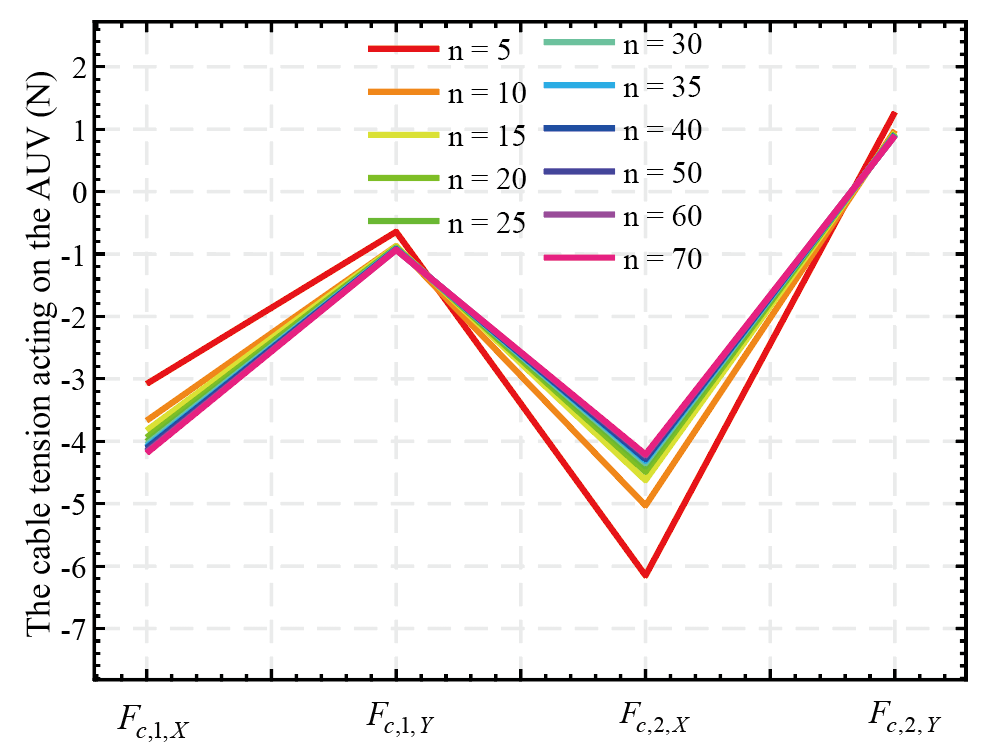}
    \captionof{figure}{Variation of AUV Force Distribution under Different Discretization Levels in a \\ Flexibly Connected Dual-AUV System} 
    \label{Force_Distribution}
\end{center}

\begin{center}
    \captionof{table}{Relative deviation of AUV tension response under different discretization levels in the \\ flexibly connected dual-AUV system (Reference: n=70)}
    \footnotesize
    \renewcommand{\arraystretch}{1.5} % 增加50%的行高
    \begin{tabularx}{1.0\textwidth}{@{}*{6}{>{\centering\arraybackslash}X}@{}} % 添加居中对齐
    \toprule
        Number of Segments $n$ & Deviations of $F_{c,1,Y}$ (N) &Deviations of $F_{c,j,X}$ (N) & Deviations of $F_{c,2,X}$ (N) & Deviations of $F_{c,2,Y}$ (N) & Average deviations (N) \\ 
    \midrule
        10	&1.111	&0.291	&1.942	&0.379	&0.931\\ 
        20	&0.522	&0.063	&0.821	&0.087	&0.373\\ 
        30	&0.366	&0.057	&0.415	&0.048	&0.221\\ 
        40	&0.254	&0.039	&0.278	&0.037	&0.152\\ 
        50	&0.186	&0.028	&0.195	&0.025	&0.108\\ 
        60	&0.141	&0.021	&0.143	&0.019	&0.081\\ 
        70	&0.106	&0.015	&0.104	&0.013	&0.060\\      \bottomrule
    \end{tabularx}
    \label{Relative_deviation_of_AUV_tension_response}
\end{center}

\subsubsection{Convergence analysis of the temporal discretization settings}

To evaluate the impact of temporal discretization on the modeling results of the flexibly connected dual-AUV system, a time step independence study is conducted in this section. The operating condition of the dual-AUV system is set to a constant speed of 1 m/s with a separation distance of 5 m. A series of simulations with different time step sizes ($t_s=10^{-1}, 10^{-2}, 10^{-3}, 10^{-4}, 10^{-5}$ s) is conducted to investigate the influence of temporal discretization accuracy on the steady-state cable configurations and endpoint force responses, thereby verifying the temporal integration convergence and selecting an appropriate time step for subsequent calculations. 

Figure \ref{Varying_Time_Discretizations} presents a comparison of the steady-state cable configurations obtained from simulations with different time step settings. According to the results, with the exception of the $t_s = 0.1$ s simulation, the differences among other cable configurations  are negligible. The overall curve profiles are highly consistent, indicating that the sensitivity of configurations to discrete time step settings is relatively low. Table \ref{Comparison_of_Cable_Shape} further provides a quantitative comparison of the midpoint, average, and maximum deviations of the cable under different time step settings, relative to the reference solution obtained using the minimum time step of $t_s = 10^{-5}$ s. The results indicate that the model is unable to produce valid solutions when the time step is set to 0.1 s. Once the time step is reduced to less than $10^{-2}$ s, the geometric deviations decrease rapidly and converge to the magnitude of $10^{-5}$ m when $t_s \leq 10^{-4}$ s, demonstrating excellent time step independence in the predictions of cable configurations.

The results in Figure \ref{Different_Discretization_Levels} and Table \ref{RelativeErrorofTensionResponse} further demonstrate the deviations of cable tension components in the X and Y directions relative to the reference solution  of $t_s = 10^{-5}$ s. These numerical simulation results demonstrate that the maximum tension deviation exceeds 0.1 N when $t_s = 10^{-2}$ s. However, as the time step falls below $10^{-3}$ s, the deviations in all force components decrease rapidly, with the maximum deviation less than $10^{-3}$ N and the average deviation is further reduced to below $10^{-5}$ N. These results confirm the temporal convergence of the proposed model with respect to the force responses.

\begin{center}
    \includegraphics[width=7cm]{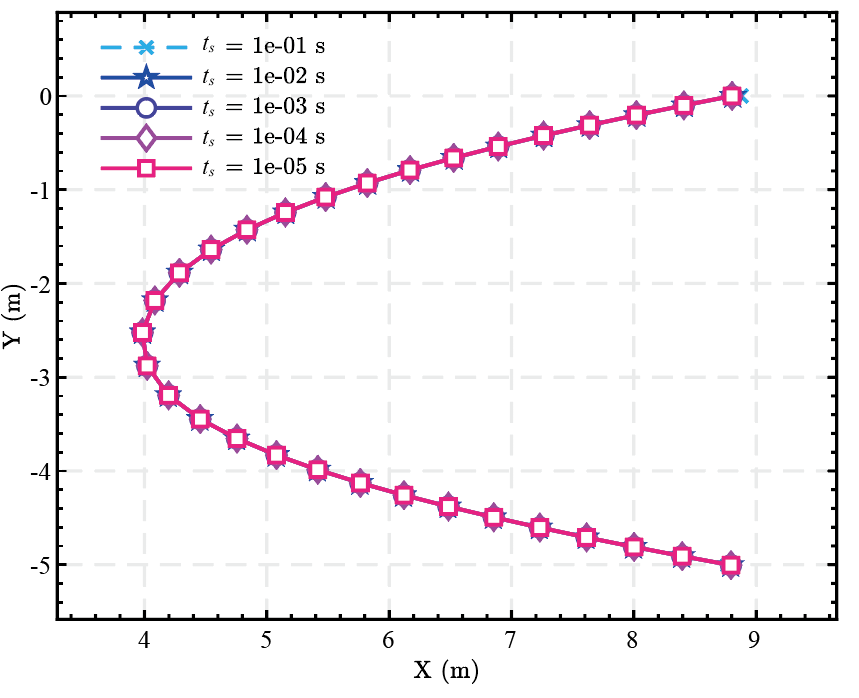}
    \captionof{figure}{Comparison of Cable Steady-State Profiles under Varying Time Discretizations} 
    \label{Varying_Time_Discretizations}
\end{center}

Based on the above analysis, the established model demonstrates convergence in both cable configurations and force responses, exhibiting excellent independence from the discrete time step settings. Considering the trade-off between computational efficiency and accuracy, $t_s = 10^{-3}$ s is selected as the standard discrete time step to facilitate high-efficiency simulations of the system.

\begin{center}
    \includegraphics[width=7cm]{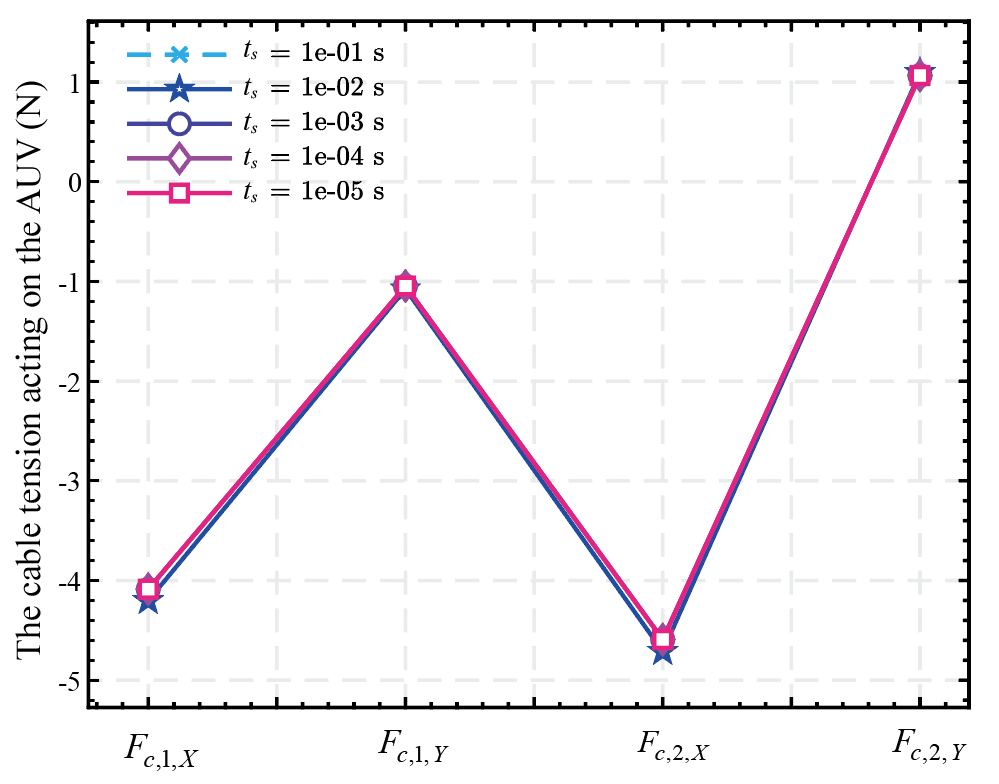}
    \captionof{figure}{Variation of cable tensions at both AUV endpoints in the X and Y directions \\ under different discretization levels} 
    \label{Different_Discretization_Levels}
\end{center}

\begin{center}
    \captionof{table}{Comparison of cable shape deviation and simulation runtime \\ under different time steps (Reference: $t_s$=10${^{-5}}$s)}
    \footnotesize
    \renewcommand{\arraystretch}{1.5} % 增加50%的行高
    \begin{tabularx}{1.0\textwidth}{@{}*{5}{>{\centering\arraybackslash}X}@{}} % 添加居中对齐
    \toprule
        Time-step  $t_s $ (s)           & Midpoint deviations (m) & Average deviations (m) & Max deviations (m) & Simulation runtime (s)\\ 
    \midrule
          $10^{-1}$	& NaN	  & NaN   & NaN	  & 5  \\ 
          $10^{-2}$	& 3.005 $\times$ $10^{-3}$	& 3.995 $\times$ $10^{-3}$	& 6.492 $\times$ $10^{-3}$	& 13  \\ 
          $10^{-3}$	& 1.640 $\times$ $10^{-4}$	& 2.170 $\times$ $10^{-4}$	& 2.650 $\times$ $10^{-4}$	& 21 \\ 
          $10^{-4}$	& 1.500 $\times$ $10^{-4}$	& 1.960 $\times$ $10^{-5}$	& 2.250 $\times$ $10^{-5}$	& 34 \\ 
          $10^{-5}$	& 0    	& 0   	& 0   	& 43 \\      \bottomrule
    \end{tabularx}
    \label{Comparison_of_Cable_Shape}
\end{center}

\newpage

\vspace*{-35pt} % 负间距

\begin{center}
    \captionof{table}{Relative Error of Tension Response in the Flexible Connection Dual-AUV System \\ under Different Time Steps (Reference: ts=10$^{-5}$ s)}
    \footnotesize
    \renewcommand{\arraystretch}{1.5} % 增加50%的行高
    \begin{tabularx}{1.0\textwidth}{@{}*{6}{>{\centering\arraybackslash}X}@{}} % 添加居中对齐
    \toprule
        Time-step  $t_s $ (s)  & Deviations of $F_{c,1,Y}$ (N) &Deviations of $F_{c,j,X}$ (N) & Deviations of $F_{c,2,X}$ (N) & Deviations of $F_{c,2,Y}$ (N) & Average deviations (N) \\ 
    \midrule
        $10^{-1}$	& NaN	& NaN	& NaN &	NaN	& NaN\\ 
        $10^{-2}$	& 1.157 $\times$ $10^{-1}$	& 2.888$\times$ $10^{-2}$	& 1.203 $\times$ $10^{-1}$	& 2.882 $\times$ $10^{-2}$	& 7.343 $\times$ $10^{-2}$ \\ 
        $10^{-3}$	&1.255 $\times$ $10^{-3}$	& 2.910 $\times$ $10^{-4}$	& 1.052  $\times$ $10^{-3}$	& 2.740 $\times$ $10^{-4}$	& 7.180 $\times$ $10^{-4}$ \\ 
        $10^{-4}$	&2.630 $\times$ $10^{-5}$	& 4.150 $\times$ $10^{-6}$	& 3.190 $\times$ $10^{-6}$	& 1.680 $\times$ $10^{-6}$	& 8.820 $\times$ $10^{-6}$ \\ 
        $10^{-5}$	&0	& 0	& 0	& 0	& 0 \\      \bottomrule
    \end{tabularx}
    \label{RelativeErrorofTensionResponse}
\end{center}

\subsection{Simulation analysis of endpoint tension responses and cable configurations under different material conditions.}

The cable material properties are among the most important factors affecting the dynamic behavior of the flexibly connected dual-AUV system. The stiffness and geometric characteristics directly determine the load distribution and formation configuration of the system in actual operations. To investigate the influence of cable material properties on the performance of the overall system, numerical simulations are conducted with the cable diameter $d$ and Young's modulus $E$ taken as variables. Similar to the previous sections,  the separation distance of the dual-AUV system is set to 5 m, while the velocities of the two AUV platforms are set to 1 m/s. The cable diameter and Young's modulus are set as $d \in$ [10$^{-3}$, 10$^{-1}$] m and $E \in$ [10$^{6}$, 10$^{10}$] Pa, respectively. Logarithmic sampling is adopted to evaluate the coupled influence of material stiffness and geometric size on the cable load responses and formation configurations.

\newlength{\imgHeightTensionResponse}
\setlength{\imgHeightTensionResponse}{5cm}
\begin{center}
    \begin{minipage}{0.48\textwidth}
        \centering
        \includegraphics[height=\imgHeightTensionResponse,keepaspectratio]{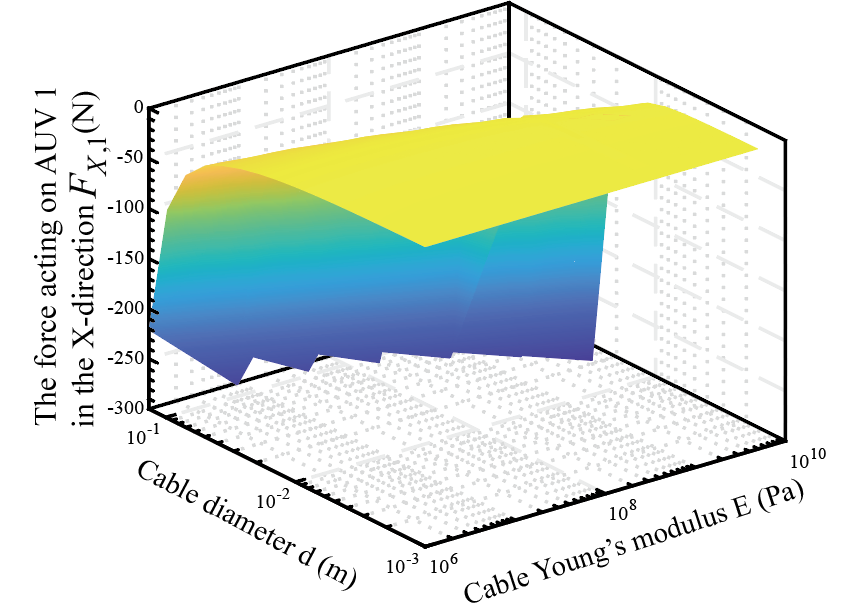}
        \vspace{0.5em}
        \text{\makecell{(a) Cable tension $F_{X,1}$ of $AUV_{1}$ in X direction}}

    \end{minipage}
    \hfill
    \begin{minipage}{0.48\textwidth}
        \centering
        \includegraphics[height=\imgHeightTensionResponse,keepaspectratio]{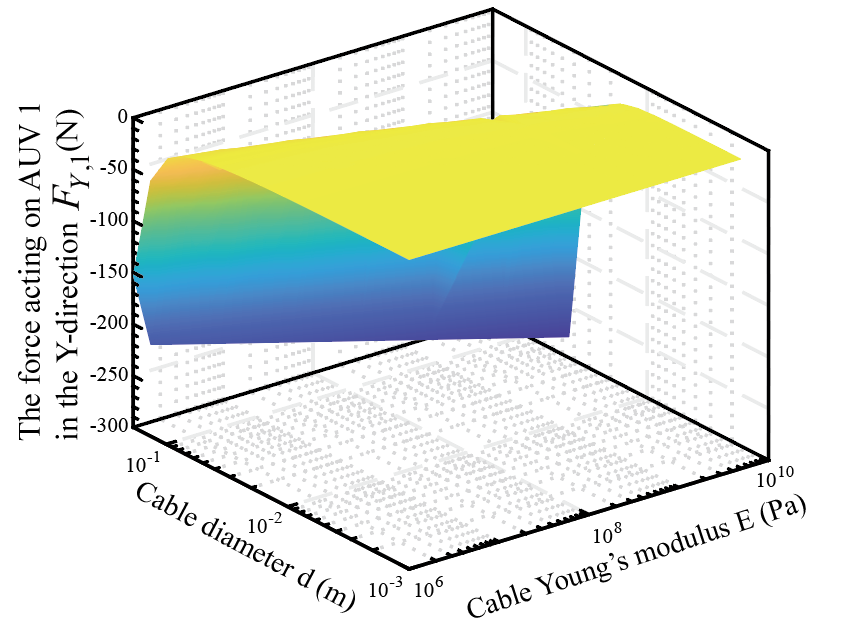}
        \vspace{0.5em}
        \text{\makecell{(b) Cable tension $F_{Y,1}$ of $AUV_{1}$ in Y direction}}
    \end{minipage}
    
   \begin{minipage}{0.48\textwidth}
        \centering
        \includegraphics[height=\imgHeightTensionResponse,keepaspectratio]{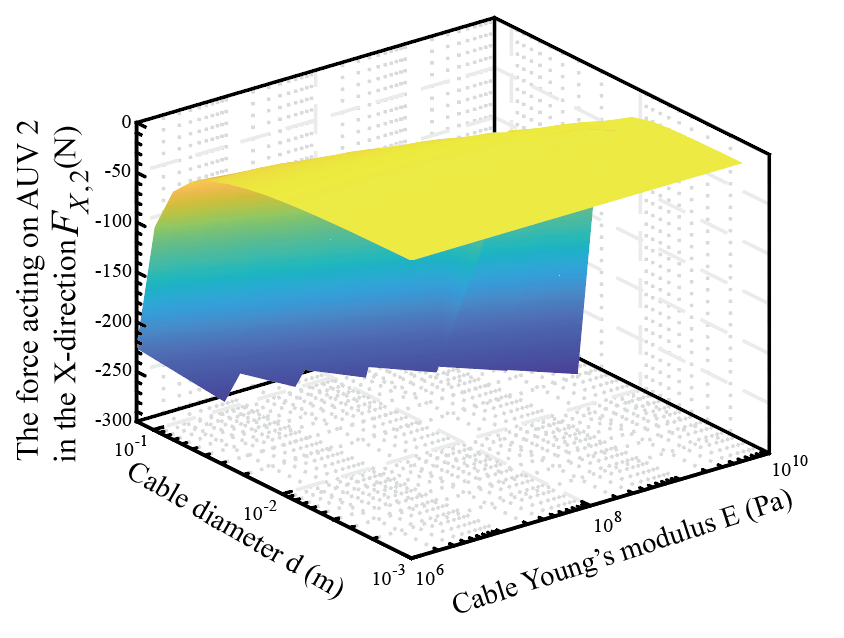}
        \vspace{0.5em}
        \text{\makecell{(c) Cable tension $F_{X,2}$ of $AUV_{2}$ in X direction}}
    \end{minipage}
    \hfill
    \begin{minipage}{0.48\textwidth}
        \centering
        \includegraphics[height=\imgHeightTensionResponse,keepaspectratio]{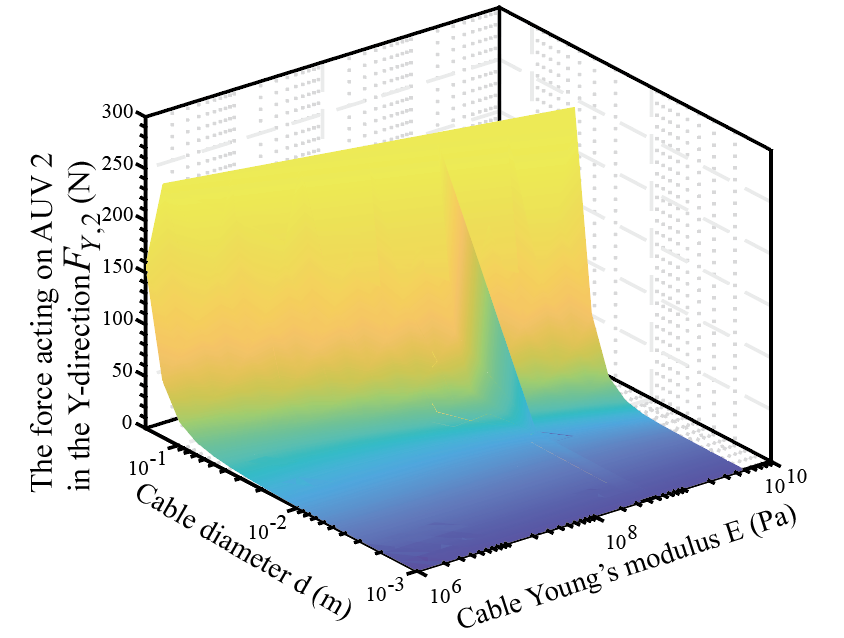}
        \vspace{0.5em}
        \text{\makecell{(d) Cable tension $F_{Y,2}$ of $AUV_{2}$ in Y direction}}
    \end{minipage}

    \captionof{figure}{Tension Response of Dual-AUV system in X and Y Directions under Different Combinations \\ of Cable Diameter and Young's Modulus} 
    \label{TensionResponse}
\end{center}

Figure \ref{TensionResponse} illustrates the three-dimensional response surfaces of force components in the X and Y directions acting on $AUV_1$ and $AUV_2$  across various combinations of $d$ and $E$. The simulation results indicate that the increase in the cable diameter has a limited influence on the tension of the system at the initial stage. However, once the diameter exceeds a certain threshold, the tension increases significantly, exhibiting  distinct nonlinear leap characteristics. In particular, this threshold is delayed by the increase in the Young's modulus, indicating that high-stiffness materials are able to suppress the sudden change in forces caused by the diameter to a certain extent. In practical applications, materials whose parameter combinations lie beyond this threshold region should be avoided to prevent sudden increases in force responses. Overall, although an increase in $E$ also leads to a rise in tension, the modulation effect is significantly weaker than that of $d$, indicating that the tension response exhibits low sensitivity to variations in $E$.

Figure \ref{morphology} further illustrates the evolution of planar configurations as $d$ and $E$ are changed separately. As demonstrated in Figure \ref{morphology} (a), the cable diameter significantly influences the cable configuration, with the maximum midpoint displacement difference approaching 1 m. As the diameter increases, the lateral position $X_{mid}$ initially moves forward followed by a backward movement. This phenomenon is attributed to the increased cable diameter, which enhances both hydrodynamic drag effects and structural rigidity. The competition between these two factors results in the observed non-monotonic deformation trend. As demonstrated in Figure \ref{morphology} (b), the increase of the Young's modulus causes  the overall cable configuration to exhibit a tendency toward the taut state, which is specifically characterized by the sustained straightening behavior, and $X_{mid}$ moves forward monotonically. Significant variations in the cable configuration occur only when $E < 10^{7}$ Pa, suggesting that the influence of material stiffness on the cable configuration exists within an effective range, and the influence tends to saturate in the high-modulus region.

\newlength{\imgHeightmorphology}
\setlength{\imgHeightmorphology}{6cm}
\begin{center}
    \begin{minipage}{0.48\textwidth}
        \centering
        \includegraphics[width=\imgHeightmorphology,keepaspectratio]{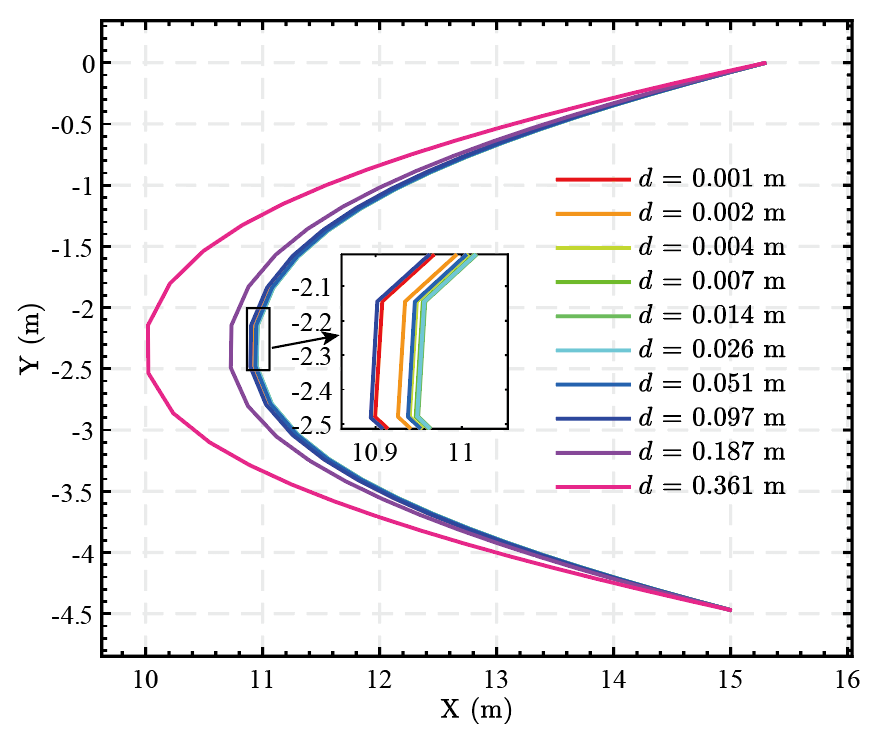}
        \vspace{0.5em}
        \text{\makecell{(a) Simulation results of fixed Young's Modulus $E$}}

    \end{minipage}
    \hfill
    \begin{minipage}{0.48\textwidth}
        \centering
        \includegraphics[width=\imgHeightmorphology,keepaspectratio]{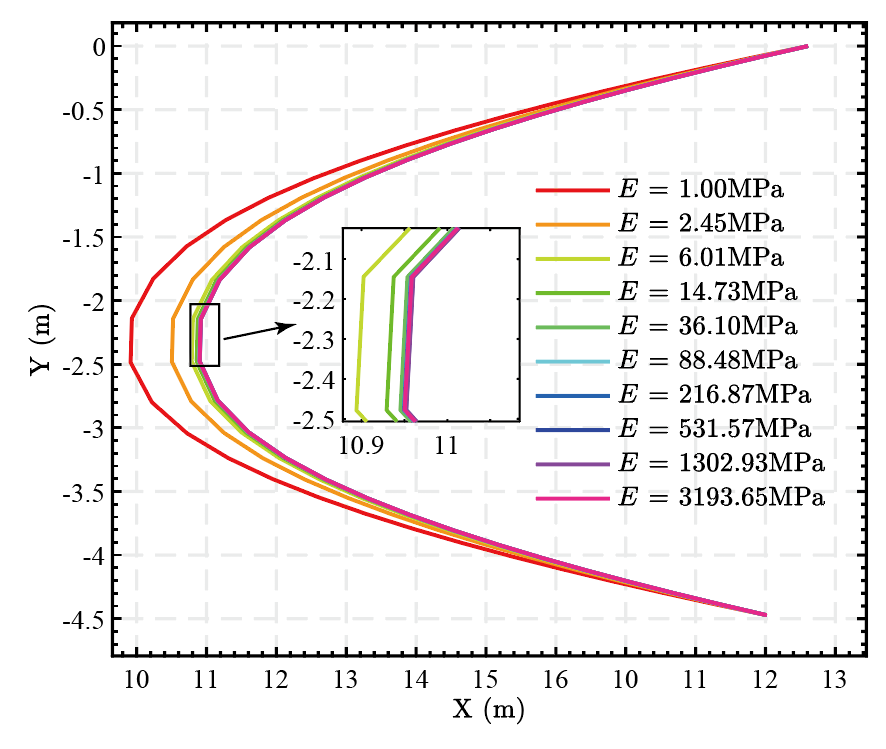}
        \vspace{0.5em}
        \text{\makecell{(b) Simulation results of fixed diameter $d$}}
    \end{minipage}

    \captionof{figure}{Changes in cable morphology under different diameters and Young's modulus conditions} 
    \label{morphology}
\end{center}

\begin{center}
    \includegraphics[width=8cm]{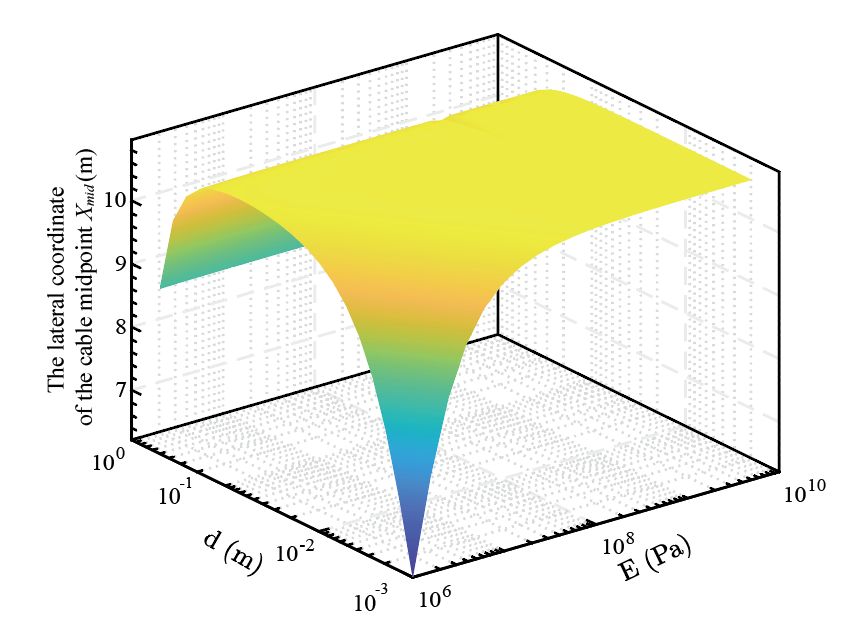}
    \captionof{figure}{Lateral Position $X_{mid}$ of Cable Midpoint under Varying Cable Diameter and Elastic Modulus} 
    \label{LateralPosition}
\end{center}

The variation of lateral position $X_{mid}$ with respect to $d$ and $E$ is further presented in Figure \ref{LateralPosition} in the form of three-dimensional response surfaces, which demonstrates the influence trends of parameter combinations on the cable configuration in a more intuitive format. As shown in this figure, instead of the monotonic relationship, a minimum point exists in the correspondence between the $X_{mid}$ and the diameter, which reflects the combined effect of the coupling between bending stiffness and hydrodynamic resistance. In contrast, $X_{mid}$ is consistently positively correlated with the Young's modulus across the entire range, which provides further validation for the stabilizing effect of the medium-to-high Young's modulus.

\subsection{Simulation analysis of endpoint tension responses and cable configurations under different cable length conditions.}

The cable length is an important factor affecting the dynamic behavior and the cable configuration of the flexibly connected dual-AUV systems, the variation of which directly effects the system stability, load distribution and control strategy design. Therefore, this section investigates the dynamic behavior and steady-state cable configuration of the flexibly connected dual-AUV system under different cable lengths. First, simulations are performed for a range of formations that follow a predefined geometric pattern, where the lateral separation is set to $X_d = 0.5 L_c$, and the longitudinal separation is $Y_d = 0$ m. The variation of cable configurations with respect to  lengths is presented in Figure \ref{CableShapeDistribution1} (a). As observed in this figure, the cable configurations maintain a high degree of similarity despite the continuous increase in $L_c$. To further reveal the characteristics of similarity, a proportional scaling is performed by normalizing the coordinates with respect to the cable length, thereby realizing a similarity transformation based on the unit length. The normalization results are shown in Figure \ref{CableShapeDistribution1} (b), demonstrating remarkable consistency among cable configurations of various lengths. These results demonstrates that the geometric pattern  remains independent of the cable length $L_c$.

% \begin{center}
%     \includegraphics[width=9cm]{CableShapeDistribution1.png}
%     \captionof{figure}{Cable Shape Distribution under the Two-AUV Spacing Configuration $X_d=0.5L_c$, $Y_d=0$} 
%     \label{CableShapeDistribution1}
% \end{center}

% \begin{center}
%     \includegraphics[width=9cm]{CableShapeDistribution2.png}
%     \captionof{figure}{Cable Shape Distribution after Similarity Transformation under the Two-AUV \\ Spacing Configuration $X_d=0.5L_c$, $Y_d=0$} 
%     \label{CableShapeDistribution2}
% \end{center}

\newlength{\imgHeightCableShapeDistribution}
\setlength{\imgHeightCableShapeDistribution}{4.8cm}
\begin{center}
    \begin{minipage}{0.48\textwidth}
        \centering
        \includegraphics[height=\imgHeightCableShapeDistribution,keepaspectratio]{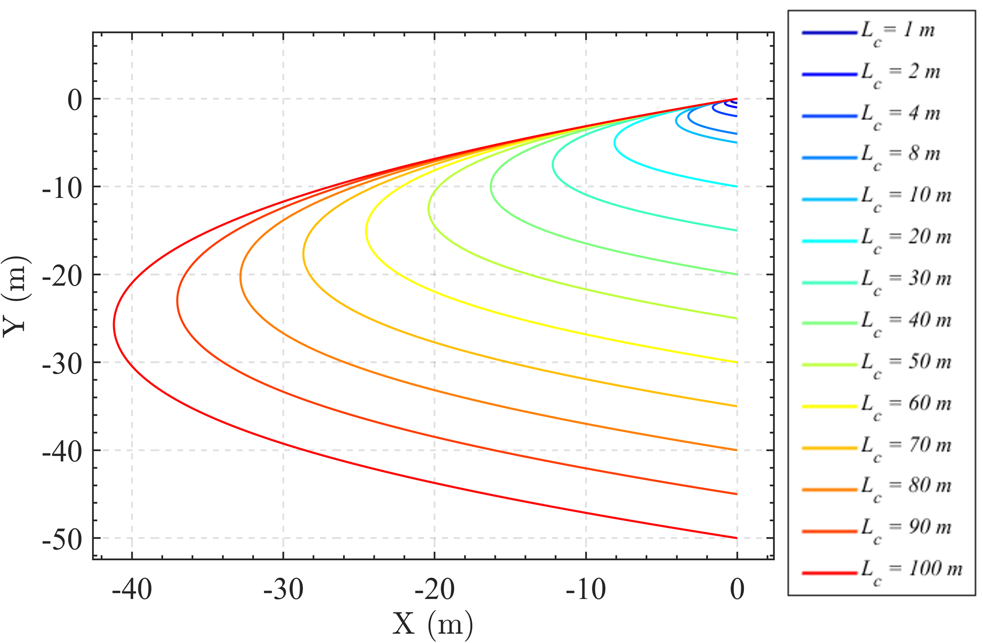}
        \vspace{0.5em}
        \text{\makecell{(a) Cable Configuration}}

    \end{minipage}
    \hfill
    \begin{minipage}{0.48\textwidth}
        \centering
        \includegraphics[height=\imgHeightCableShapeDistribution,keepaspectratio]{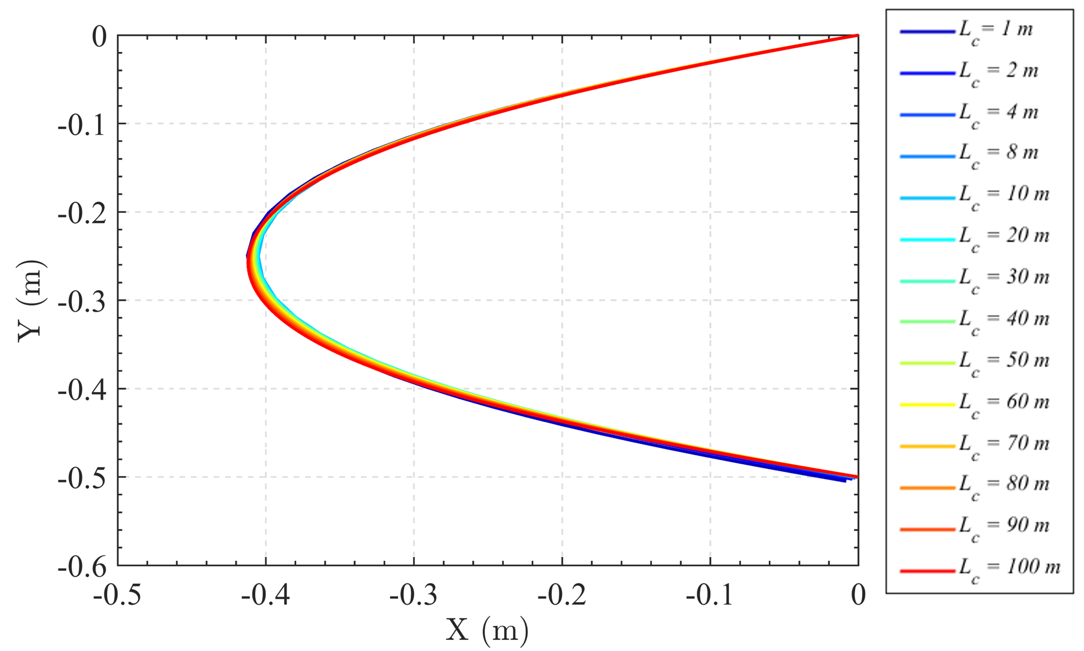}
        \vspace{0.5em}
        \text{\makecell{(b) Cable Configuration after Similarity Transformation}}
    \end{minipage}

    \captionof{figure}{Cable shape distribution under the Two-AUV spacing configuration $X_d=0.5L_c$, $Y_d=0$ and \\ the corresponding normalization results} 
    \label{CableShapeDistribution1}
\end{center}

Subsequently, the behavior of another dual-AUV formation is investigated, with lateral separation $X_d = 0.8L_c$ and longitudinal separation $Y_d = 0.3 L_c$. The original cable configurations are shown in Figure \ref{CableShapeDistribution2} (a), while the corresponding normalization results are provided in Figure \ref{CableShapeDistribution2} (b). Through the comparison of these simulation results, a significant reduction in similarity of the cable configuration is observed. As the separation distance approaches $L_c$, the effects of elastic tension and bending forces become more significant, which leads to a disruption of the previously observed similarity. Finally, Figure \ref{CableShapeDistribution5} illustrates the maximum coordinate differences at the midpoint for cable configurations of various lengths. These differences are calculated based on the scaling transformation, obtained by dividing the midpoint coordinates by  $L_c$. The results indicate that as the separation distance approaches $L_c$, the cable gradually transitions toward the taut state, causing the geometric similarity between cable configurations of various lengths to diminish significantly. Conversely, in the slack state, the system is dominated by the hydrodynamic drag, which makes the geometric similarity more evident.

% \begin{center}
%     \includegraphics[width=9cm]{CableShapeDistribution3.png}
%     \captionof{figure}{Cable Shape Distribution under the Two-AUV \\ Spacing Configuration $X_d=0.8L_c$, $Y_d=0.3L_c$} 
%     \label{CableShapeDistribution3}
% \end{center}

% \begin{center}
%     \includegraphics[width=9cm]{CableShapeDistribution4.png}
%     \captionof{figure}{ Cable Shape Distribution after Similarity Transformation under the Two-AUV \\ Spacing Configuration $X_d=0.8L_c$, $Y_d=0.3L_c$} 
%     \label{CableShapeDistribution4}
% \end{center}

% \newlength{\imgHeightCableShapeDistributionbig}
% \setlength{\imgHeightCableShapeDistributionbig}{8cm}
\begin{center}
    \begin{minipage}{0.48\textwidth}
        \centering
        \includegraphics[height=\imgHeightCableShapeDistribution,keepaspectratio]{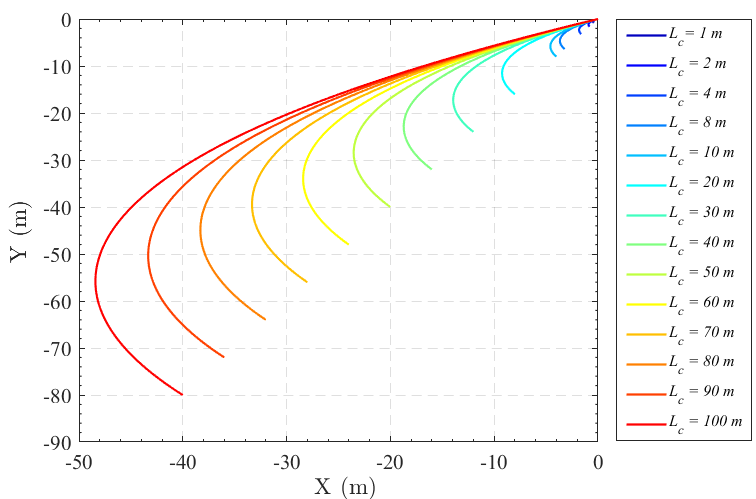}
        \vspace{0.5em}
        \text{\makecell{(a) Cable Configuration}}

    \end{minipage}
    \hfill
    \begin{minipage}{0.48\textwidth}
        \centering
        \includegraphics[height=\imgHeightCableShapeDistribution,keepaspectratio]{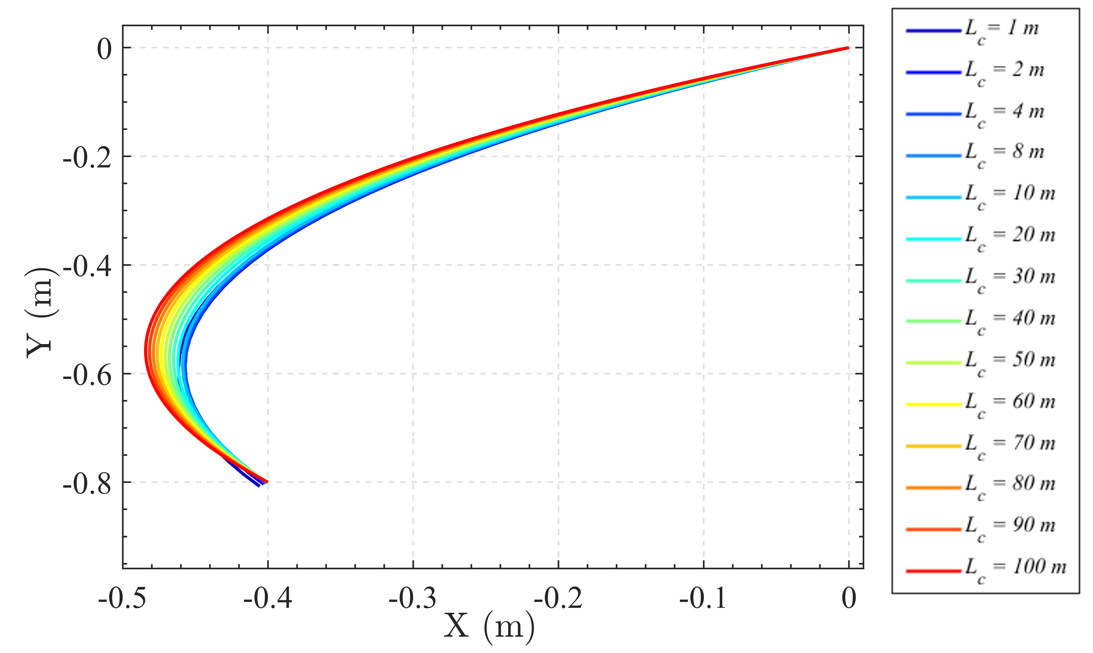}
        \vspace{0.5em}
        \text{\makecell{(b) Cable Configuration after Similarity Transformation}}
    \end{minipage}

    \captionof{figure}{Cable  configuration under the Two-AUV spacing configuration $X_d=0.8L_c$, $Y_d=0.3$ and \\ the corresponding normalization results} 
    \label{CableShapeDistribution2}
\end{center}

In this section,  an analysis of the dynamic behavior of the flexibly connected dual-AUV system is presented, accounting for variations in both cable length and formation geometry. To enhance clarity,  lateral and longitudinal separation distances of the system are normalized to the range of [0,1]. The variations of the cable forces acting on the AUV platforms with respect to cable lengths are presented in Figure \ref{CableShapeDistribution6}, in which various cable configurations are included. According to the results presented,  increasing the cable length does not result in a corresponding growth in the forces, indicating that the two variables are not positively correlated. In contrast, under extremely short cable length conditions ($L_c$=1 m), the endpoint forces increase significantly, which is attributed to the dominant influence of bending stiffness and the resulting substantially larger bending moments. Although the hydrodynamic drag is decreased to a certain extent, the magnitude of the bending effect outweighs this reduction, which ultimately leads to the increase of resultant forces exerted on the endpoints.

\begin{center}
    \includegraphics[width=8cm]{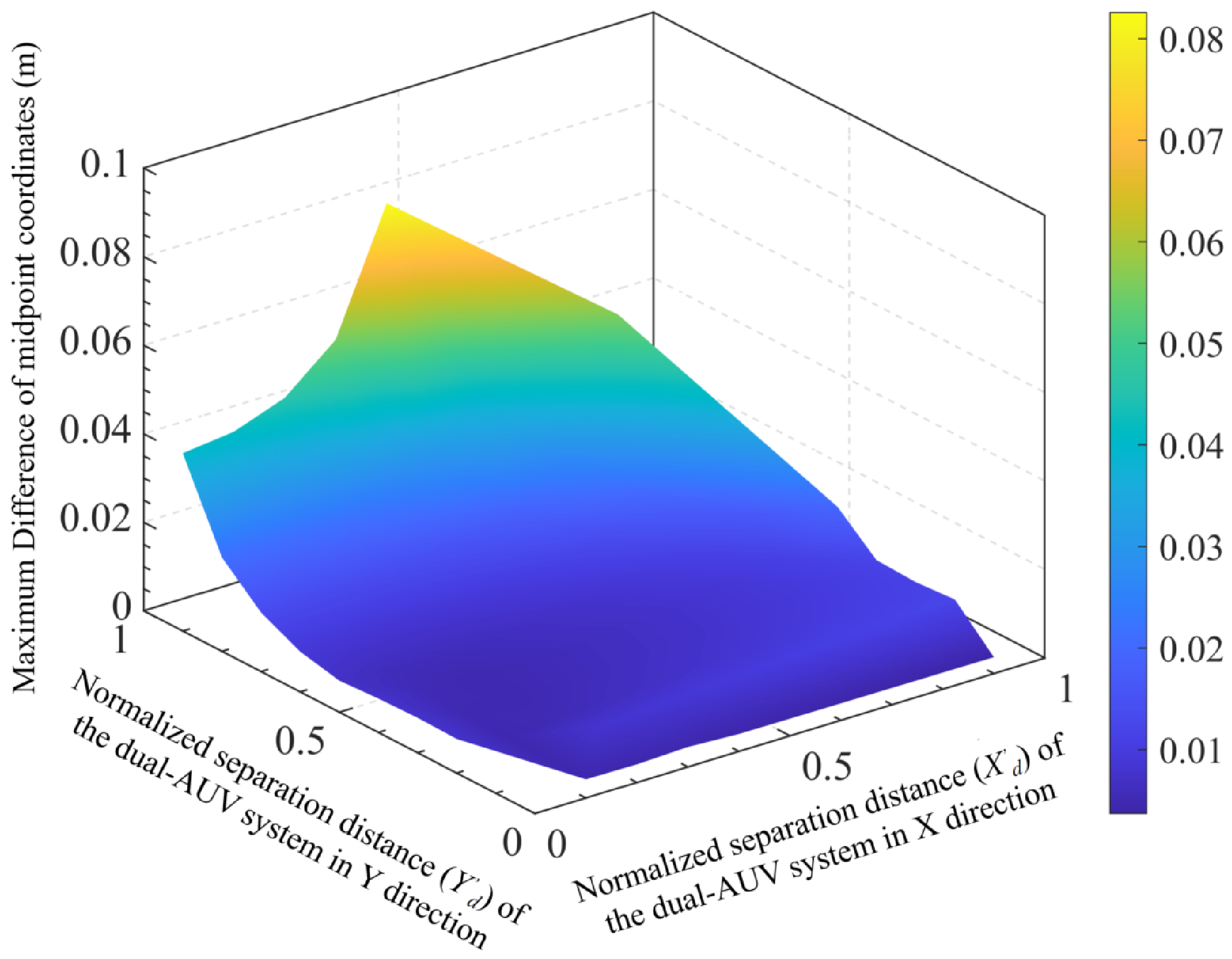}
    \captionof{figure}{Distribution of Midpoint Coordinate Differences of Cables \\ under Different Similar Configurations} 
    \label{CableShapeDistribution5}
\end{center}

To provide a more explicit illustration of endpoint forces varying with cable lengths under typical configurations, Figure \ref{CableShapeDistribution7} is further depicted. In this typical configuration, the lateral separation distance is gradually extended to $X_d=0.9L_c$, while the longitudinal separation is fixed at $Y_d = 0$. The illustrated surfaces corresponding to this extension indicate that, instead of a monotonic relationship, the endpoint forces initially increase and then decrease as the cable length increases. The analysis results demonstrate that both cable stiffness effects and hydrodynamic drag significantly affect the dynamic response of the AUV platform. In the  taut state, system forces are  dominated by elastic effects, while the hydrodynamic drag become the governing factor in the slack state. These two mechanisms are coupled under different cable length conditions, jointly determining the load distribution and response characteristics at the endpoints. Within the studied conditions, when the cable length increases to 8 m, the system undergoes a transition from the taut state dominated by elastic forces to the slack state  governed by hydrodynamic effects.

\captionsetup{justification=centering}
% 定义高度
\setlength{\imgHeight}{7.2cm}

\begin{center}
    \begin{minipage}{0.4\textwidth}
        \centering
        \includegraphics[width=\imgHeight,keepaspectratio]{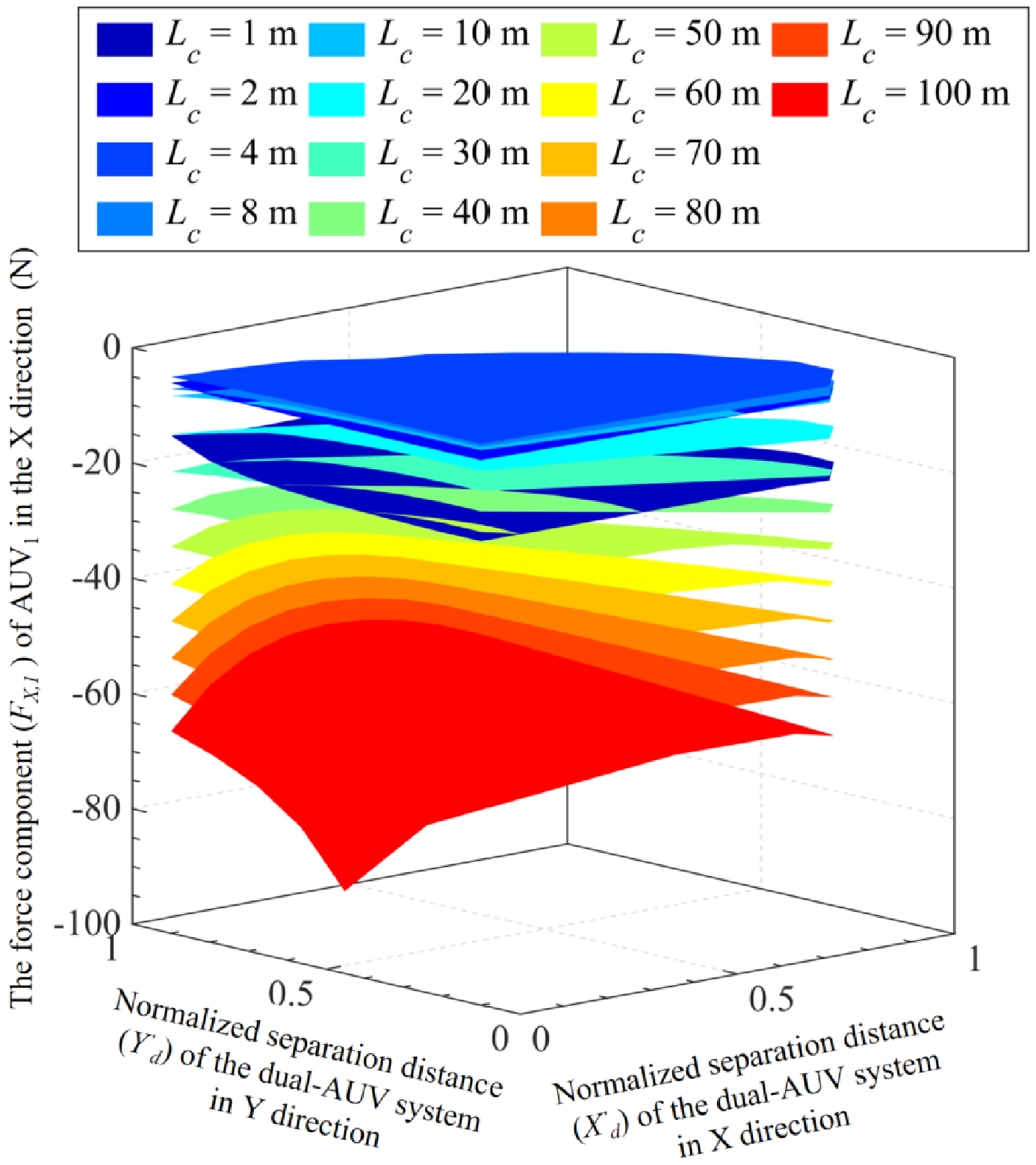}
        \vspace{0.5em}
        \text{\makecell{(a) Cable tension $F_{X,1}$ of $AUV_{1}$ in X direction}}

    \end{minipage}
    \hspace{2cm}
    \begin{minipage}{0.4\textwidth}
        \centering
        \includegraphics[width=\imgHeight,keepaspectratio]{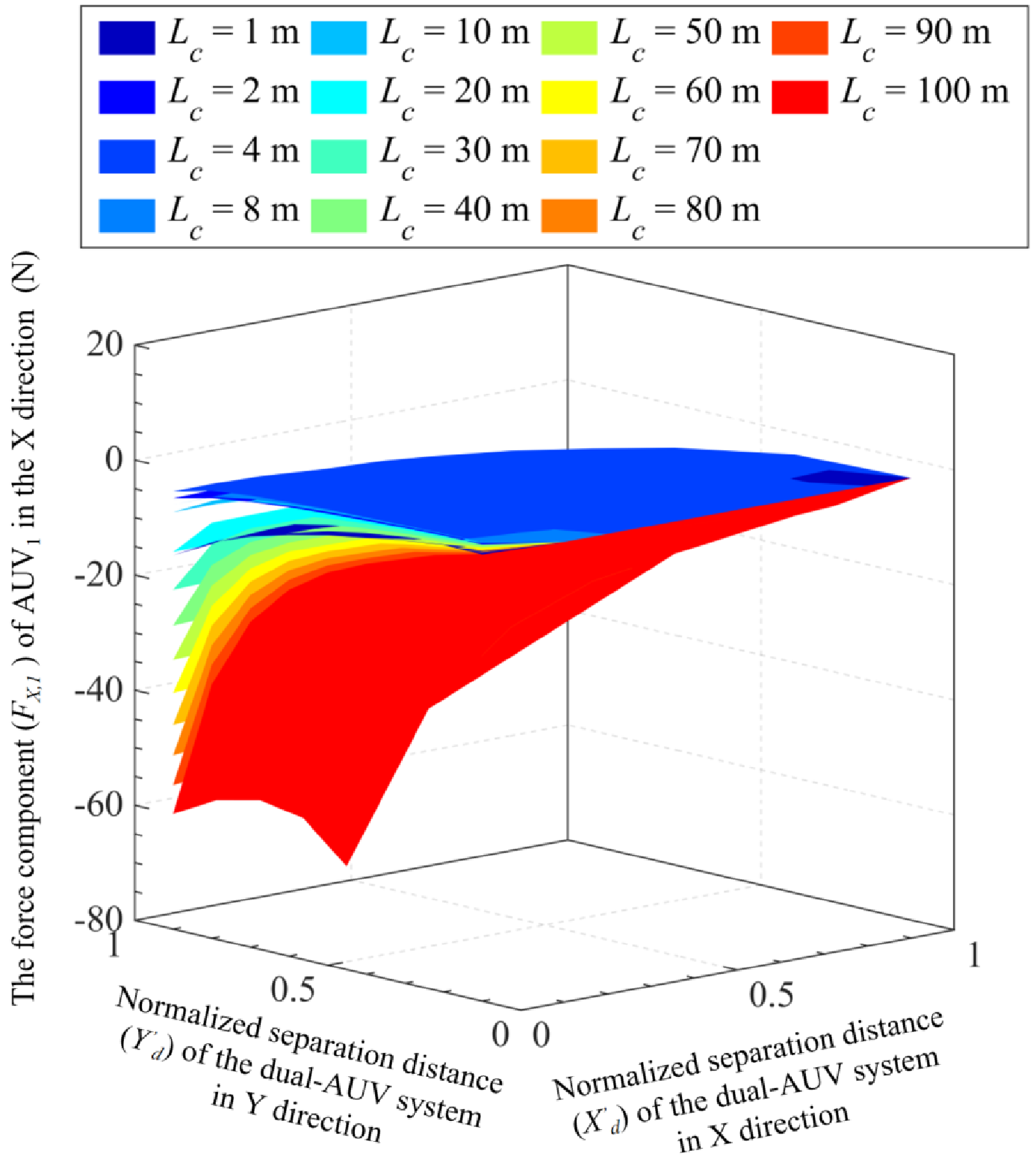}
        \vspace{0.5em}
        \text{\makecell{(b) Cable tension $F_{Y,1}$ of $AUV_{1}$ in Y direction}}
    \end{minipage}
    
   \begin{minipage}{0.4\textwidth}
        \centering
        \includegraphics[width=\imgHeight,keepaspectratio]{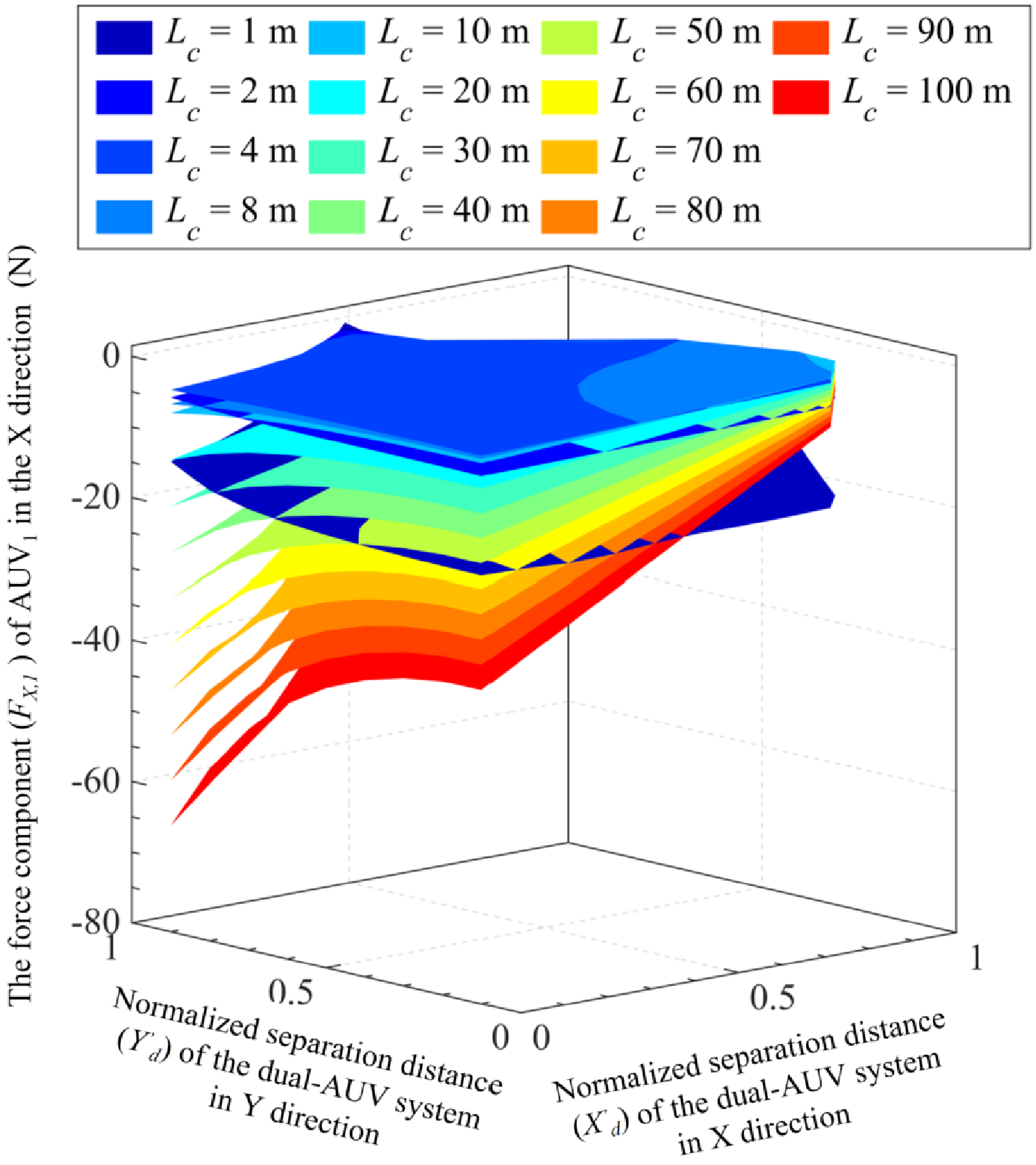}
        \vspace{0.5em}
        \text{\makecell{(c) Cable tension $F_{X,2}$ of $AUV_{2}$ in X direction}}
    \end{minipage}
    \hspace{2cm}
    \begin{minipage}{0.4\textwidth}
        \centering
        \includegraphics[width=\imgHeight,keepaspectratio]{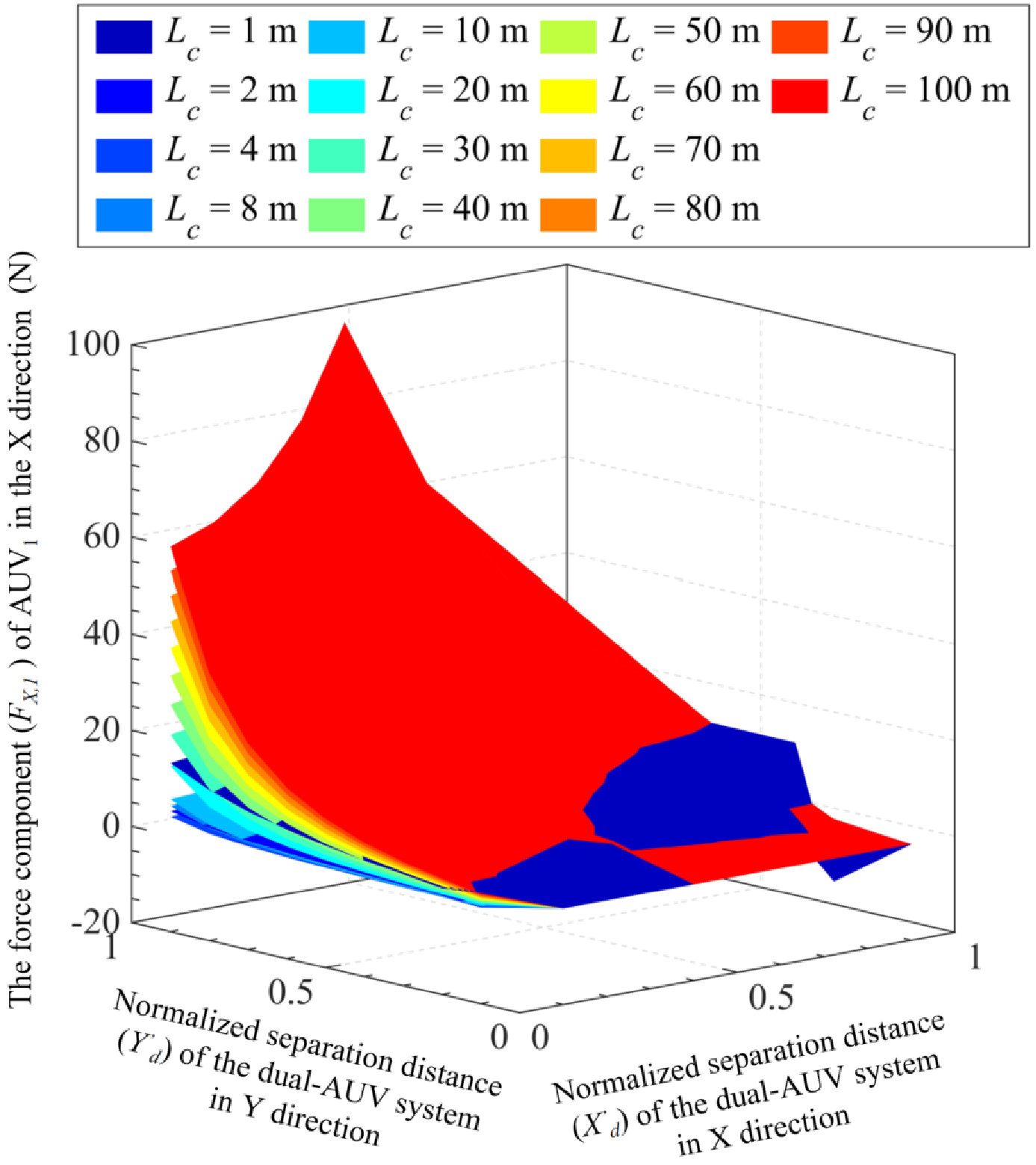}
        \vspace{0.5em}
        \text{\makecell{(d) Cable tension $F_{Y,2}$ of $AUV_{2}$ in Y direction}}
    \end{minipage}

    \captionof{figure}{Variation of Forces on Endpoints of the AUV under Different Configurations and Cable Lengths} 
    \label{CableShapeDistribution6}
\end{center}

\begin{center}
    \begin{minipage}{0.48\textwidth}
        \centering
        \includegraphics[width=\linewidth]{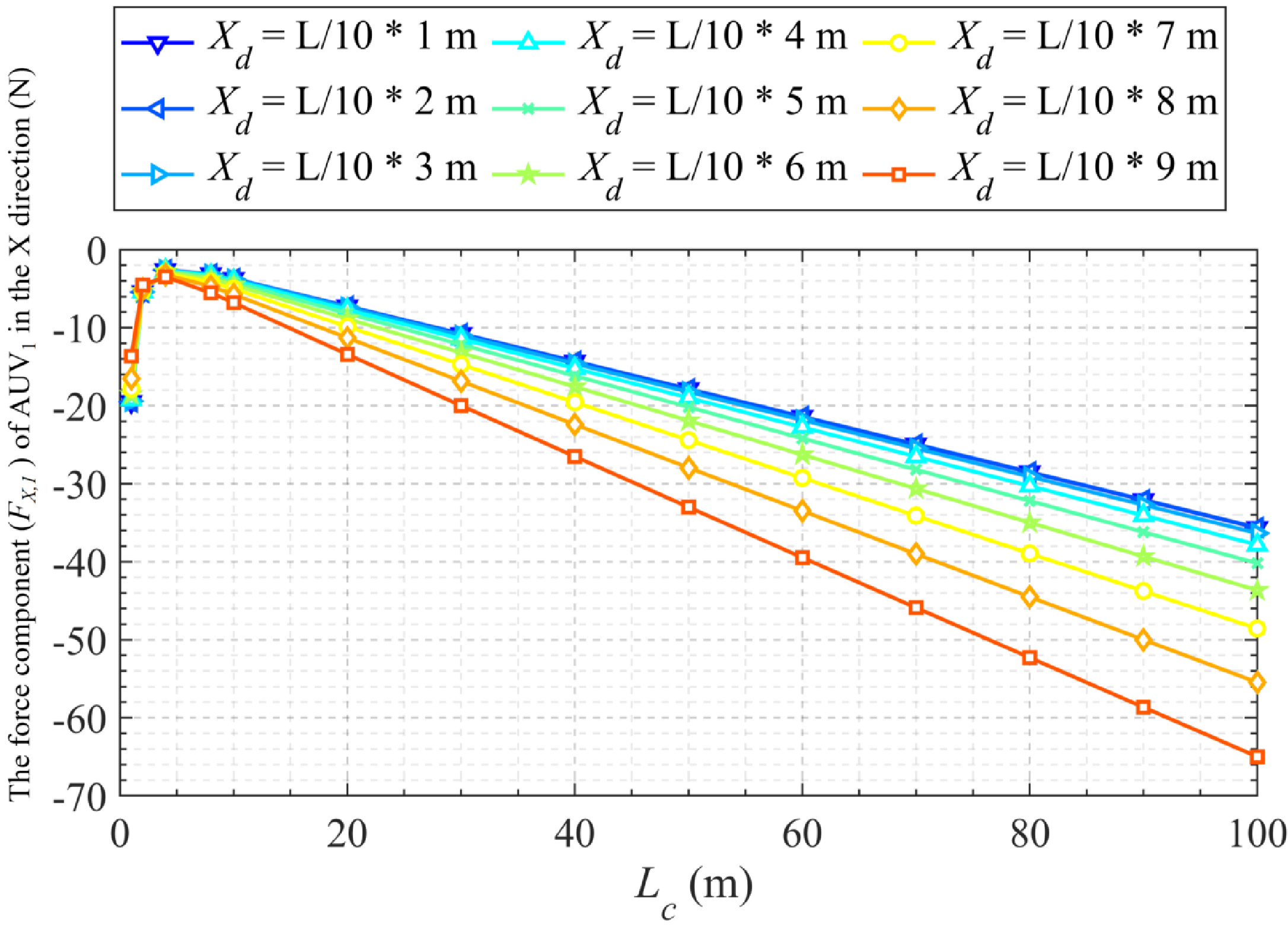}
        \vspace{0.5em}
        \text{\makecell{(a) Cable tension $F_{X,1}$ of $AUV_{1}$ in X direction}}

    \end{minipage}
    \hfill
    \begin{minipage}{0.48\textwidth}
        \centering
        \includegraphics[width=\linewidth]{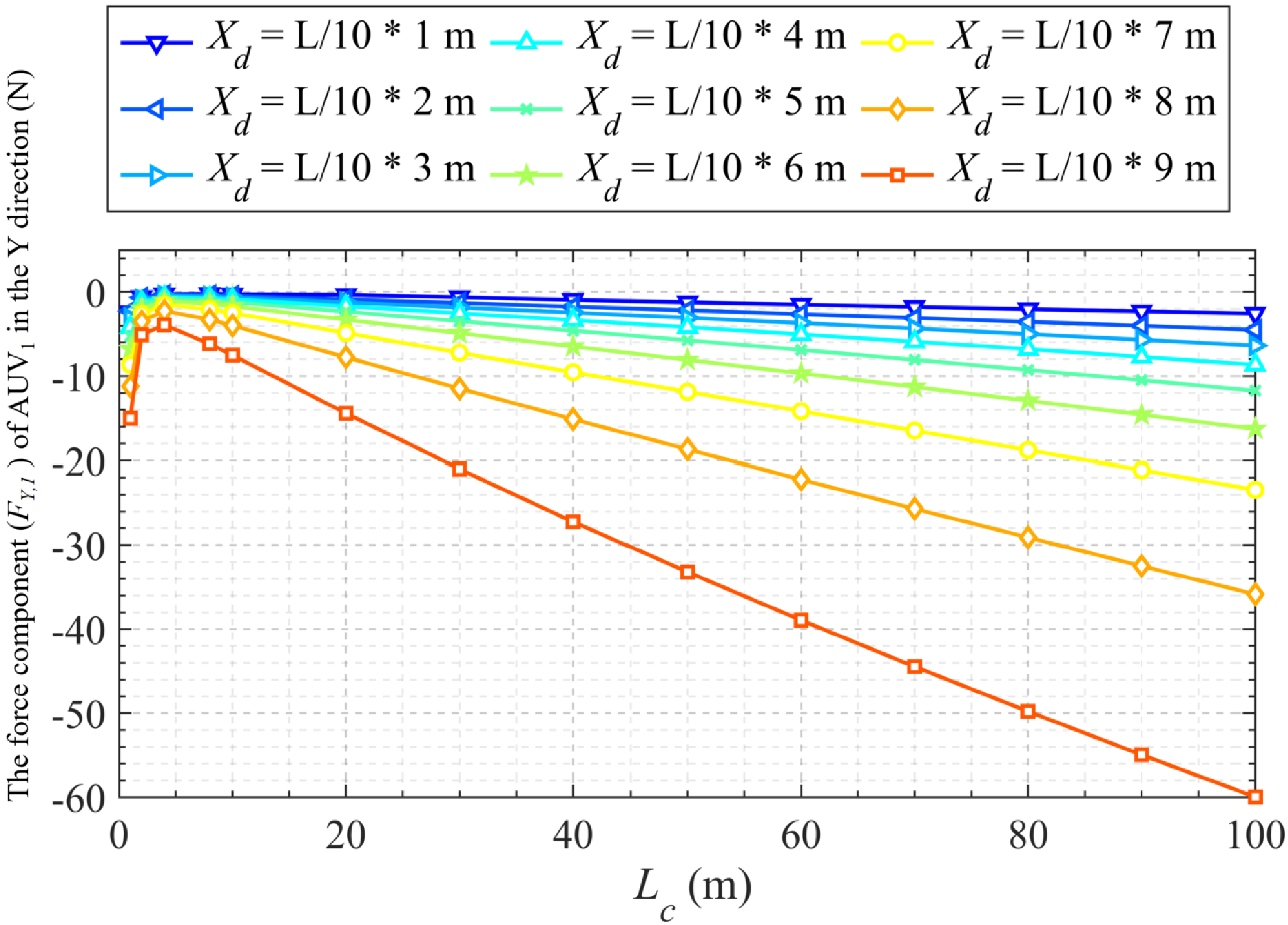}
        \vspace{0.5em}
        \text{\makecell{(b) Cable tension $F_{Y,1}$ of $AUV_{1}$ in Y direction}}
    \end{minipage}
    
   \begin{minipage}{0.48\textwidth}
        \centering
        \includegraphics[width=\linewidth]{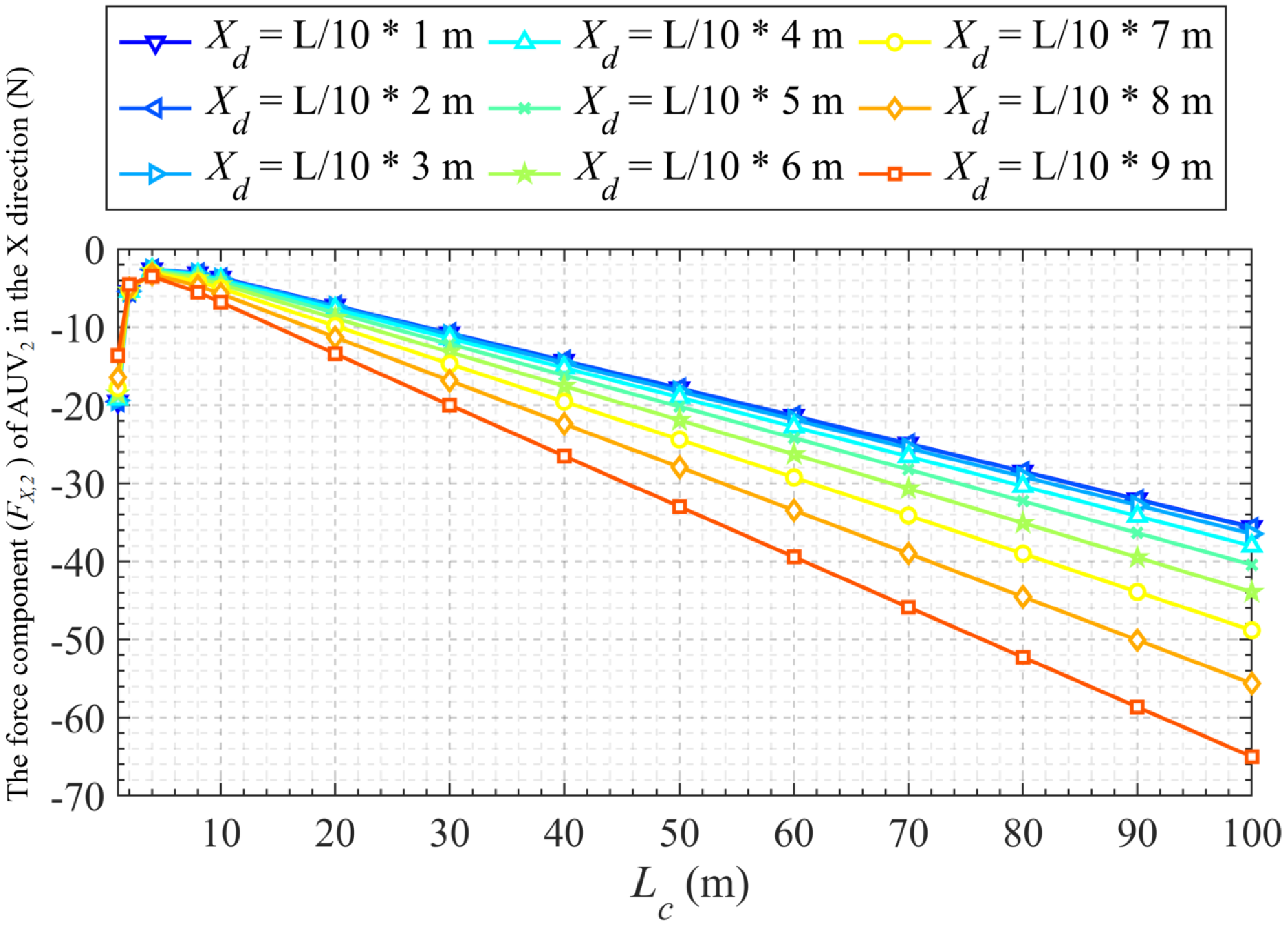}
        \vspace{0.5em}
        \text{\makecell{(c) Cable tension $F_{X,2}$ of $AUV_{2}$ in X direction}}
    \end{minipage}
    \hfill
    \begin{minipage}{0.48\textwidth}
        \centering
        \includegraphics[width=\linewidth]{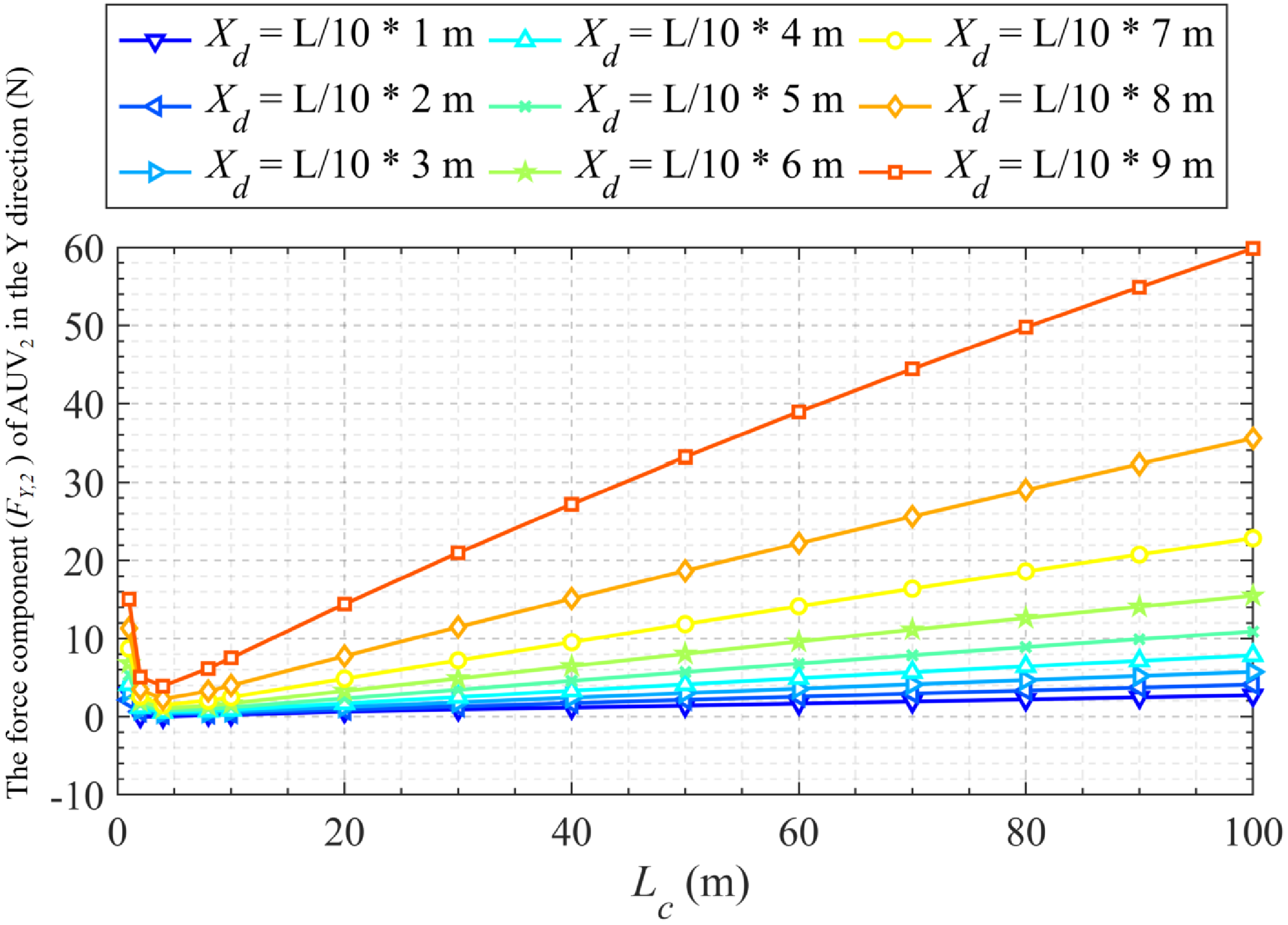}
        \vspace{0.5em}
        \text{\makecell{(d) Cable tension $F_{Y,2}$ of $AUV_{2}$ in Y direction}}
    \end{minipage}

    \captionof{figure}{ Variation of AUV End Forces with Cable Length under a Typical Configuration} 
    \label{CableShapeDistribution7}
\end{center}

\subsection{Simulation analysis of endpoint forces and cable configurations in the  transition from horizontal to vertical formations}

In practical marine operations, the flexibly connected dual-AUV system requires the ability of formation reconfiguration from a horizontal to a vertical alignment. To further investigate the dynamic characteristics of the formation transition under varying cable configurations, a series of simulations involving non-uniform motion states is conducted. The main modification from the previous analysis is that the cable no longer maintains neutral buoyancy, and the cable density is now specified as 1300 kg/m$^3$. Meanwhile, the two AUVs  are initially aligned along the X direction with a fixed distance of 8 m, with no relative displacement in the Y and Z directions. For the final state after the transition, the separation distance between the two AUVs is 8 m along the Z direction, with no relative displacement in the X and Y directions. The response of velocity components in X, Y and Z directions is presented in Figure \ref{AUV_velocity_transformation_curve}. According to the curves illustrated, the system starts  from a zero-velocity state. Following the synchronous acceleration initiated at $t=10$ s, the velocities of both AUV platforms eventually converge to a steady-state value of 0.5 m/s. Starting at $t$=55 s, the introduction of Y and Z velocity components leads to a decrease in lateral separation distance and an increase in vertical distance, while the system maintains its forward motion in the X direction.

\begin{center}
    \includegraphics[width=8cm]{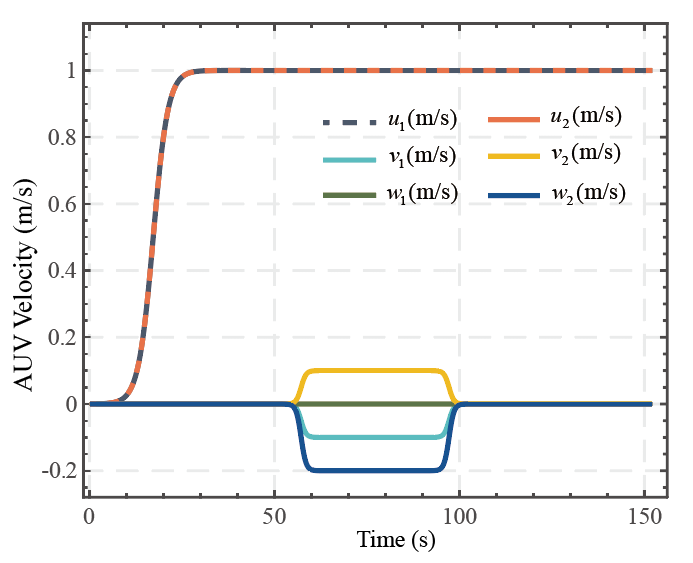}
    \captionof{figure}{AUV velocity transformation curve} 
    \label{AUV_velocity_transformation_curve}
\end{center}

Figure \ref{perspectives} demonstrates the time-varying evolution process of the three-dimensional cable configuration from different perspectives.  The color gradient represents the simulation time (ranging from blue to yellow for 0 s to 150 s), while the circles and triangles denote $AUV_1$ and $AUV_2$. In Figure \ref{perspectives}, (a)-(d) correspond to the spatial, top, side, and front views, respectively. As indicated by these results,  the gravitational effect is fully manifested in the initial stage, causing the midpoint of the cable to exhibit significant vertical displacement. However, the configuration also shows a clear evolution trend from a gravity-dominated state towards a state governed by hydrodynamic forces. As the system accelerates in the X direction and reaches the position of $x=5$ m in the Earth-fixed coordinate system, hydrodynamic forces become increasingly dominant, suppressing  the vertical displacement of the midpoint and resulting in an overall backward movement of the cable. Consequently, the geometry of the cable gradually approaches a horizontally planar  configuration. In the next stage, as the system reaches the  position of $x=40$ m, the velocity components in Y and Z directions are introduced, initiating the transition from horizontal to vertical formations. During this transition, the cable deviates from the taut state, and the backward movement is gradually reduced. In the final stage, as the system reaches the  position of $x=90$ m, the transition process is eventually completed. The dual-AUV system maintains this configuration while sustaining forward motion along the $X$ direction until the simulation is completed.

\newlength{\imgHeightspectrala}
\setlength{\imgHeightspectrala}{3.7cm}

\begin{center}
    \begin{minipage}{0.6\textwidth}
        \centering
        \includegraphics[height=\imgHeightspectrala,keepaspectratio]{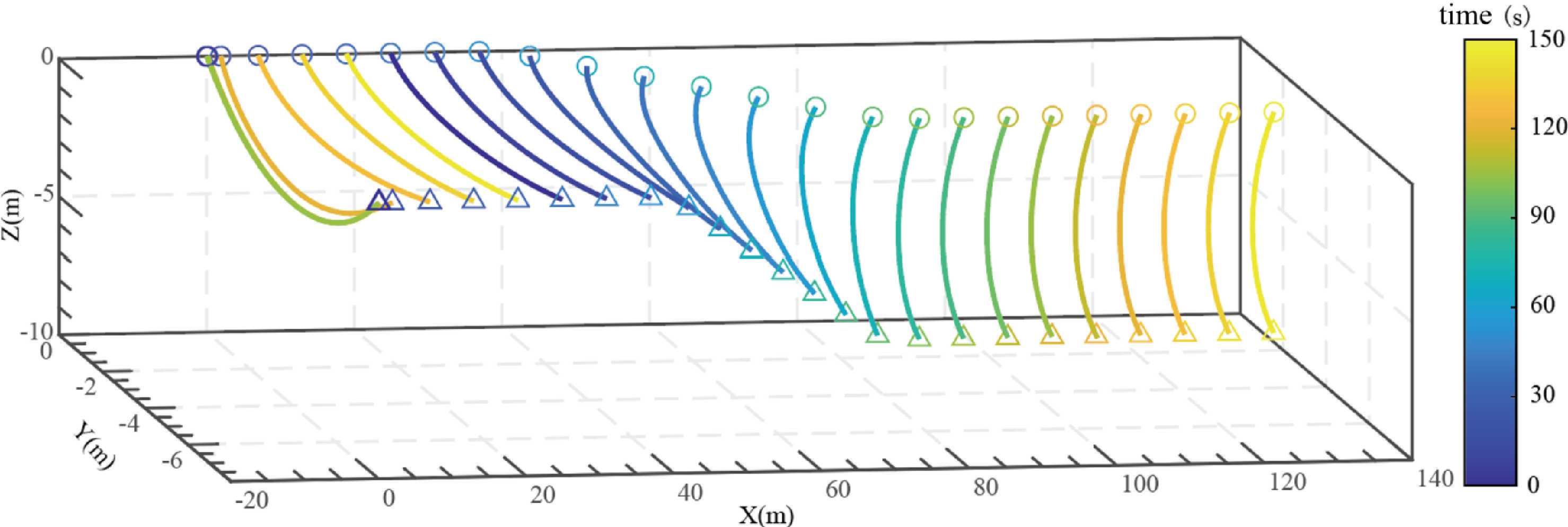}
        \vspace{0.5em}
        \text{\makecell{(a) Spatial view}}

    \end{minipage}

    \begin{minipage}{0.6\textwidth}
        \centering
        \includegraphics[height=\imgHeightspectrala,keepaspectratio]{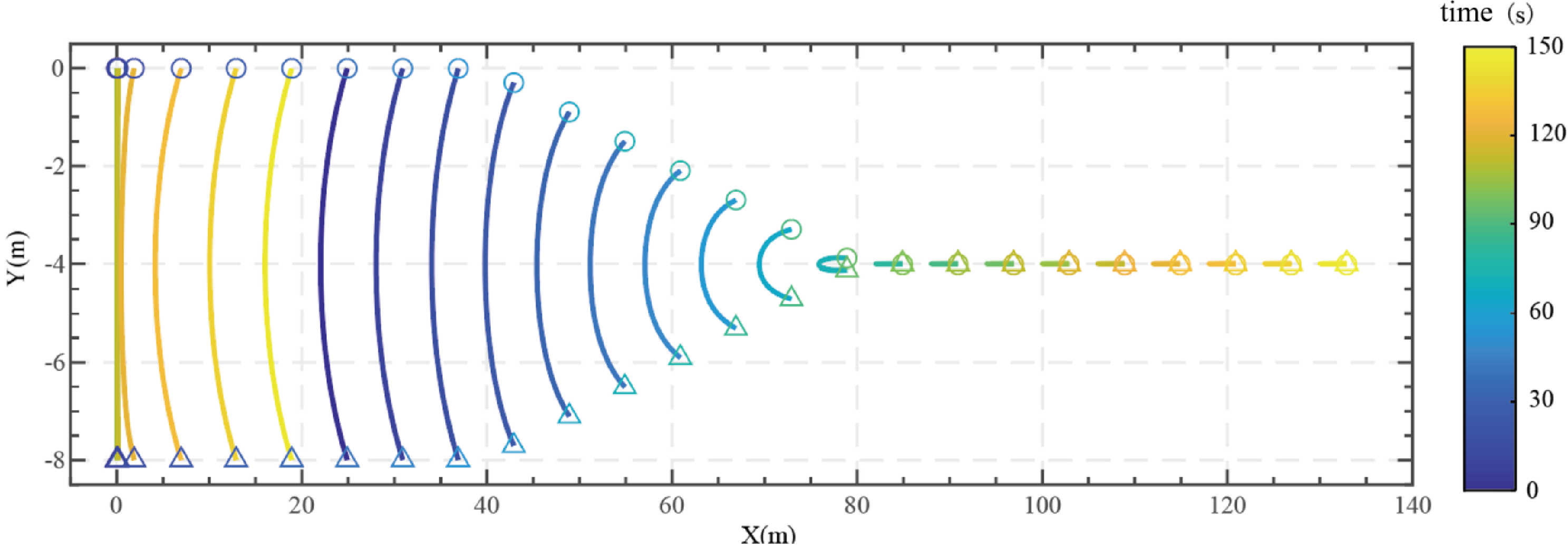}
        \vspace{0.5em}
        \text{\makecell{(b) Top view}}
    \end{minipage}
    
   \begin{minipage}{0.6\textwidth}
        \centering
        \includegraphics[height=\imgHeightspectrala,keepaspectratio]{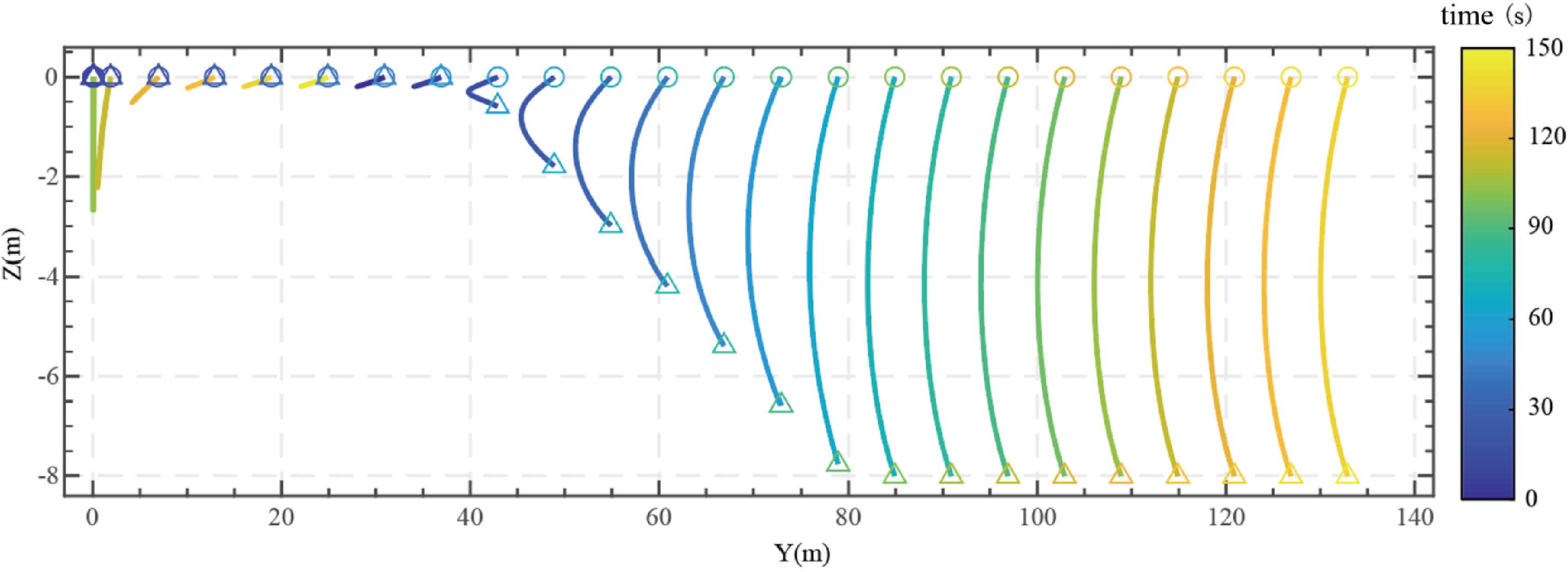}
        \vspace{0.5em}
        \text{\makecell{(c) Side view}}
    \end{minipage}

    \begin{minipage}{0.45\textwidth}
        \centering
        \includegraphics[height=\imgHeightspectrala,keepaspectratio]{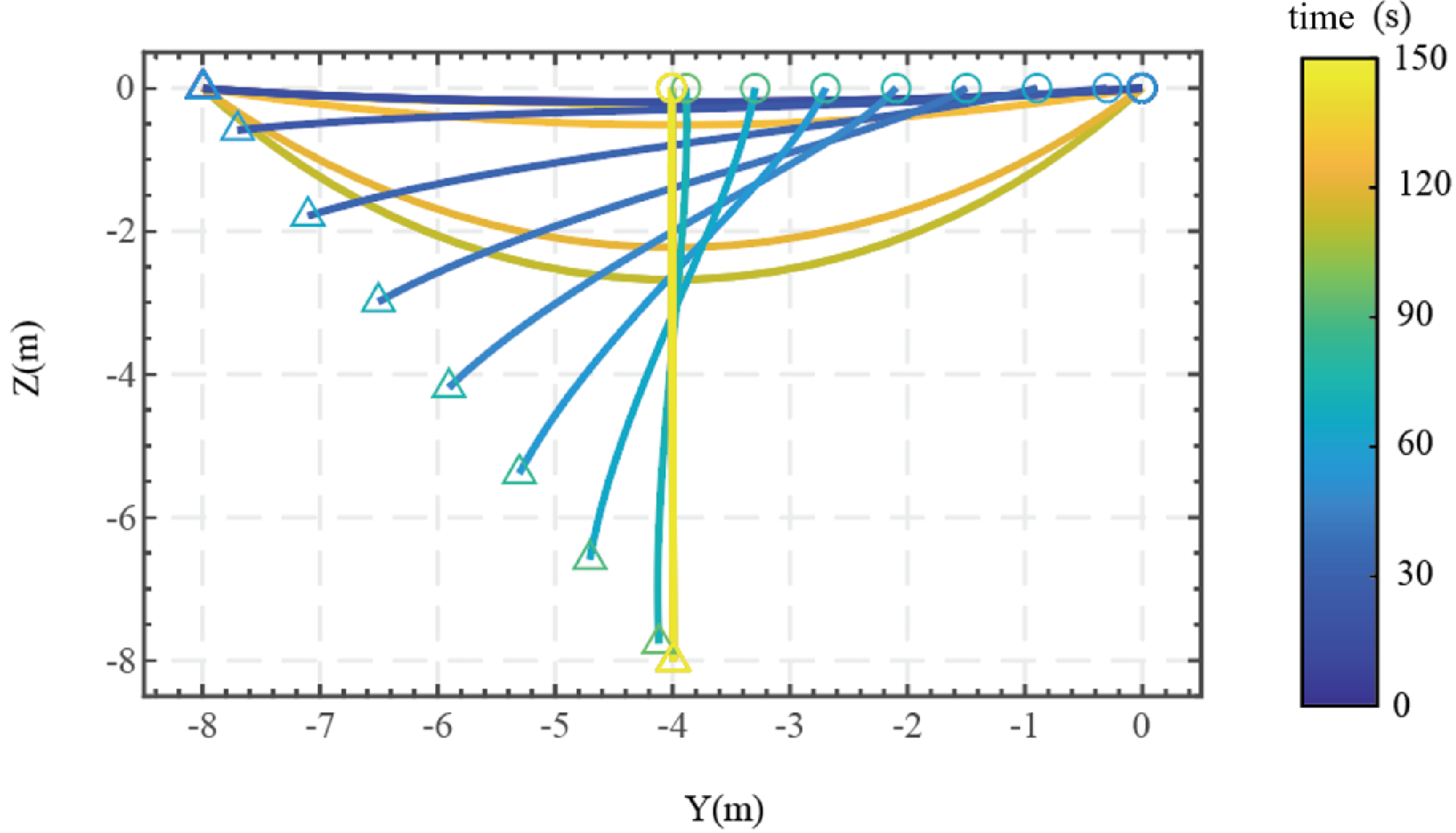}
        \vspace{0.5em}
        \text{\makecell{\centering (d) Front view}}
    \end{minipage}

    \captionof{figure}{ Cable array transformation diagram during the transformation from \\ horizontal array to vertical array} 
    \label{perspectives}
\end{center}

% \begin{center}
%     \begin{tabular}{@{}c@{\hspace{1em}}c@{}} % 两列，中间间距1em
%         % 第一行左
%         \begin{minipage}{0.45\textwidth}
%             \centering
%             \includegraphics[height=\imgHeightspectrala,keepaspectratio]{perspectives_1.png}
%             \vspace{0.5em}\\
%             (a) Spatial view
%         \end{minipage}
%         &
%         % 第一行右
%         \begin{minipage}{0.45\textwidth}
%             \centering
%             \includegraphics[height=\imgHeightspectrala,keepaspectratio]{perspectives_2.png}
%             \vspace{0.5em}\\
%             (b) Top view
%         \end{minipage}
%         \\[1em] % 行间距
        
%         % 第二行左
%         \begin{minipage}{0.45\textwidth}
%             \centering
%             \includegraphics[height=\imgHeightspectrala,keepaspectratio]{perspectives_3.png}
%             \vspace{0.5em}\\
%             (c) Side view
%         \end{minipage}
%         &
%         % 第二行右
%         \begin{minipage}{0.45\textwidth}
%             \centering
%             \includegraphics[width=\linewidth,height=\imgHeightspectrala,keepaspectratio]{perspectives_4.png} % 添加width参数
%             \vspace{0.5em}\\
%             (d) Front view
%         \end{minipage}
%     \end{tabular}
    
%     \captionof{figure}{Cable array transformation diagram during the transformation from horizontal array to vertical array} 
%     \label{perspectives}
% \end{center}

Figure \ref{Timevarying_characteristics_of_forces} illustrates the corresponding  time-varying cable tension components in the X, Y and Z directions. At the initial static stage, the system is dominated  by gravity. The cable tensions exerted on the AUVs  are mainly concentrated in the X and Y directions. At $t \approx 10$ s, the dual-AUV system initiates acceleration along the X direction. Consequently, the cable gradually transitions to the taut state, leading to a significant increase in the force components exerted on the AUVs in the X and Y directions. As the system reaches a steady motion state at about $t$=25 s, due to symmetric geometry of the dual-AUV system, the two AUVs experience almost identical forces in the X and Z directions, while the forces in the Y direction are equal in magnitude but opposite in direction. The steady motion state persists until $t$=55 s, after which the lateral separation in the Y direction gradually decreases while the vertical separation in the Z direction increases. Correspondingly, the cable transitions from the taut state to slack state, with a reduction in tension exerted  on the endpoints. As shown in this figure, a subsequent tension peak is observed for a short period at $t \approx 100$ s, when the cable  transitions into another taut state and reaches an extension of nearly 8 m. The generation of this tension peak is primarily attributed to the existence of velocity components in the X, Y, and Z directions, inducing a transient enhancement of the cable tension in the system response. As the velocity components in Y and Z directions gradually decrease and eventually reach zero, the cable tensions are reduced correspondingly, and the system reaches a steady motion state along the X direction. At this stage, the symmetric property of the cable configuration and force conditions are no longer valid. As observed in Figure \ref{Timevarying_characteristics_of_forces}, the $AUV_2$, which is located at the lower position of the simulation domain, experiences larger forces than $AUV_1$ in both the X and Z directions. In addition, the magnitude of steady-state forces acting on this platform is also higher than that in the horizontal formation conditions. This phenomenon indicates that  the gravity components are redistributed to the X and Z  directions, leading to a gradual decrease in the force component in the Y direction, which eventually approaches zero.

\begin{center}
    \includegraphics[width=10cm]{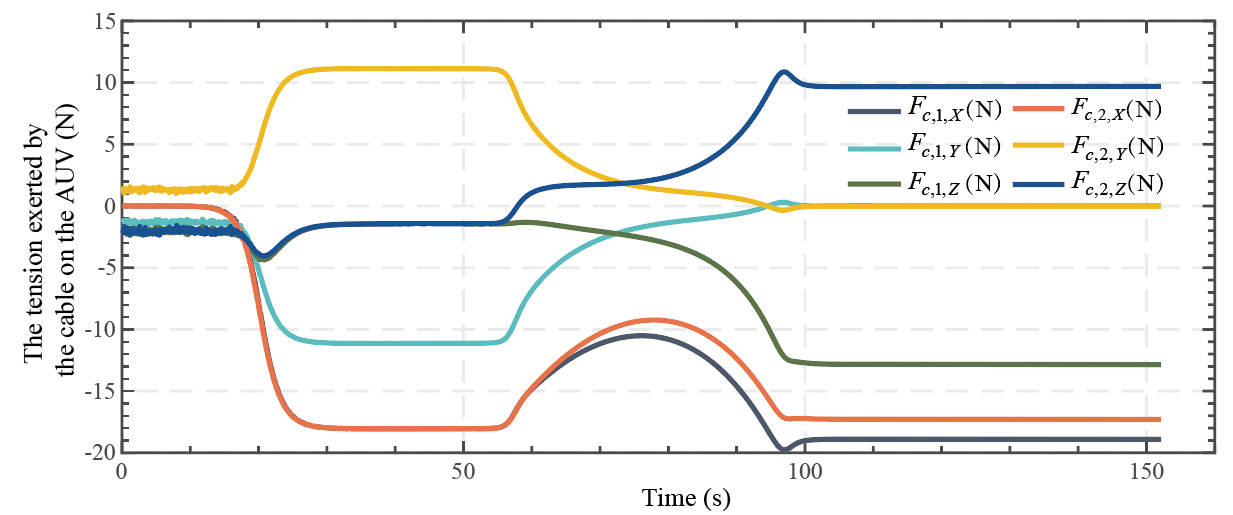}
    \captionof{figure}{Time-varying characteristics of the forces exerted on both ends of the cable by the AUV \\ when switching from a horizontal to a vertical formation} 
    \label{Timevarying_characteristics_of_forces}
\end{center}

\subsection{Simulation analysis of endpoint forces and cable configurations during the transition between taut and slack states}

To further investigate the influence of "taut" and "slack" cable states on the dynamic behavior of the system, a series of representative simulation experiments is designed and conducted. In this experiment, dynamic operating conditions involving the transition between the two  states are studied. The cable configuration is initialized as a straight line. The initial position of $AUV_1$ is set to [0, 0, 0] while $AUV_2$ is positioned at [8.74, 0, 0], ensuring the straight-line connection between the two platforms corresponds to the direction vector $t_c = [2, 0, 1]$. For the velocity settings, both platforms are simulated to move along the X direction.  A velocity of $u_{j,1} = 0.5$ m/s is prescribed for $AUV_1$, while all other velocity components are set to zero. As depicted by the curve in Figure \ref{sin_curve}, a time-varying velocity function, which is formulated as $u_{j,2} = 0.5$sin(0.0575$t$)+0.5 m/s,  is assigned to $AUV_2$ along the X direction. Similar to $AUV_1$, all other velocity components of $AUV_2$ are set to zero. The prescribed velocity settings simulate a periodic directional transition between $[2, 1, 0]$ and $[-2, 1, 0]$, involving substantial angular deviations. In practical operations, such angular deviations significantly impact sensing and detection performance. Therefore, the prescribed settings are both representative and practically relevant, reflecting common operational characteristics in marine exploration. 

\begin{center}
    \includegraphics[width=8cm]{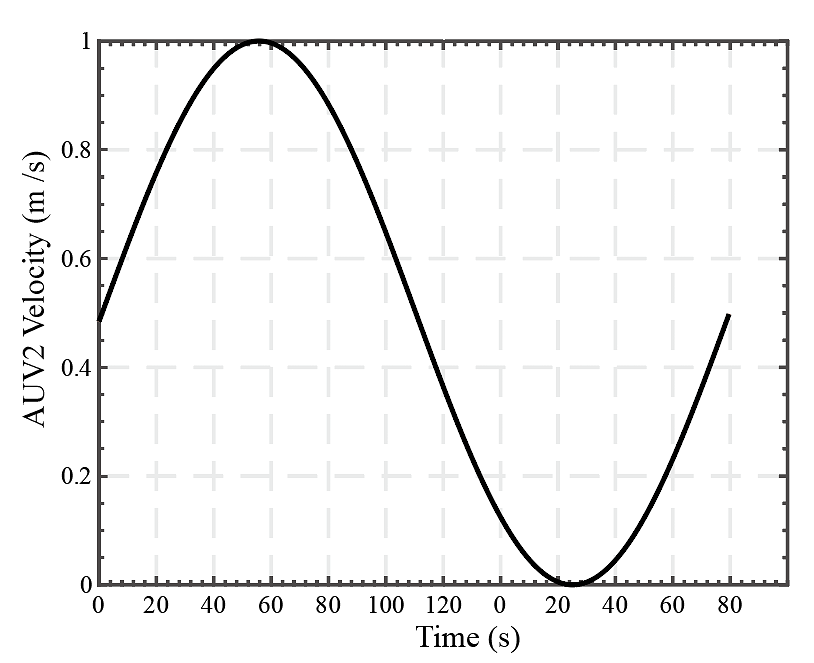}
    \captionof{figure}{$AUV_2$ velocity settings in the $X$ direction} 
    \label{sin_curve}
\end{center}

The spatial position variation of the flexible connected dual-AUV system and the cable configuration is presented in Figure \ref{Spatial_position_variation}, where the time-step is set to 3 s. The  time-varying spatial trajectories of system position are illustrated in Figure \ref{Spatial_position_variation} (a). The formation transition process in the horizontal plane is presented in Figure \ref{Spatial_position_variation} (b), while the variation of cable configurations in vertical plane is reflected in Figure \ref{Spatial_position_variation} (c).

\begin{center}
    \begin{minipage}{0.6\textwidth}
        \centering
        \includegraphics[width=9.5cm]{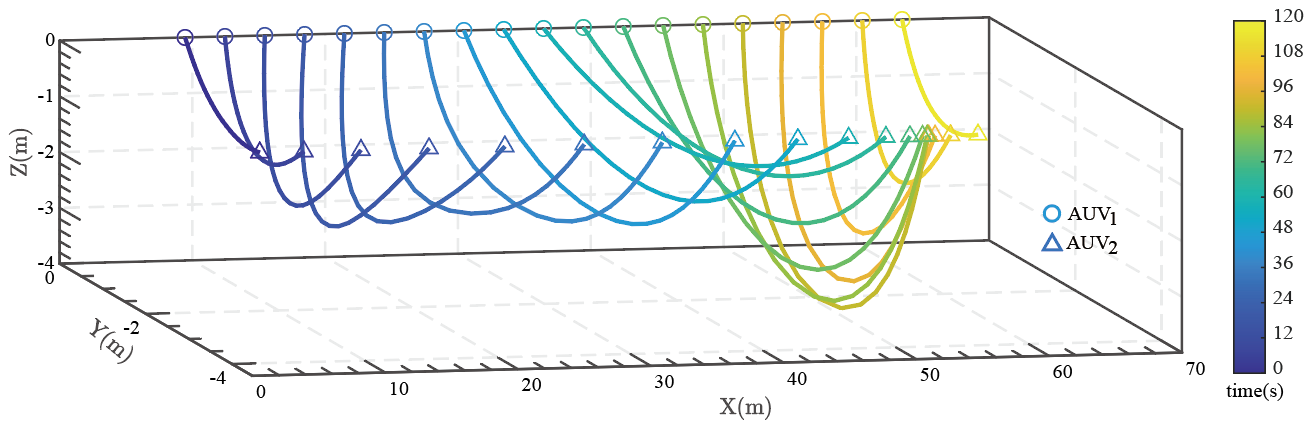}
        \vspace{0.5em}
        \text{\makecell{(a) Position variation of the cable and dual-AUV system in  spatial view}}

    \end{minipage}

    \begin{minipage}{0.6\textwidth}
        \centering
        \includegraphics[width=9.5cm]{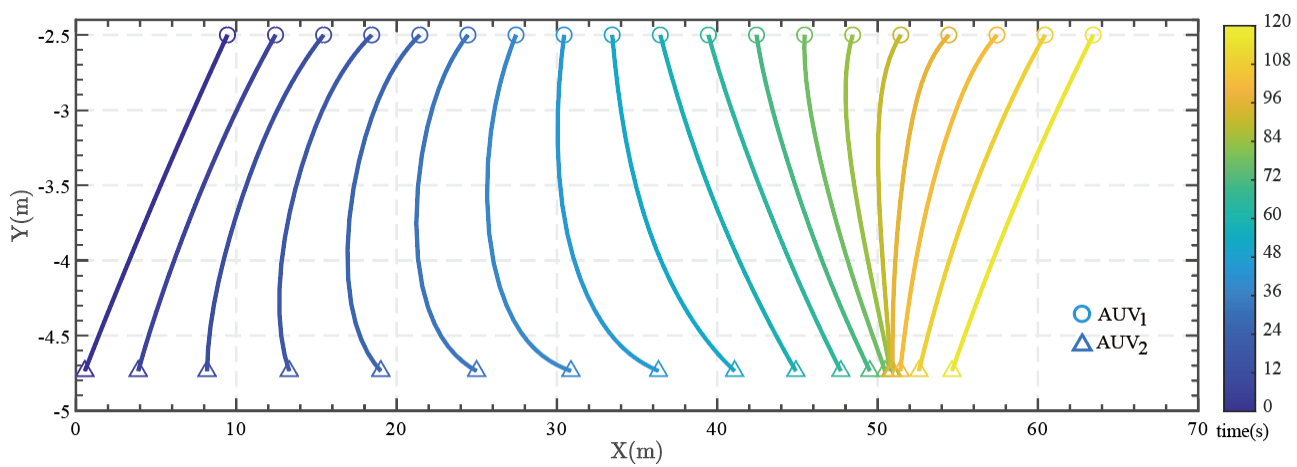}
        \vspace{0.5em}
        \text{\makecell{(b) Position variation of the cable and dual-AUV system in top view}}
    \end{minipage}
    
   \begin{minipage}{0.6\textwidth}
        \centering
        \includegraphics[width=9.5cm]{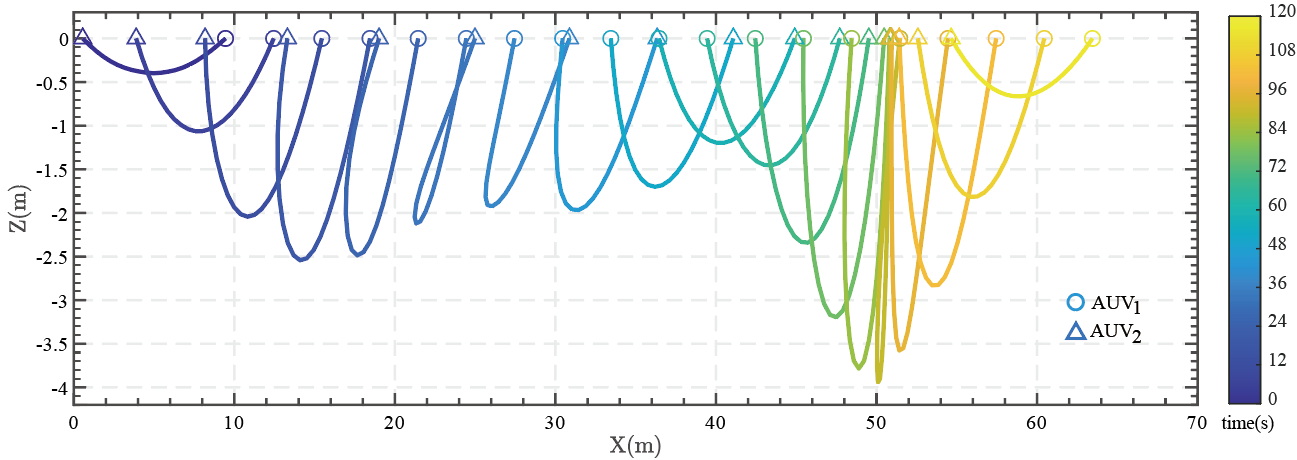}
        \vspace{0.5em}
        \text{\makecell{(c) Position variation of the cable and dual-AUV system in side view}}
    \end{minipage}

    \captionof{figure}{Spatial position variation of the flexible connected dual-AUV system and the cable configuration} 
    \label{Spatial_position_variation}
\end{center}

The time-varying characteristics of the external forces exerted on $AUV_1$ and $AUV_2$ are depicted in Figures \ref{MechanicalResponseCharacteristics} (a) and (b). Based on the system state determination criteria proposed in the previous research \citep{chen2024reduced}, the red markers represent the instances where the cable state is identified as the taut state, while the other instances correspond to the slack state. As presented in this figure, the expansion of the separation distance drives the system into the taut state, resulting in a significant increase in the tension exerted on the cable endpoints. Thereafter, the separation distance between the two AUV platforms is reduced in the subsequent motions, leading the cable to transition into the slack state. Correspondingly, the system tension gradually returns to a lower level. The results presented above indicate that the physical states of the cable, including the slack and taut state, have a significant influence on the tension distribution. This influence is particularly evident during the state transition process, exhibiting highly nonlinear characteristics in dynamic responses.

To further investigate the fundamental dynamic characteristics of the system under the cable mode transition process, the spectral properties of the linearized state-transition matrices $A_r^{c,d}$, established for both the 'slack' and 'taut' configurations, are analyzed to characterize the dynamic properties of the system. The results of spectral analysis are illustrated in Figure \ref{spectral}, where the eigenvalues in the complex plane for the cable in the 'slack' and  'taut' mode are  presented in Figure \ref{spectral} (a) and (b) respectively. The horizontal and vertical axes represent the real and imaginary parts of the eigenvalues. The dashed unit circle serves as the stability boundary, demarcating the stable and unstable regions of the system modes. The blue markers in the figure correspond to eigenvalues with magnitudes less than 1, representing the stable modes of the system, while the red markers correspond to eigenvalues with magnitudes greater than or equal to 1, indicating unstable modes. As observed in these figures, most eigenvalues are located inside the unit circle, indicating the great stability of the reduced-order model under the current motion conditions. The unstable modes account for a lower percentage and are primarily concentrated around the boundary of the unit circle , indicating  a relatively constrained influence on the overall system dynamics.

\newlength{\imgHeightCableMechanicalResponseCharacteristics}
\setlength{\imgHeightCableMechanicalResponseCharacteristics}{6cm}

\begin{center}
    \begin{minipage}{0.48\textwidth}
        \centering
        \includegraphics[width=\imgHeightCableMechanicalResponseCharacteristics,keepaspectratio]{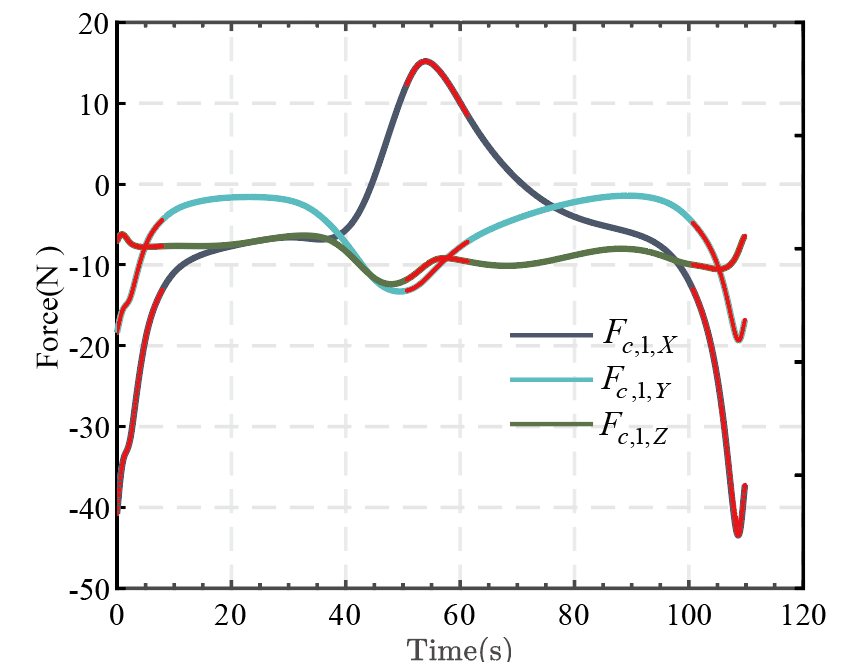}
        \vspace{0.5em}
        \text{\makecell{(a) The time-varying forces exerted on $AUV_1$}}

    \end{minipage}
    \hfill
    \begin{minipage}{0.48\textwidth}
        \centering
        \includegraphics[width=\imgHeightCableMechanicalResponseCharacteristics,keepaspectratio]{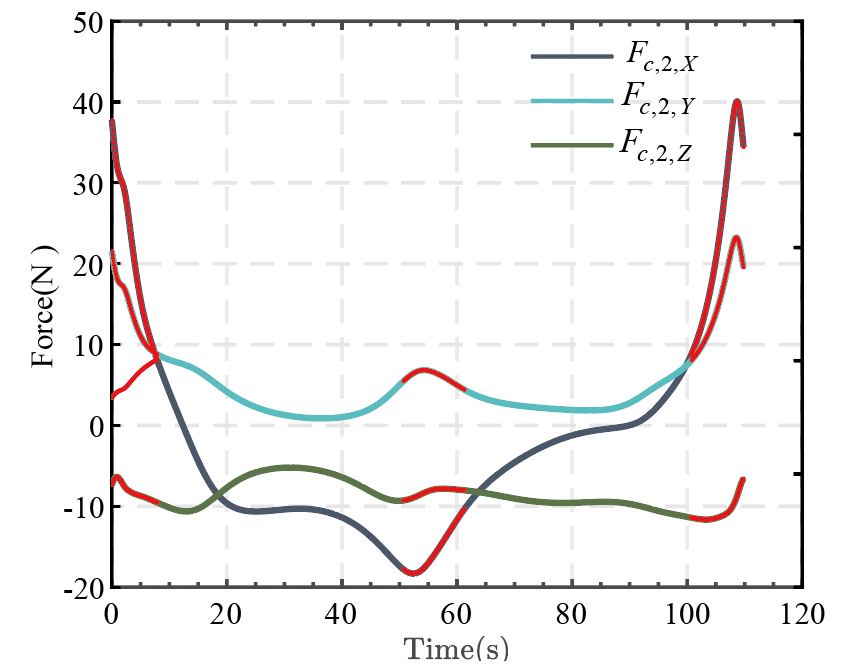}
        \vspace{0.5em}
        \text{\makecell{(b) The time-varying forces exerted on $AUV_2$}}
    \end{minipage}

    \captionof{figure}{Mechanical response characteristics of the flexible connected dual-AUV system} 
    \label{MechanicalResponseCharacteristics}
\end{center}

In the slack state, the phase angle remains below 0.523 rad, which suggests that the system is governed by low-frequency oscillations. In contrast, in the taut state, the upper bound of phase angles is raised to 0.963 rad, exhibiting higher oscillation frequencies. This phenomenon is primarily attributed to the distinct dominant oscillation mechanisms in the two modes. For a system operating in the taut state, the structural elastic forces of the cable become the dominant factor. The strengthening of the inherent dynamic effects leads to  higher-frequency oscillatory responses. In contrast, for a system in the slack state,  the hydrodynamic forces dominate the dynamics, resulting in lower-frequency system responses. The analysis results suggest that for the flexibly connected dual-AUV system experiencing multiple cable mode transitions, the associated variations in dynamic characteristics should be fully considered.

To quantify the impact of the main system modes on the response of the flexible cable, the eigenvectors and norms of the first six dominant modes are computed to evaluate the contribution of each mode in the state variables of different dimensions. Figure \ref{Distribution_of_Dominant_Mode_Bent} demonstrates the spatial distribution of the dominant mode influence on the state variables under the slack state of the cable.  The color gradient from blue to yellow represents the magnitude of the vector norm, indicating the progression in modal influence from weak to strong. Figure \ref{Distribution_of_Dominant_Mode_Bent} (a) - (c) illustrate  influence of the dominant mode on the cable nodes along the X, Y and Z axes of the Earth-fixed system. The results indicat that, in the slack state, the third and fourth dominant modes exert the most significant influence on the middle segment of the cable in the X and Y directions. This result diverges from the previous  simulation, in which the strongest modal influence is observed at the high-index nodes near ${AUV}_2$. In contrast to the previous asymmetric condition involving one stationary end and one moving end, in the current case, the entire system moves collectively in the X direction. As a result, this condition  produces the largest displacements at the midpoints, resulting in greater contribution to the modal response.

% \begin{center}
%     \begin{minipage}{0.48\textwidth}
%         \centering
%         \includegraphics[width=8cm]{spectral1.png}
%         \vspace{0.5em}
%         \text{\makecell{(a) The distribution of eigenvalues for a cable \\ in the bent mode}}

%     \end{minipage}
%     \hfill
%     \begin{minipage}{0.48\textwidth}
%         \centering
%         \includegraphics[width=8.5cm]{spectral2.png}
%         \vspace{0.5em}
%         \text{\makecell{(b) The distribution of eigenvalues for a cable \\ in the stretched mode}}
%     \end{minipage}

%     \captionof{figure}{Spectral Distribution of the Flexible Cable Dual-AUV System under “Bent” and “Stretched” Cable States} 
%     \label{spectral}
% \end{center}

\newlength{\imgHeightspectral}
\setlength{\imgHeightspectral}{6cm}

\begin{center}
    \begin{minipage}{0.48\textwidth}
        \centering
        \includegraphics[width=\imgHeightspectral,keepaspectratio]{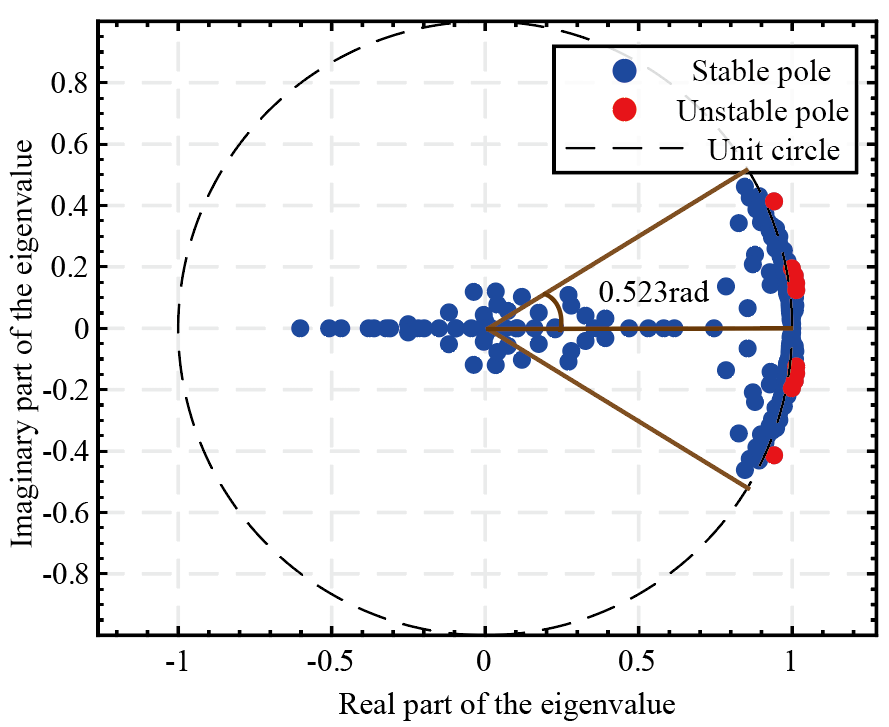}
        \vspace{0.5em}
        \text{\makecell{(a) The distribution of eigenvalues for a cable \\ in the bent state}}

    \end{minipage}
    \hfill
    \begin{minipage}{0.48\textwidth}
        \centering
        \includegraphics[width=\imgHeightspectral,keepaspectratio]{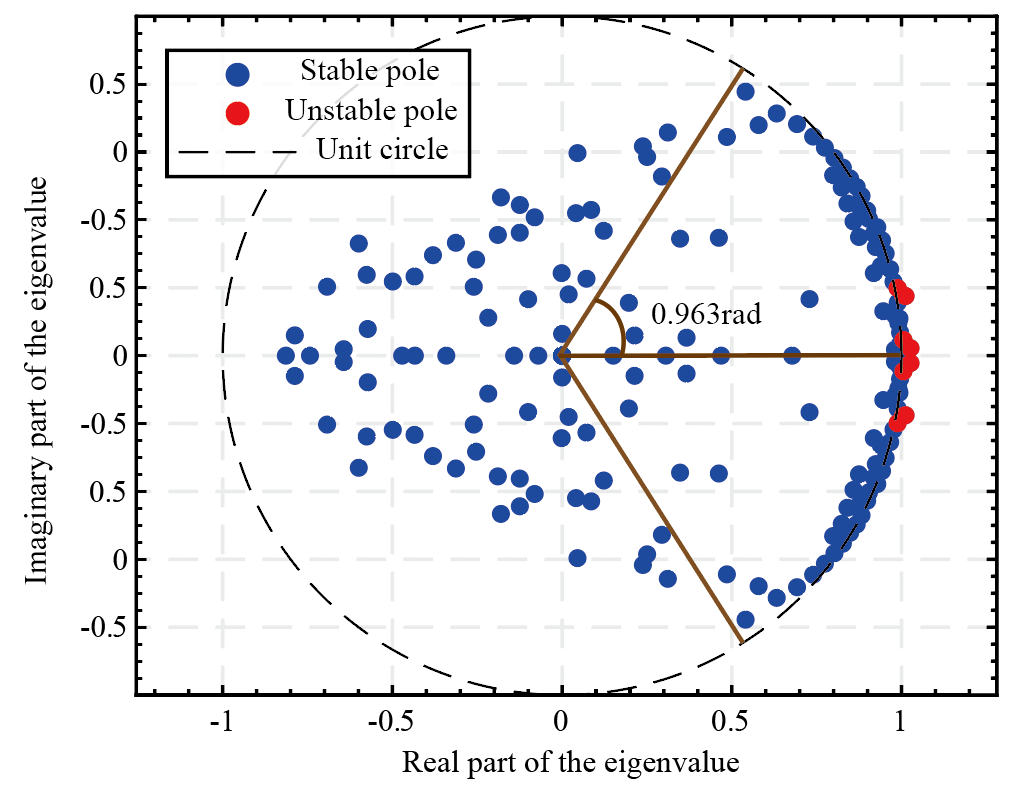}
        \vspace{0.5em}
        \text{\makecell{(b) The distribution of eigenvalues for a cable \\ in the stretched state}}
    \end{minipage}

    \captionof{figure}{Spectral Distribution of the Flexible Cable Dual-AUV System under “Bent” and “Stretched” Cable States} 
    \label{spectral}
\end{center}

\newlength{\imgHeightDistributionofDominantModeBent}
\setlength{\imgHeightDistributionofDominantModeBent}{6cm}

\begin{center}
    \begin{minipage}{0.48\textwidth}
        \centering
        \includegraphics[width=\imgHeightspectral,keepaspectratio]{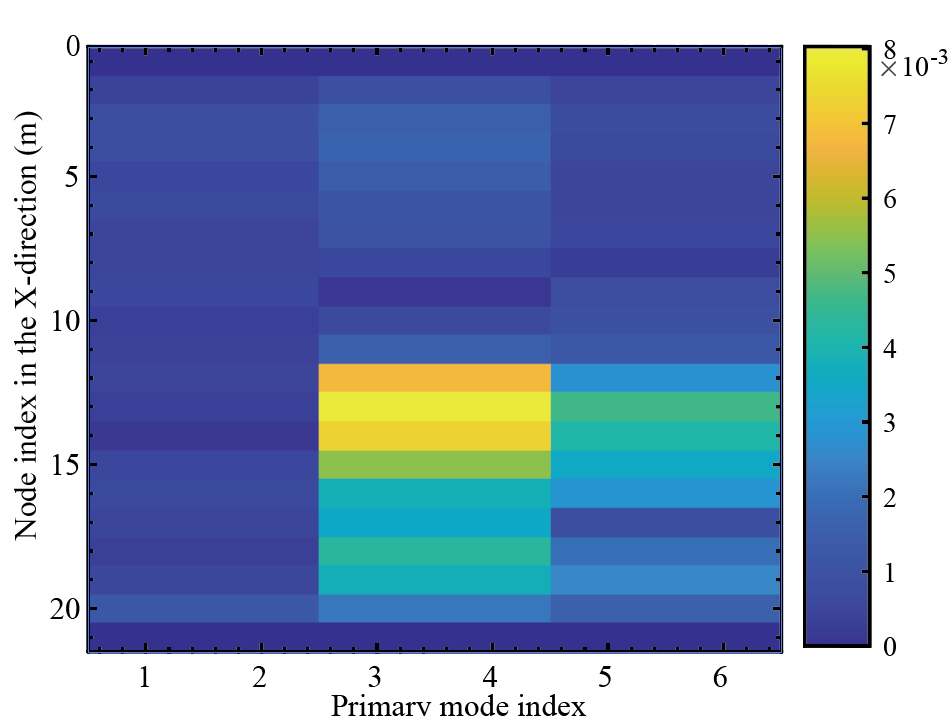}
        \vspace{0.5em}
        \text{\makecell{(a) Influence from dominant mode on cable node \\ positions in $X$ direction}}

    \end{minipage}
    \hfill
    \begin{minipage}{0.48\textwidth}
        \centering
        \includegraphics[width=\imgHeightspectral,keepaspectratio]{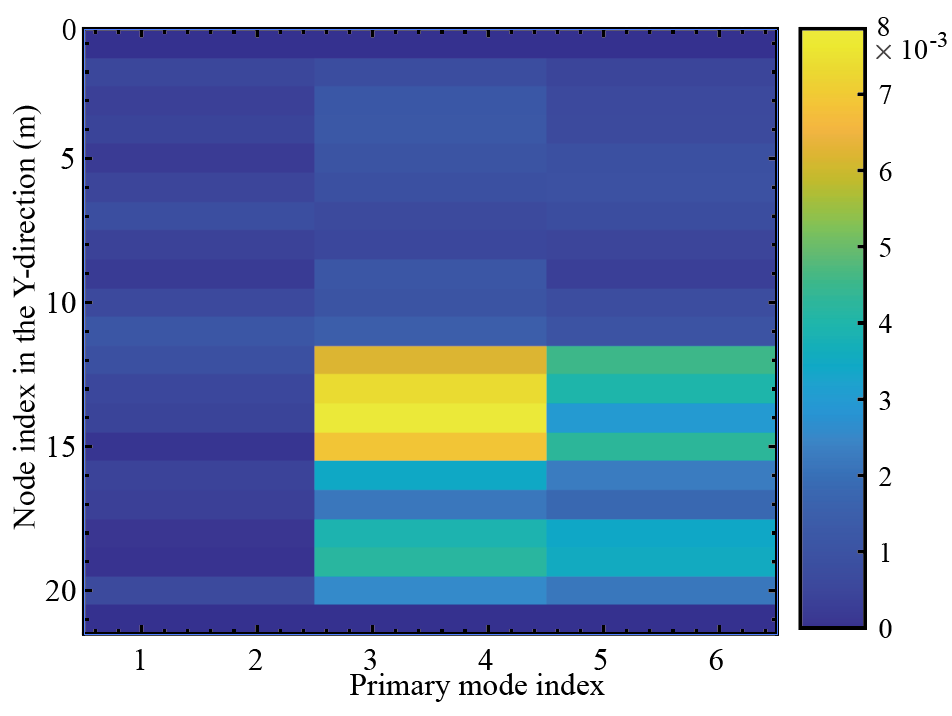}
        \vspace{0.5em}
        \text{\makecell{(b) Influence from dominant mode on cable node \\ positions in $Y$ direction}}
    \end{minipage}
    
   \begin{minipage}{0.48\textwidth}
        \centering
        \includegraphics[width=\imgHeightspectral,keepaspectratio]{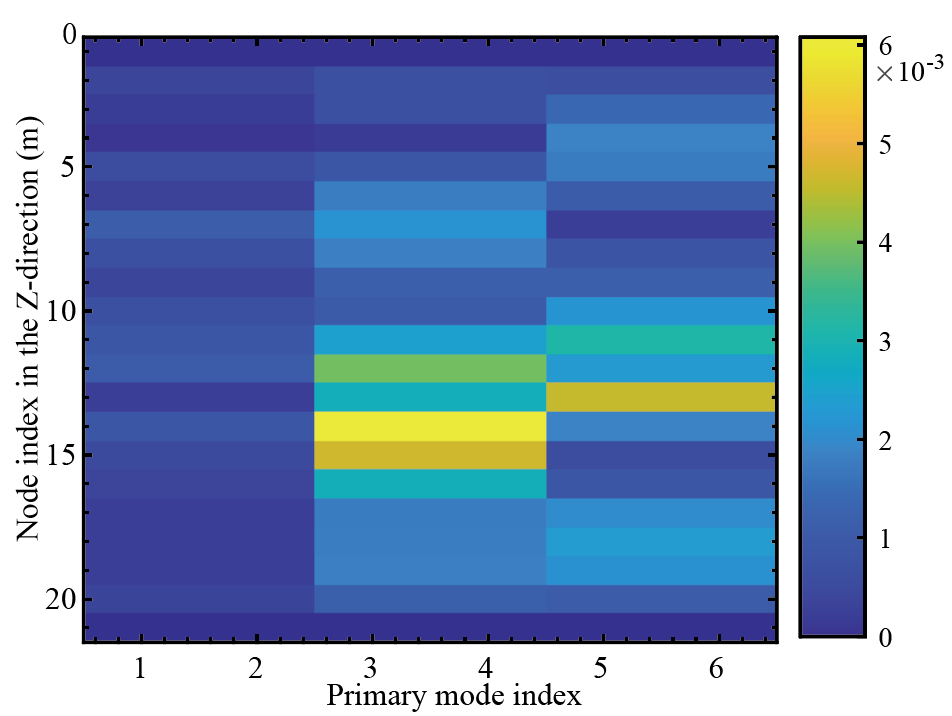}
        \vspace{0.5em}
        \text{\makecell{(c) Influence from dominant mode on the cable node \\ positions in $Z$ direction}}
    \end{minipage}
    \hfill
    \begin{minipage}{0.48\textwidth}
        \centering
        \includegraphics[width=\imgHeightspectral,keepaspectratio]{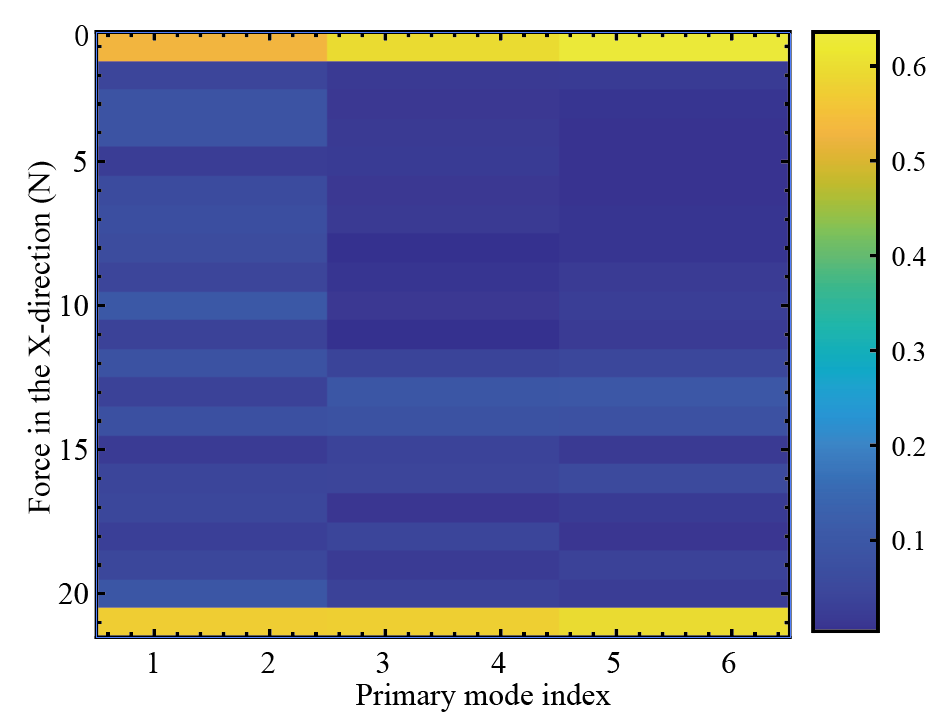}
        \vspace{0.5em}
        \text{\makecell{(d) Influence from dominant mode on the cable node \\ forces in $X$ direction}}
    \end{minipage}

   \begin{minipage}{0.48\textwidth}
        \centering
        \includegraphics[width=\imgHeightspectral,keepaspectratio]{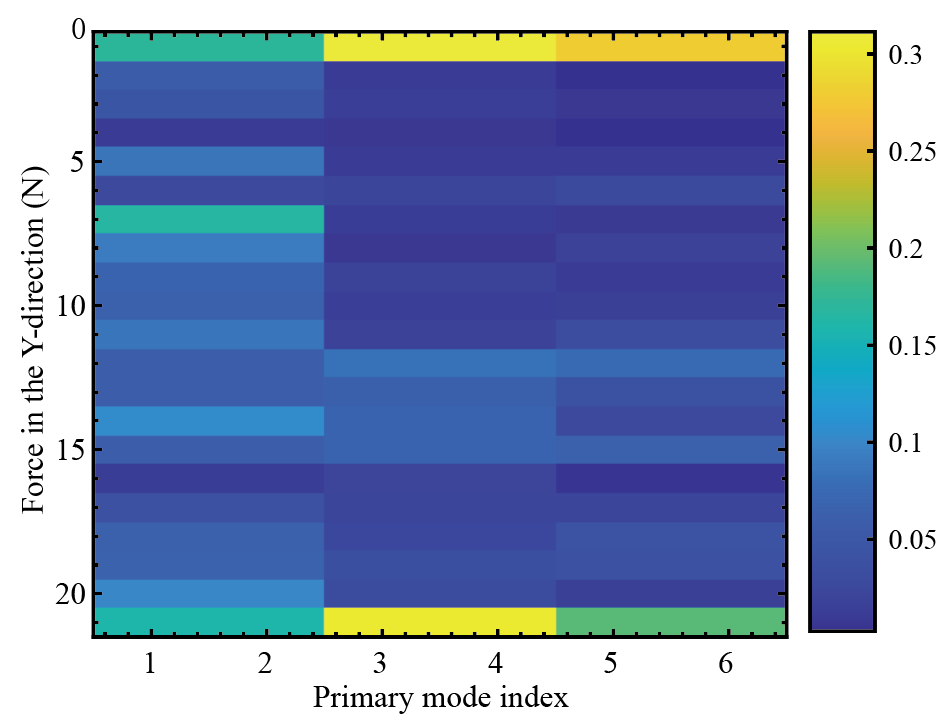}
        \vspace{0.5em}
        \text{\makecell{(e) Influence from dominant mode on cable node \\ forces  in $Y$ direction}}
    \end{minipage}
    \hfill
    \begin{minipage}{0.48\textwidth}
        \centering
        \includegraphics[width=\imgHeightspectral,keepaspectratio]{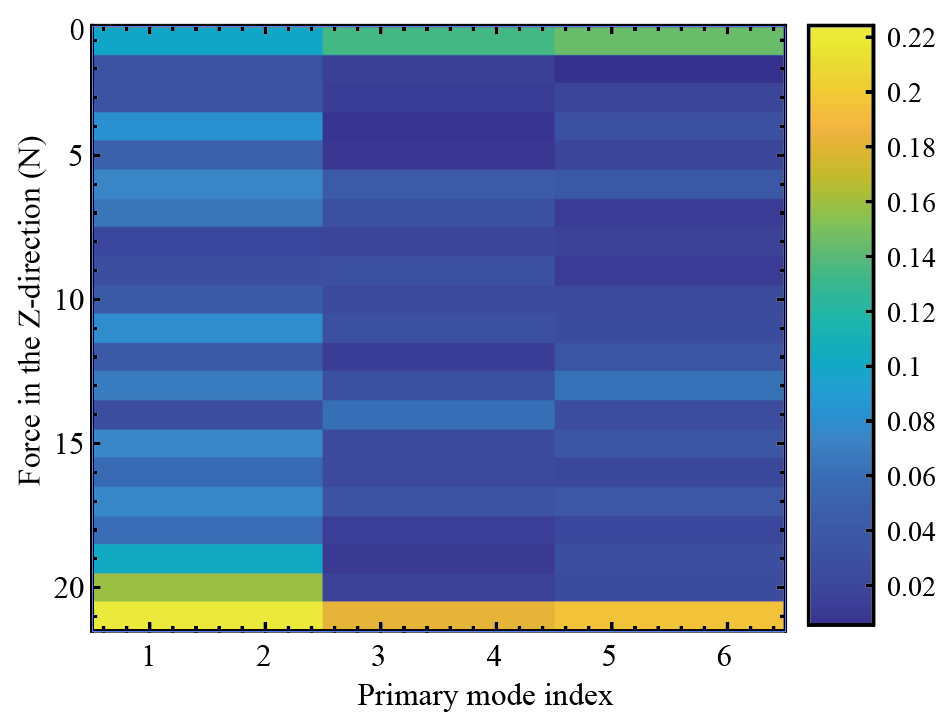}
        \vspace{0.5em}
        \text{\makecell{(f) Influence from dominant mode on cable node \\ forces  in $Z$ direction}}
    \end{minipage}

    \captionof{figure}{Distribution of Dominant Mode Influence on Cable Nodes and Forces under \\ the “Bent” State of the System} 
    \label{Distribution_of_Dominant_Mode_Bent}

\end{center}

Figure \ref{Distribution_of_Dominant_Mode_Stretched} illustrates the distribution characteristics of each state variable under the first six dominant modes when the cable is in the taut state. The color transition from blue to yellow corresponds to an increase in the magnitude of the eigenvector norm, indicating the gradual strengthening of modal influence on system responses. The results show that the cable tension is primarily dominated by the first, second, fifth and sixth modes, while the third and fourth modes mainly participate in the position response and contribute relatively less to the force response. From the perspective of spatial distribution, attributed to reduced boundary constraints and inherently higher degrees of freedom, the central segment of the cable is the most sensitive to modal variations. In contrast, the mechanical response reaches the maximum level near both endpoints, implying that the internal forces are ultimately concentrated and transmitted to the AUV located at the cable ends, which is consistent with physical intuition. Furthermore, a certain degree of decoupling between position and force modes is observed in the taut state. This finding further indicates that when the cable is in the taut state, the system position response is predominantly governed by internal elasticity and structural properties, while the hydrodynamic effects closely related to positional variation are no longer the prevailing factor. Consequently, the overall system becomes less sensitive to motion inputs from the two AUVs.

\begin{center}
    \begin{minipage}{0.48\textwidth}
        \centering
        \includegraphics[width=\imgHeightspectral,keepaspectratio]{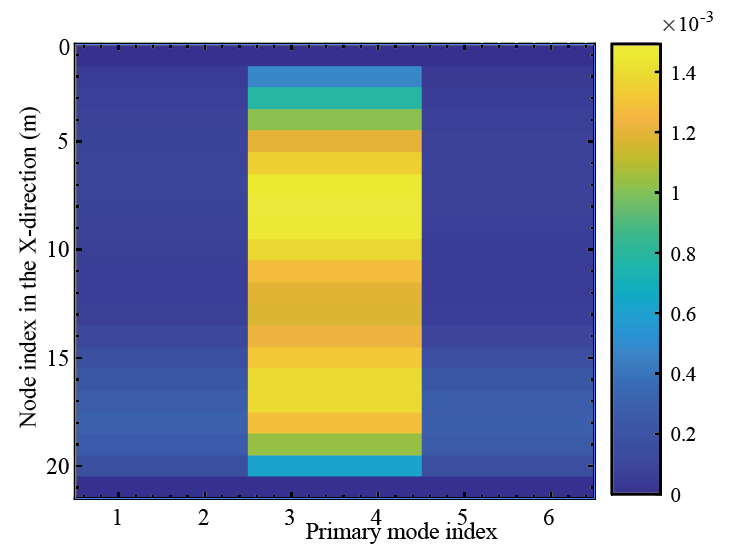}
        \vspace{0.5em}
        \text{\makecell{(a) Influence from dominant mode on cable node \\ positions in $X$ direction}}
    \end{minipage}
    \hfill
    \begin{minipage}{0.48\textwidth}
        \centering
        \includegraphics[width=\imgHeightspectral,keepaspectratio]{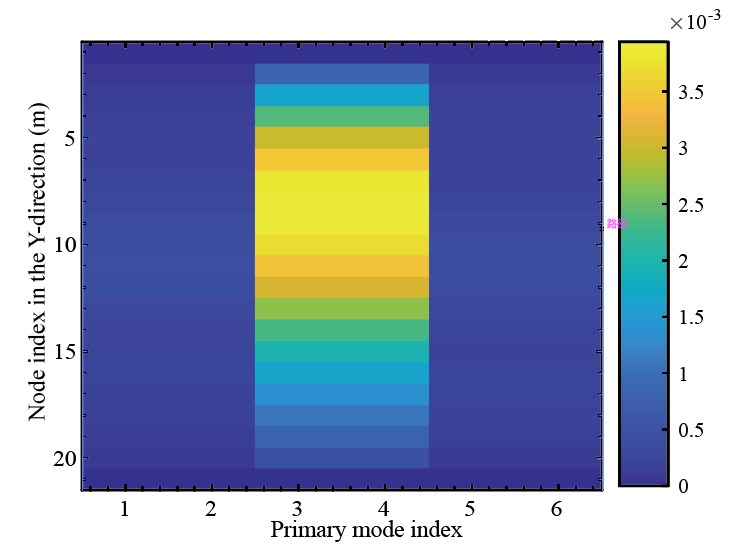}
        \vspace{0.5em}
        \text{\makecell{(b) Influence from dominant mode on cable node \\ positions in $Y$ direction}}
    \end{minipage}
    
   \begin{minipage}{0.48\textwidth}
        \centering
        \includegraphics[width=\imgHeightspectral,keepaspectratio]{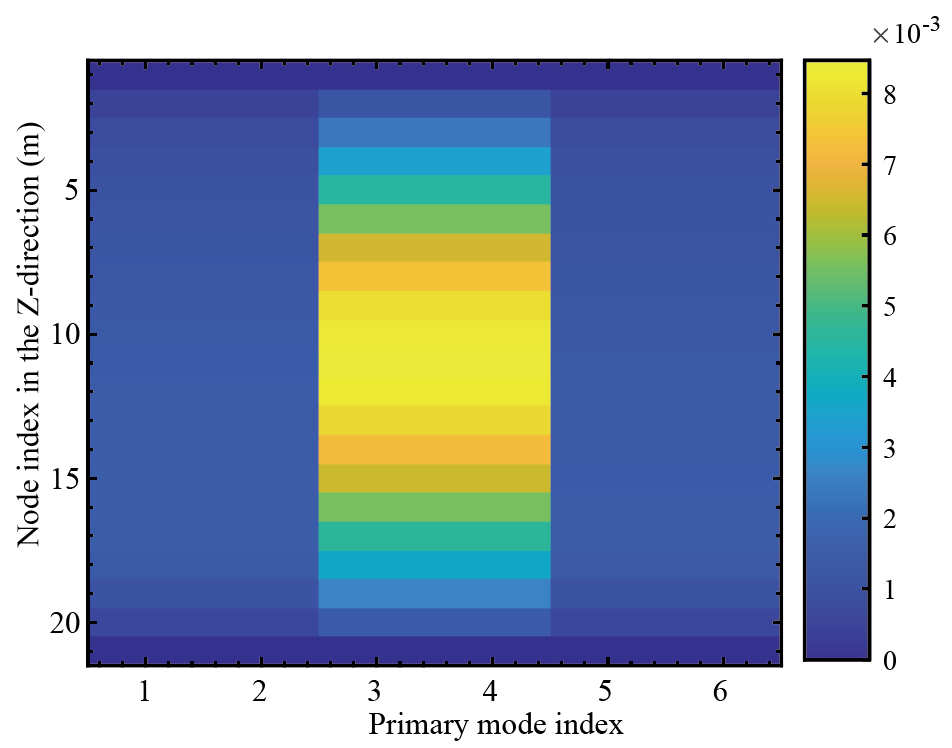}
        \vspace{0.5em}
        \text{\makecell{(c) Influence from dominant mode on the cable node \\ positions in $Z$ direction}}
    \end{minipage}
    \hfill
    \begin{minipage}{0.48\textwidth}
        \centering
        \includegraphics[width=\imgHeightspectral,keepaspectratio]{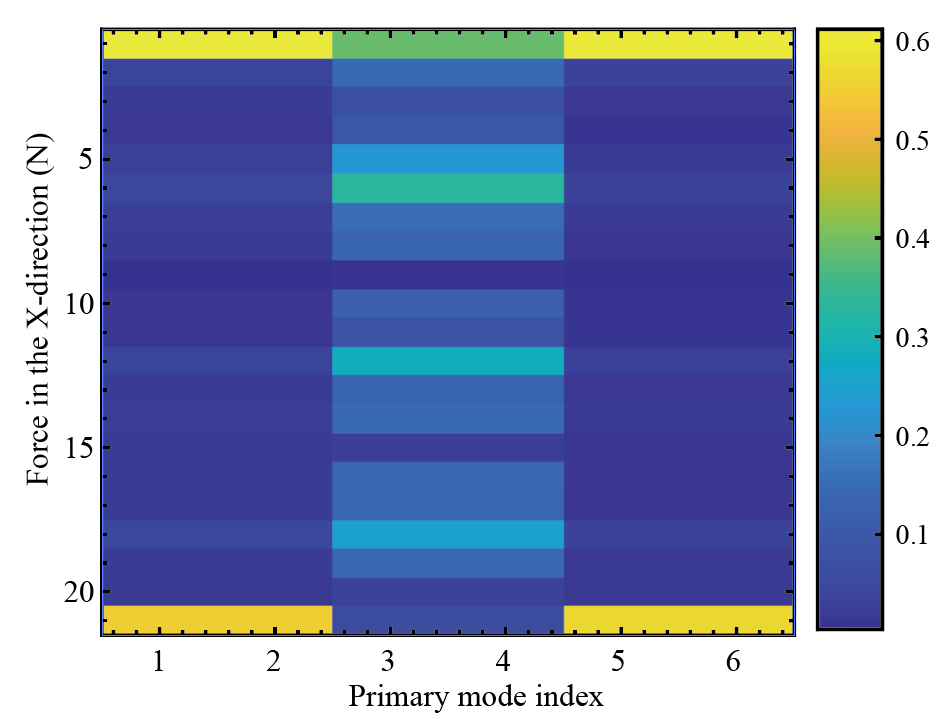}
        \vspace{0.5em}
        \text{\makecell{(d) Influence from dominant mode on the cable node \\ forces in $X$ direction}}
    \end{minipage}

   \begin{minipage}{0.48\textwidth}
        \centering
        \includegraphics[width=\imgHeightspectral,keepaspectratio]{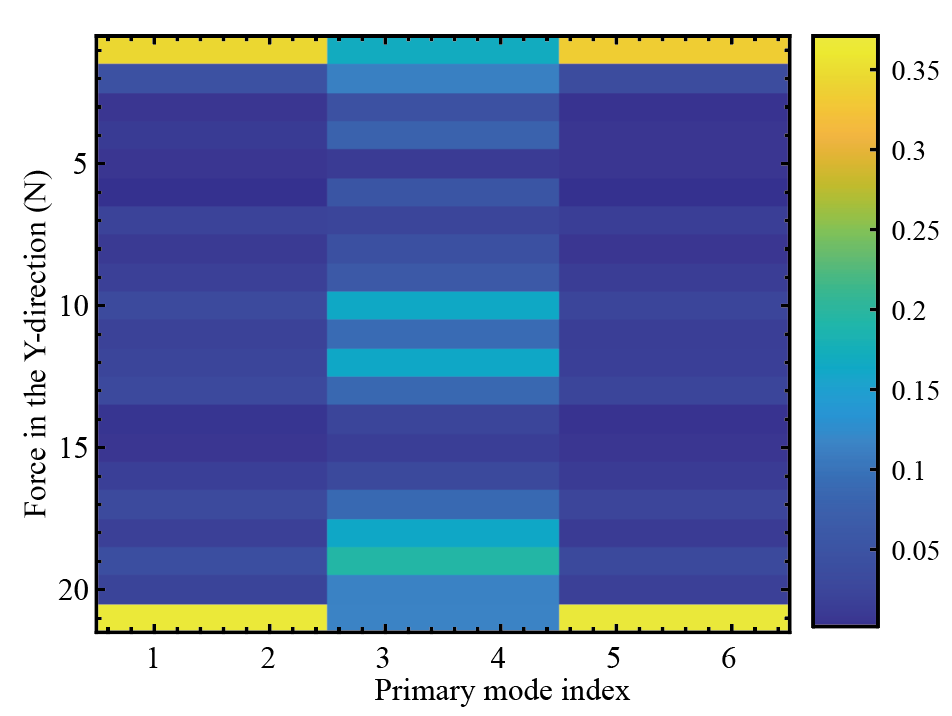}
        \vspace{0.5em}
        \text{\makecell{(e) Influence from dominant mode on cable node \\ forces  in $Y$ direction}}
    \end{minipage}
    \hfill
    \begin{minipage}{0.48\textwidth}
        \centering
        \includegraphics[width=\imgHeightspectral,keepaspectratio]{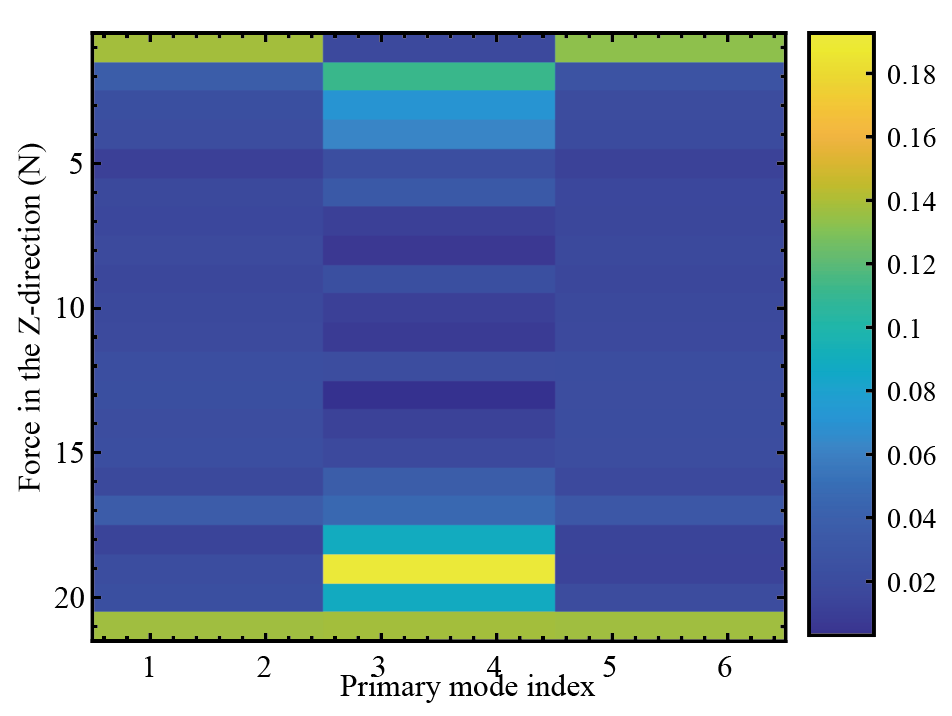}
        \vspace{0.5em}
        \text{\makecell{(f) Influence from dominant mode on cable node \\ forces  in $Z$ direction}}
    \end{minipage}

    \captionof{figure}{Distribution of Dominant Mode Influence on Cable Nodes and Forces under \\ the “Stretched” State of the System} 
    \label{Distribution_of_Dominant_Mode_Stretched}

\end{center}

% These results demonstrate that the dynamic characteristics of the slack state are more sensitive to geometric variations, further confirming that hydrodynamics play a decisive role in the overall system response under such conditions. 

Figure \ref{Distribution_of_Dominant_Mode_Bent} illustrates the distribution characteristics of the  positional and tension responses associated with the first six dominant modes under the slack condition. Similar to the taut state, the positional response remains most sensitive at the middle segment, while the mechanical response is most pronounced near the endpoints. However, unlike the decoupling behavior observed in the taut condition,  the slack state exhibits significant modal coupling between positional and tension responses. The generation mechanism of this coupling behavior is attributed to the dominance of hydrodynamic forces in the system dynamics, of which the magnitude and distribution are highly sensitive to the cable configuration. Therefore, positional variations and mechanical responses become tightly coupled at the modal level. These results demonstrate that the dynamic characteristics of the cable in the slack state  are more sensitive to geometric variation, providing further evidence of the decisive influence of hydrodynamics on the overall system response.

\section{Conclusions}

A comprehensive physical model of the flexibly connected dual-AUV system is established in this paper, accompanied by systematic simulation and experimental analyses. First, the coordinate systems are defined, and the dynamic equations of both the flexible cable and the AUV are derived. In the modeling process, hydrodynamic effects, cable axial stiffness and bending characteristics, AUV rigid-body dynamics and the corresponding boundary conditions are integrated, thereby establishing a comprehensive mathematical model capable of accurately capturing the coupled interaction of the dual-AUV system and the flexible cable. On this basis, numerical simulations are conducted to investigate the influence of critical factors, including discretization resolution, current velocity, cable material properties, spatial configuration, and non-uniform AUV motions, which provide a systematic analysis of the dynamic characteristics and the configuration transitions of the cable. The simulation results demonstrate that the proposed model is capable of accurately describing the dynamic responses of the flexibly connected dual-AUV system under different operating conditions, providing a reliable foundation for system design and control in complex marine environments. The research presented in this paper serves as a crucial basis for future developments in related technologies, such as towing system design, performance optimization, and advanced control strategy development.

\section*{Acknowledgments }

This work was supported by the National Key R\&D Program of China (Grant No. 2022YFE0204600), the Program of the State Key Laboratory of Robotics at Shenyang Institute of Automation, Chinese Academy of Sciences (Grant No. 2024-Z05), the National Natural Science Foundation of China (GrantNo. 41976183 \& 52271353), the Fundamental Research Program of Shenyang Institute of Automation, Chinese Academy of Sciences (Grant No. 2022JC3K05 \& 202400047601 \& 2024JC3K02)

\end{document}